\documentclass[11pt]{article}
\usepackage{fullpage}

\usepackage[utf8]{inputenc} 
\usepackage[T1]{fontenc}    
\usepackage{hyperref}       
\usepackage{url}            
\usepackage{booktabs}       
\usepackage{amsfonts}       
\usepackage{nicefrac}       
\usepackage{microtype}      
\usepackage{tcolorbox}
\usepackage{diagbox}
\usepackage{enumerate}
\usepackage[shortlabels]{enumitem}
\usepackage{algorithmic}
\usepackage[vlined, linesnumbered, ruled]{algorithm2e}

\usepackage{subfigure}
\usepackage{graphicx} 
\usepackage{caption}
\usepackage{amsmath}
\usepackage{amsthm}
\usepackage{amssymb}
\usepackage{tikz}
\usepackage{tablefootnote}
\usepackage{multirow}
\usepackage{enumerate}
\usepackage{color}
\usepackage{xcolor}
\usepackage{natbib}

\usepackage{titletoc}
\usepackage[page, header, toc, page]{appendix}
\allowdisplaybreaks[4]

\usepackage{mathrsfs}

\usepackage{hyperref}
\usepackage{bm,bbm,todonotes}

\allowdisplaybreaks

\newtheorem{Rema}{Remark}[section]

\newcommand{\eq}[1]{Eq.~\eqref{#1}}

\hypersetup{
	colorlinks=true,
	filecolor=blue,
	citecolor = blue,
	urlcolor=cyan,
}

\newcommand{\1}{\mathbbm{1}}				
\renewcommand{\P}{\mathbb{P}}				
\newcommand{\E}{\mathbb{E}}					
\newcommand{\agp}[1]{\langle #1 \rangle}	

\newcommand{\normmm}[1]{\lvert\kern-0.25ex\lvert\kern-0.25ex\lvert #1 \rvert\kern-0.25ex\rvert\kern-0.25ex\rvert}

\newcommand{\R}{\mathbb{R}}		

\newcommand{\argmin}{\operatornamewithlimits{arg\,min}}
\newcommand{\argmax}{\operatornamewithlimits{arg\,max}}
\newtheorem{Def}{Definition}[section]
\newtheorem{Lem}[Def]{Lemma}
\newtheorem{Thm}[Def]{Theorem}
\newtheorem{Cor}[Def]{Corollary}

\title{
	Last-Iterate Analyses of FTRL with the 1/2-Tsallis Entropy in Stochastic Bandits
}

\author{
Jingxin Zhan
\thanks{School of Mathematical Sciences, Peking University, Beijing, China 100871; email: \texttt{bjdxzjx@pku.edu.cn}.}
\and
Yuze Han
\thanks{Center for Applied Statistics and School of Statistics, Renmin University of China, Beijing, China; email: \texttt{hanyuze97@ruc.edu.cn}.}
\and
Zhihua Zhang
\thanks{School of Mathematical Sciences, Peking University; email: \texttt{zhzhang@math.pku.edu.cn}.}
}

\begin{document}

\maketitle

\begin{abstract}
The convergence analysis of online learning algorithms 
    is central to machine learning theory, where the last-iterate convergence is particularly important, 
as it captures the learner’s actual decisions and describes the evolution of the learning process over time. 
However, in multi-armed bandits, most existing algorithmic analyses mainly focus on the order of regret, 
while the last-iterate
(simple regret)
convergence rate remains less explored---especially for the widely studied 
Follow-the-Regularized-Leader (FTRL) algorithms. 
Recently, FTRL with the $1/2$-Tsallis entropy regularizer $\Psi(p) = -4\sum_{i=1}^d \sqrt{p_i}$  \citep[the $1/2$-Tsallis-INF algorithm, by][]{zimmert2021tsallis} was shown to achieve the desirable Best-of-Both-Worlds (BOBW) guarantees and perform well in both adversarial and stochastic
settings.
Nevertheless, its last-iterate convergence rate has not yet been fully studied. This paper studies the $1/2$-Tsallis-INF algorithm in stochastic bandits and shows that its sampling simple regret decays at rate $\mathcal{O}(t^{-1})$, without requiring the optimal arm to be unique. Under a unique optimal arm, we further show that the expected Bregman divergence induced by $\Psi$ between the point mass on the optimal arm and the sampling distribution at iteration $t$ decays at rate $\mathcal{O}(t^{-1/2})$. Matching lower bounds under the same uniqueness condition show that both exponents of $t$ are tight.
\end{abstract}

\section{Introduction}
The multi-armed bandit (MAB) problem is a central model in online learning \citep{thompson1933likelihood,robbins1952some}. 
Early work focused on the stochastic setting \citep{thompson1933likelihood,lai1985asymptotically,auer2002finite}, where the losses $\ell_1,\ell_2,\ldots\in[0,1]^d$ are i.i.d.\ draws from an unknown distribution $\mathcal{D}$. 
In this regime, a rich class of algorithms based on Upper Confidence Bounds (UCB) \citep{lai1985asymptotically} has been developed, many of which attain optimal logarithmic regret \citep{auer2010ucb,auer2002finite}. 
To address scenarios where stochastic assumptions may not hold, the adversarial bandit model \citep{auer2002nonstochastic}, which places no restrictions on the loss generation process, has also been widely investigated.




In recent years,  algorithms based on the Follow-The-Regularized-Leader (FTRL) framework have attracted renewed attention because of high adaptivity, compared to the classical UCB algorithms. At each time step, the algorithm samples $I_t$ according to a probability
distribution $p_t$ over $d$ arms:
\[
    p_t=\argmin_{p\in\Delta^{d-1}}\eta_t\agp{p,\hat{L}_t}+\Psi(p),
\]
where $\Delta^{d-1}$ is the probability simplex in $\R^d$, $\eta_t$ is the learning rate, $\Psi(p)$ is a convex regularization function and $
\hat{L}_t = \sum_{s=1}^{t-1} \hat{\ell}_s
$ is the cumulative estimated loss.
FTRL was first introduced into bandit research through studies of the adversarial bandit problem \citep{auer1995gambling}. 
However, for a long time, it remained theoretically unclear whether there exists an FTRL algorithm 
that can match the performance of these classical algorithms in stochastic bandits, 
namely achieving logarithmic regret. 
Only recently has it been shown that when choosing 1/2-Tsallis entropy 
\(\Psi(p) = -4\sum_{i=1}^d \sqrt{p_i}\) and \(\eta_t \sim t^{-1/2}\), 
FTRL, which is also named  $1/2$-Tsallis-INF, can achieve the Best-of-Both-Worlds (BOBW) property \citep{zimmert2021tsallis}, 
performing well in both adversarial and stochastic settings by attaining the optimal regret bound in each case (achieving the logarithmic regret bound in stochastic bandits, especially). 
It's also worth noting that the BOBW property of $1/2$-Tsallis-INF and its variants is quite broad. 
It is relevant not only for the MAB setting studied here, but also for other frameworks, including linear bandits \citep{zhao2025heavytailedlinearbanditsadversarial}, semi-bandits \citep{zimmert2019beating}, and MDPs \citep{chen2026bestofbothworldsheavytailedmarkovdecision}.

However, most existing studies on FTRL algorithms in the bandit setting 
have focused on the expected regret or its concentration properties \citep{zimmert2021tsallis,pogodin2020first}. 
To the best of our knowledge, their last-iterate behavior in stochastic bandits has not yet been fully studied.
By contrast, in the games setting, there has been a recent line of work analyzing the last-iterate convergence properties of FTRL algorithms \citep{cai2024fast,cai2023last,cai2025average,ito2025instancedependentregretboundslearning,fiegel2026optimallastiterateconvergencematrix}. In fact, the last-iterate analysis itself plays a crucial role in online optimization, 
as it captures the learner’s actual decisions and describes the evolution of the learning process over time. 
For example, a rich line of work has been devoted to the last-iterate behavior of traditional stochastic gradient descent (SGD) \citep{zhang2004solving,shamir2013stochastic,jain2019making}.

The study of last-iterate behavior is closely related to sampling simple regret, which is an important measure for bandit algorithms in the pure exploration setting, where the cost of exploration is largely ignored and the focus is on identifying the best arm as quickly as possible. This setting arises in many practical applications, such as clinical drug testing, and has received considerable attention \citep{even2002pac,mannor2004sample,even2006action}. Therefore, establishing a pointwise guarantee for the native FTRL sampling distribution provides insight beyond the BOBW guarantee into whether the same algorithm can support both regret minimization and pure-exploration objectives.

As mentioned earlier, the $1/2$-Tsallis-INF algorithm achieves logarithmic regret in stochastic bandits. The cumulative pseudo-regret through horizon $T$ is the sum of the sampling simple regrets over the first $T$ rounds, and summing a $t^{-1}$ rate yields a logarithmic dependence on $T$. Although logarithmic cumulative regret does not by itself control a prescribed iterate, it suggests the following natural question:
\begin{quote}
\textit{Does the sampling simple regret of $1/2$-Tsallis-INF decay at rate $\mathcal{O}(t^{-1})$?}
\end{quote}

Beyond sampling simple regret, we also consider last-iterate convergence in the natural geometry of FTRL. For FTRL and mirror descent, the Bregman divergence is the canonical last-iterate quantity: in Euclidean gradient descent, the natural potential is the squared Euclidean distance, while in the non-Euclidean setting it is the Bregman divergence induced by the regularizer. Accordingly, under a unique optimal arm, we study the expected Bregman divergence induced by $\Psi$ between the point mass on the optimal arm and the sampling distribution $p_t$. Since the optimal arm may vary drastically over time in the adversarial setting, we henceforth focus on the stochastic setting.

\subsection{Contribution}
In this work, we prove that, for every suboptimal arm, the expected probability assigned to that arm by the unmodified $1/2$-Tsallis-INF algorithm decays at rate $\mathcal{O}(t^{-1})$. Consequently, its sampling simple regret satisfies $\operatorname{Reg}^{\mathrm{sim}}_t=\mathcal{O}(t^{-1})$. These bounds do not require the optimal arm to be unique. Under a unique optimal arm, we establish a lower bound matching the $t^{-1}$ exponent. To our knowledge, this is the first $\mathcal{O}(t^{-1})$ sampling simple-regret guarantee for the native iterate of an unmodified BOBW FTRL algorithm in stochastic bandits. Importantly, our analysis leaves $1/2$-Tsallis-INF unchanged and fully preserves its BOBW property.

As a complementary result in the geometry induced by the FTRL regularizer, under a unique optimal arm we prove that $\E[D_{\Psi}(e_{i_*},p_t)]$ decays at rate $\mathcal{O}(t^{-1/2})$ and establish a lower bound matching this exponent of $t$.

For the sampling simple-regret bound, we develop a Lyapunov analysis that handles the moving empirical minimum over the optimal arms, uses stopped increments and an adaptive skeleton with state-dependent block lengths to control the larger importance-weighted fluctuations, and retains a strict local contraction to obtain a uniform moment bound. For the Bregman-divergence bound, we introduce an FTRL decomposition tailored to the last iterate and combine it with a refined self-bounding argument and a fixed-arm second-moment estimate.
These techniques may also be applicable to the analysis of $1/2$-Tsallis-INF in other BOBW regimes, as well as to more general Tsallis-entropy FTRL algorithms under importance-weighted bandit feedback.

\subsection{Related Works}
\paragraph{Best-of-Both-Worlds (BOBW)}
In many practical scenarios, it is unclear whether the environment is stochastic or adversarial, 
motivating the design of algorithms with regret guarantees in both settings, which has led to the development of 
BOBW algorithms that aim for optimal performance in both regimes. In bandit problems, the first such algorithm was proposed by \citet{pmlr-v23-bubeck12b}, 
and later the well-known Tsallis-INF algorithm considered in this paper was introduced by \citet{pmlr-v89-zimmert19a}. Subsequent research has extensively explored BOBW algorithms in various online learning settings. For instance linear bandits \citep{ito2023bestofthreeworldslinearbanditalgorithm,kong2023bestofthreeworldsanalysislinearbandits,zhao2025heavytailedlinearbanditsadversarial}, combinatorial semi-bandits \citep{wei2018adaptivealgorithmsadversarialbandits,zimmert2019beating,zhan2025followtheperturbedleaderapproachesbestofbothworldsmset}, 
dueling bandits \citep{pmlr-v162-saha22a}, contextual bandits \citep{pmlr-v238-kuroki24a,kato2023best} and episodic Markov decision processes \citep{jin2021bestworldsstochasticadversarial,ito2025adaptingstochasticadversariallosses}.

\paragraph{Pure Exploration}
Pure exploration problems have been extensively studied under various theoretical frameworks in recent years. 
Among them, the line of work most closely related to last-iterate analysis is simple regret analysis, 
where \citet{lattimore2016causal} derived minimax regret bounds, \citet{wuthrich2021regret} established regret bounds for variants of the expected improvement 
and UCB algorithms, and \citet{zhao2023revisiting} provided bounds for the sequential halving algorithm. 
Another line of research focuses on PAC-based analyses, which aim to identify the optimal arm 
(Best-Arm Identification) within a limited budget \citep{even2002pac,mannor2004sample,kalyanakrishnan2012pac}. 
More recently, several studies have investigated the Pareto frontier between the two optimization objectives 
of regret minimization and pure exploration \citep{zhong2021achieving}.

\paragraph{Organization}
The remainder of this paper is organized as follows. Section~\ref{Preliminaries} gives preliminaries, and Section~\ref{main result} presents our main results. Section~\ref{sec:proof-sketch} develops the proof strategy for the sampling simple-regret bound, while Section~\ref{finalpf} proves the Bregman-divergence upper bound. Concluding remarks are given in Section~\ref{cr}, and numerical results are deferred to Appendix~\ref{Experiments}.

\section{Preliminaries}
\label{Preliminaries}
In this section, we formulate the problem and introduce the FTRL policy. 

\subsection{The Problem Setting and Notation}
We consider the multi-armed bandit problem with arm set  
$
\mathcal{I}=\{1, \ldots, d\}
$.  
At each round \( t = 1, 2, \ldots, \) the learner chooses an arm \( I_t \in \mathcal{I} \), while the environment generates a loss vector \( \ell_t \in [0,1]^d \). The learner observes and incurs $\ell_{t,I_t}$, i.e., the loss for the chosen arm. We consider the stochastic setting and these loss vectors are i.i.d. samples from a fixed but unknown distribution \( \mathcal{D} \). Let \( \mu_i := \mathbb{E}_{\ell \sim \mathcal{D}}[\ell_i] \) denote the expected loss of the arm \( i \), and set
\[
\mu_*:=\min_{i\in\mathcal I}\mu_i,
\qquad
\mathcal O:=\{i\in\mathcal I:\mu_i=\mu_*\},
\qquad
\mathcal S:=\mathcal I\setminus\mathcal O.
\]
We allow the optimal arm to be non-unique. Choose and fix any
$i_*\in\mathcal O$; statements that require a unique optimal arm will
explicitly assume that $i_*$ is unique, equivalently,
$\mathcal O=\{i_*\}$. Define the suboptimality gap of arm $i$ as
$\Delta_i:=\mu_i-\mu_*$ and, when $\mathcal S\ne\varnothing$, define the
minimum positive gap as $\Delta:=\min_{i\in\mathcal S}\Delta_i$.
If $\mathcal S=\varnothing$, the simple regret defined below is identically
zero. We therefore restrict attention to the
nontrivial case $\mathcal S\ne\varnothing$. We also consider the sampling simple regret
\begin{equation}
\label{simpleregret}
\operatorname{Reg}^{\mathrm{sim}}_t
:=\mathbb{E}[\Delta_{I_t}]
=\sum_{i\in\mathcal S}\Delta_i\E[p_{t,i}].
\end{equation}
In the bandit setting, we do not observe the complete loss vector $\ell_t$. Instead, an unbiased estimator $\hat{\ell}_t$ 
of $\ell_t$ is used for updating the cumulative losses, 
$
\hat{L}_t = \sum_{s=1}^{t-1} \hat{\ell}_s
$. The common way of constructing unbiased loss estimators is through importance-weighted
sampling:
\begin{equation}
\label{eq:loss-vector-est}
\hat{\ell}_{t,i}=\frac{\ell_{t,i}A_{t,i}}{p_{t,i}},\,\text{where $A_{t,i}=\1_{\{I_t=i\}}$} .
\end{equation}

\textbf{Notation}\quad
Throughout, $C>0$ denotes a generic universal constant, whereas $C_\alpha>0$ denotes a generic constant depending only on $\alpha$. Their values may change from line to line, and neither hides dependence on any other problem parameter. We use $C'$ only when it must be distinguished from an immediately preceding $C$. In the following, we denote by $\Delta^{d-1}$ the probability simplex. For any $\lambda =(\lambda_1, \ldots, \lambda_d)^\top\in\R^d$, let $\underline{\lambda}=\lambda-\min\limits_{i=1,\cdots,d}\lambda_i\mathbf{1},$ where $\mathbf{1}:=(1, \ldots, 1)^{\top}$, and in this paper, we will primarily use this notation for $\hat{L}_t$, i.e., $\underline{\hat{L}}_{t}=\hat{L}_t-\min _{i\in\mathcal{I}} \hat{L}_{t,i}\mathbf{1}$
. Given a convex function $\Psi$ on $\R^d$, we define the Bregman divergence induced by $\Psi$ as
\[
D_{\Psi}(x,y):=\Psi(x)-\Psi(y)-\agp{x-y,\nabla\Psi(y)}, \; \forall x, y \in \R^d.
\]
For any $\lambda\in\R^d$, we define
\begin{equation}\label{defiphi}
    \phi(\lambda):=\argmax_{p\in\Delta^{d-1}}\agp{p,\lambda}-\Psi(p).
\end{equation}
There are some important properties about $\phi$ in Appendix \ref{important}. We denote by $\mathscr{F}_t$ the filtration $\sigma(I_1,\ell_{1,I_1}, \ldots, I_t,\ell_{t,I_t})$ and define $\E_t[\cdot]:=\E[\cdot\mid \mathscr{F}_{t}]$. For any \( i \in   \mathcal{I}  \), let \( e_i \) denote the \( i \)-th coordinate vector in \( \mathbb{R}^d \). For any $a,b\in\R$, we denote $a^+=\max(a,0)$, $a\wedge b=\min(a,b)$ and $a\vee b=\max(a,b)$.

For later use, define
\begin{equation}
\label{eq:gap-process}
\begin{aligned}
M_t&:=\min_{j\in\mathcal I}\hat L_{t,j},
&M_t^{\mathcal O}&:=\min_{a\in\mathcal O}\hat L_{t,a},\\
N_t^{\mathcal S}&:=\min_{i\in\mathcal S}\hat L_{t,i},
&D_t&:=M_t^{\mathcal O}-N_t^{\mathcal S}.
\end{aligned}
\end{equation}
Here, $M_t$ is the minimum cumulative estimated loss over all arms, while
$M_t^{\mathcal O}$ and $N_t^{\mathcal S}$ are the minimum cumulative
estimated losses over the optimal and suboptimal arms, respectively. Thus,
$D_t$ compares the smallest cumulative estimated loss within $\mathcal O$
with that within $\mathcal S$. In particular,
$M_t=\min\{M_t^{\mathcal O},N_t^{\mathcal S}\}$; when $D_t<0$, the global
minimum is attained by an optimal arm, whereas when $D_t>0$, it is attained
by a suboptimal arm; $D_t=0$ corresponds to a tie between the two sets.
Using fixed tie-breaking rules, define predictably
\[
i_t^*\in\argmin_{i\in\mathcal I}\hat L_{t,i},
\qquad
i_t^{\mathcal O}\in\argmin_{a\in\mathcal O}\hat L_{t,a}.
\]
Thus $i_t^*$ minimizes over all arms,
whereas $i_t^{\mathcal O}$ minimizes only over $\mathcal O$. Moreover,
$p_t$, $\hat L_t$, $i_t^*$, and $i_t^{\mathcal O}$ are
$\mathscr F_{t-1}$-measurable.

\begin{algorithm}[htbp]
\caption{$1/2$-Tsallis-INF}
\label{alg: FTRL}
\begin{algorithmic}
   \STATE {\bfseries Input:} $0<\alpha < 1$
   \STATE {\bfseries Initialize:} $\hat{L}_1 = \mathbf{0}, \eta_t = \alpha/\sqrt{t}$
   \FOR{$t=1, 2, \dots$}
   \STATE choose $p_t=\argmin\limits_{p\in\Delta^{d-1}} \; \frac{\alpha}{\sqrt{t}}\agp{p,\hat{L}_t}-4\sum_{i=1}^d \sqrt{p_i}$\;
   \STATE play $I_t\sim p_t$, and
   observe $\ell_{t,I_t}$\;
   \STATE construct estimator $\hat{\ell}_{t,i}=\frac{\ell_{t,i}\1_{\{I_t=i\}}}{p_{t,i}}$,
   and update $\hat L_{t+1} = \hat L_{t}+\hat\ell_t$\;
   \ENDFOR
\end{algorithmic}
\end{algorithm}

\subsection{FTRL Policy}
We study the Follow-The-Regularized-Leader (FTRL) algorithm. At each time step, the algorithm chooses a probability
distribution $p_t$ over $d$ arms:
\begin{equation}
\label{ptdefi}
    p_t=\argmin_{p\in\Delta^{d-1}}\eta_t\agp{p,\hat{L}_t}+\Psi(p),
\end{equation}
where $\eta_t$ is the learning rate and $\Psi(p)$ is a convex regularization function. One can also see that $p_t=\phi(-\eta_t\hat{L}_t)$. 

In simple settings, $\Psi(p)$ is typically chosen so that its Hessian matrix satisfies
$\nabla^2 \Psi(p)=\operatorname{diag}\{p_1^\xi,\ldots,p_d^\xi\}$,
where $\xi$ is a fixed constant. Common examples include the negentropy ($\Psi(p)=\sum_{i=1}^d p_i\log p_i$, corresponding to the EXP3 algorithm), the log-barrier ($\Psi(p)=-\sum_{i=1}^d\log p_i$), and the $\theta$-Tsallis entropy ($\Psi(p)=-\sum_{i=1}^d p_i^\theta$, $0<\theta<1$). \citet{pmlr-v89-zimmert19a} showed that the $1/2$-Tsallis-INF algorithm (Algorithm~\ref{alg: FTRL}), that is, FTRL with $\Psi(p)=-4\sum_{i=1}^d \sqrt{p_i}$, achieves the BOBW guarantee, and provided intuitive arguments indicating that other regularizers cannot. Also, the BOBW property enjoyed by 
\(1/2\)-Tsallis-INF and its variants is fairly general. It plays an important role not only in the MAB setting considered in this paper, but also in problems such as linear bandits \citep{zhao2025heavytailedlinearbanditsadversarial}, semi-bandits \citep{zimmert2019beating}, and MDPs \citep{chen2026bestofbothworldsheavytailedmarkovdecision}.

In this paper, we focus on $1/2$-Tsallis-INF. Then by Lemma \ref{pti-2}, the solution of \eq{ptdefi} takes the form
\begin{equation}
\label{pti}
    p_{t,i}=4\left(\eta_t\underline{\hat{L}}_{t, i}+2p_{t,i_{t}^*}^{-\frac{1}{2}}\right)^{-2}.
\end{equation}

\section{Main Results}
\label{main result}
\label{sec:main-results}

In this section, we first present pointwise upper bounds for the
sampling distribution and simple regret, then give the Bregman-divergence
upper bound, and finally state the corresponding lower bounds.

\subsection{\texorpdfstring{Sampling-Distribution and Simple-Regret Upper Bounds}{Sampling-Distribution and Simple-Regret Upper Bounds}}

We first state the pointwise last-iterate guarantees for the native FTRL
sampling distribution.  The result does not require a unique optimal arm and
is proved through a uniform Lyapunov moment bound stated below.

\begin{Thm}[Multiple optimal arms]
\label{thm:main}
Suppose that $0<\alpha<1$ and $\mathcal S\ne\varnothing$.  There exists a
constant $C_\alpha>0$ such that, for every $t\ge1$ and every
$i\in\mathcal S$,
\begin{equation}
\label{eq:main-armwise-mass}
\E[p_{t,i}]
\le
\frac{C_\alpha d^4\Delta^{-8}\log^4(2d)}{t}.
\end{equation}
Consequently,
\begin{equation}
\label{eq:main-simple-regret}
\operatorname{Reg}^{\mathrm{sim}}_t
\le
\frac{C_\alpha d^5\Delta^{-8}\log^4(2d)}{t}.
\end{equation}
\end{Thm}

Theorem~\ref{thm:main} gives an $\mathcal{O}(t^{-1})$ pointwise guarantee for the
sampling distribution produced by the unmodified FTRL update, without
averaging, additional exploration, or a separate recommendation rule.  To
the best of our knowledge, this is the first $\mathcal{O}(t^{-1})$ sampling
simple-regret guarantee for the native iterate of an unmodified
best-of-both-worlds FTRL algorithm in stochastic bandits.
The lower bound in Theorem~\ref{thm:lower} shows that the
$t^{-1}$ time dependence is tight, and
summing this rate over time is consistent with logarithmic cumulative
pseudo-regret.

For comparison, logarithmic-regret analyses of FTRL under multiple optimal
arms were developed by \citet{pmlr-v134-ito21a} for $1/2$-Tsallis-INF and by
\citet{jin2023improved} for a broader family of regularizers.  Such cumulative
guarantees do not by themselves control a prescribed iterate.  Their proofs
rely on involved regret-decomposition analyses, whereas our Lyapunov proof
provides an alternative route, although our resulting regret bound has worse
dependence on the dimension and the gaps.

The pointwise problem is delicate even in the stochastic setting.  At the
target scale $p_{t,i}\asymp t^{-1}$, an importance-weighted loss estimate has
one-step second moment of order $t$, so its cumulative fluctuation is on the
same linear scale as the mean gap.  Standard concentration therefore does not
by itself yield a decaying fixed-time bound for the empirical minima.  With multiple optimal arms,
there is an additional obstruction: the optimal side of the gap process is
the moving minimum $M_t^{\mathcal O}$, and the identity of its minimizing arm
may change over time.

Following the Lyapunov framework of \citet{zhan2026does}, we reduce the
sampling-mass problem to constructing a nonnegative family satisfying
$p_{t,i}\le g_t(D_t)/t$ for every $i\in\mathcal S$ and proving a uniform
bound on its expectation.  We use
\begin{equation}
\label{eq:potential}
g_t(y):=
\begin{cases}
\displaystyle\frac{4t^2}{(2\sqrt t-\alpha y)^2}, & y<0,\\[7pt]
\displaystyle\frac{(y+\Delta t)^2}{\Delta^2t}, & y\ge0.
\end{cases}
\end{equation}
If $D_t\ge0$, then \eq{eq:potential} gives
$g_t(D_t)=(D_t+\Delta t)^2/(\Delta^2t)\ge t$, and hence
$p_{t,i}\le1\le g_t(D_t)/t$.  If $D_t<0$, then $M_t=M_t^{\mathcal O}$ and
$\underline{\hat L}_{t,i}\ge-D_t$ for every $i\in\mathcal S$.  Hence
\eq{pti} gives
\begin{equation}
\label{eq:negative-potential-dominates}
p_{t,i}
\le
4\left(2+\frac{\alpha(-D_t)}{\sqrt t}\right)^{-2}
=\frac{g_t(D_t)}{t},
\qquad i\in\mathcal S.
\end{equation}
Thus $p_{t,i}\le g_t(D_t)/t$ on both half-lines.

The two branches meet at $g_t(0)=t$.  Up to the normalization by
$\Delta^{-2}$, the positive branch is the $q=1$ member of the quadratic
family $t^{q-2}(y+\Delta t)^2$ constructed by \citet{zhan2026does}.  The
inverse-square negative branch is instead dictated by the exact
sampling-probability identity in \eq{pti} and also has value $t$ at the interface;
matching the interface within this normalized quadratic family therefore
selects $q=1$.

The negative branch is also consistent with the diffusion heuristic of
\citet{zhan2026does}.  On the negative half-line, their toy calculation only
requires the generator drift of the candidate Lyapunov function to be
nonpositive, and the exponential profile used there is one convenient
separated choice.  At sufficiently large times, the inverse-square profile
has the required nonpositive toy-generator drift on a sufficiently narrow
linear strip, although not on the entire negative half-line.  The formal proof
does not rely on this heuristic and establishes the required contraction
directly in discrete time.

The main probabilistic input is the following all-time moment estimate.  Its
proof is sketched in Section~\ref{sec:proof-sketch} and given in full in
Appendix~\ref{sec:global-proofs}.

\begin{Thm}[Uniform Lyapunov moment]
\label{thm:coarse-uniform-moment}
Suppose that $0<\alpha<1$ and $\mathcal S\ne\varnothing$.  There exists a
constant $C_\alpha>0$, depending only on $\alpha$, such that every $t\ge1$
satisfies
\begin{equation}
\label{eq:coarse-uniform-moment}
\E[g_t(D_t)]
\le
C_\alpha d^4\Delta^{-8}\log^4(2d).
\end{equation}
\end{Thm}

By \eq{eq:coarse-uniform-moment}, Theorem~\ref{thm:main} is a direct
consequence of Theorem~\ref{thm:coarse-uniform-moment}.

Our proof is not a direct application of the analysis of
\citet{zhan2026does}.  First, we overcome the difficulty caused by multiple
optimal arms through a careful analysis of the moving minimum
$M_t^{\mathcal O}$.  Second, we study the behavior of $D_t$ on a deeper
region where the importance-weighted fluctuations are substantially larger.
Specifically, we introduce a stopped-increment argument to sharpen the
control of $N_t^{\mathcal S}$, extending the corresponding estimates of
\citet{zhan2026does} to the wider strip, and use an adaptive skeleton to
accommodate the resulting state-dependent block lengths.  Finally, we
strengthen the local supermartingale inequality by retaining a strict
contraction margin, which is needed to control the larger entrance budget.
These changes are explained in detail in
Section~\ref{subsec:structural-changes}.

The moving-minimum comparison may also provide a route for replacing the
fixed-optimal-arm step in the empirical-recommendation argument of
\citet{zhan2026does}.  A complete version of that separate extension would
also require their exponential negative branch and fixed-grid global
analysis, which we do not pursue here.

The construction of \eq{eq:potential}, the proof strategy, and the
comparison with \citet{zhan2026does} are developed in
Section~\ref{sec:proof-sketch}; all formal local estimates and their proofs
are given in the appendices.

\subsection{\texorpdfstring{Bregman-Divergence Upper Bound}{Bregman-Divergence Upper Bound}}
\label{nd}

We first adapt the standard FTRL regret decomposition to the
last-iterate setting and then state the resulting upper bound for the Bregman
divergence.
Let $\widetilde p_{t+1}=\phi(-\eta_t\hat L_{t+1})$.
\begin{Lem}\label{decompo}
    For any $t\ge 1$, we have
\begin{equation}\label{decom}
    \begin{aligned}
        \agp{p_t-e_{i_*},\hat{\ell}_t}
        \leq{}&\underbrace{\frac{1}{\eta_t}D_{\Psi}(p_t,\widetilde p_{t+1})}_{\mathbf{\uppercase\expandafter{\romannumeral1}}}\\
        &+\underbrace{\frac{1}{\eta_t}D_{\Psi}(e_{i_*},p_t)
        -\left(\frac{2}{\eta_t}-\frac{1}{\eta_{t+1}}\right)D_{\Psi}(e_{i_*},p_{t+1})}_{\mathbf{\uppercase\expandafter{\romannumeral2}}}.
    \end{aligned}
\end{equation}
\end{Lem}
\noindent We leave the proof in Appendix \ref{proofdecompo}.
Following the convention, we refer to the first and second term of \eq{decom} as the stability term and penalty
term. By Lemma~\ref{stabequi}, the first term is exactly the one-step FTRL quantity controlled in Lemma~11(2) of \citet{pmlr-v89-zimmert19a}, which is why we call it the stability term. The second term combines the usual penalty contribution with the contribution caused by the change in the learning rate.

Our convergence result cannot be obtained from this decomposition in a straightforward manner. A Bregman divergence appears on the right-hand side of the inequality, while no such term is present on the left-hand side. As in the classical self-bounding technique, we therefore need to establish a lower bound for the left-hand side that is linked to the Bregman divergence. Through a delicate analysis, we find that achieving this goal requires controlling the second moment of $\underline{\hat{L}}_{t,i_*}$, namely the gap between the estimated losses of the optimal arm and the empirically optimal arm. This step is technically challenging, and we elaborate it in Appendix~\ref{sec:fixed-arm-moments}.

Our next result characterizes the decay rate of the Bregman divergence between
the probability distribution \(p_t\) obtained at iteration \(t\) and the point mass \(e_{i_*}\)
corresponding to the optimal arm.
\begin{Thm}\label{rate}
    Assume that $i_*$ is unique. For Algorithm \ref{alg: FTRL}, if $0<\alpha<1$, then for any $t\ge 1$, we have
    {
    \[
    \E[D_{\Psi}(e_{i_*},p_t)]\leq C_{\alpha,\mu_{i_*}}\frac{d}{\Delta}t^{-1/2},
    \]
    where $C_{\alpha,\mu_{i_*}}$ can be chosen as $C e^{\alpha^2\mu_{i_*}}/[\alpha(1-\alpha^2\mu_{i_*})]$ for a universal constant $C>0$.}
\end{Thm}
The proof is in Section \ref{finalpf}.
In the one-player specialization corresponding to stochastic bandits,
Proposition~1 of
\citet{ito2025instancedependentregretboundslearning} yields
$\E[D_{\Psi}(e_{i_*},p_t)]
=\mathcal{O}\!\left(\frac{\log t}{\sqrt{t}}\right)$.
Theorem~\ref{rate} removes this extra logarithmic factor.
The matching lower bound in Theorem~\ref{thm:lower} shows
that the $t^{-1/2}$ time dependence is tight.

The $t^{-1}$ simple-regret bound in
Theorem~\ref{thm:main} and the $t^{-1/2}$ Bregman-divergence bound above are
complementary: neither directly recovers the other with its stated dependence
on $t$ and the problem parameters.  Thus neither guarantee is stronger than
the other at the level established here.

\subsection{\texorpdfstring{Lower Bounds}{Lower Bounds}}

We conclude with lower bounds for both the Bregman divergence
and the sampling simple regret.  The following identity will be used in the
Bregman-divergence argument.
\begin{equation}\label{2sqrt}
    \begin{aligned}
        D_{\Psi}(e_{i_*}, p_t)
=&2\sum_{i\ne i_*}^d \sqrt{p_{t,i}} +\frac{2(1-\sqrt{p_{t,i_*}})^2}{\sqrt{p_{t,i_*}}}
\ge& 2\sum_{i \ne i_*} \sqrt{p_{t,i}}\ge 2p_{t,i}.
    \end{aligned}
\end{equation}

\begin{Thm}\label{thm:lower}
    Assume that $i_*$ is unique. For Algorithm \ref{alg: FTRL}, if $0<\alpha<1$, then there exists $C_{\alpha}>0$ depending only on $\alpha$ such that for any $t\ge C_{\alpha}\frac{d}{\Delta^2}$,
    {
    \[
    \E[D_{\Psi}(e_{i_*},p_t)]\ge \frac{2}{\alpha}\sum_{i\ne i_*}\frac{1}{\Delta_i}t^{-1/2}
    \ge \frac{2}{\alpha\Delta}t^{-1/2}.
    \]}
    {
    Moreover,
    \[
    \operatorname{Reg}^{\mathrm{sim}}_t
    \ge \frac{1}{\alpha^2}\sum_{i\ne i_*}\frac{1}{\Delta_i}t^{-1}
    \ge \frac{1}{\alpha^2\Delta}t^{-1}.
    \]}
\end{Thm}
We leave the proof in Appendix \ref{sec:pflower}.

\section{\texorpdfstring{Proof Strategy for the Simple Regret Bound}{Proof Strategy for the Simple Regret Bound}}
\label{sec:proof-sketch}

This section separates the assembly of the proof from the reasons that its
ingredients differ from those of \citet{zhan2026does}.  We first follow the
adaptive-skeleton argument that proves the uniform Lyapunov moment, and then
explain the four structural changes behind that argument.

\subsection{Proof sketch}
\label{subsec:proof-assembly}

For every $i\in\mathcal S$, the pathwise inequality
$p_{t,i}\le g_t(D_t)/t$ from
Section~\ref{sec:main-results} reduces the problem to controlling
$\E[g_t(D_t)]$ uniformly in time.  Following the organization at the
beginning of Appendix~\ref{sec:global-proofs}, we first control the last
adaptive-skeleton node before $t$, bound the weighted reentry contributions
accumulated up to $t$, and only then cross the final incomplete block.

Appendix~\ref{sec:parameters} fixes the block scale $A$, boundary coefficient
$\beta$, and burn-in time
$N$.  We call $\{D_u\ge-\beta u\}$ the controlled
strip.  For each $u\ge N$, define the predictable horizon
\begin{equation}
\label{eq:sketch-state-dependent-horizon}
h_u:=
\begin{cases}
\displaystyle
\left\lceil
A\left(\sqrt d+\frac{\alpha(-D_u)^+}{2\sqrt u}\right)^2
\right\rceil,
&D_u\ge-\beta u,\\[2mm]
1,&D_u<-\beta u.
\end{cases}
\end{equation}
Starting from $N$, define the skeleton by
\begin{equation}
\label{eq:sketch-skeleton}
s_1:=N,
\qquad
s_{j+1}:=s_j+h_{s_j}.
\end{equation}
The effective-sampling origin of $h_u$ and the need for the adaptive
skeleton are explained in
Section~\ref{subsec:state-dependent-blocks}.

For every $u\ge N$ satisfying $D_u\ge-\beta u$,
Lemma~\ref{lem:one-block-contraction} supplies
\begin{equation}
\label{eq:sketch-local-contraction}
\E\left[
g_{u+h_u}(D_{u+h_u})
\,\middle|\,\mathscr F_{u-1}
\right]
\le
\left(1-\frac{h_u}{C_\alpha u}\right)g_u(D_u),
\qquad u\ge N,\quad D_u\ge-\beta u.
\end{equation}
The proof controls the two contributions in
\[
D_{u+h_u}-D_u+\Delta h_u
=
\bigl(M_{u+h_u}^{\mathcal O}-M_u^{\mathcal O}-\mu_*h_u\bigr)
+
\bigl((\mu_*+\Delta)h_u
-(N_{u+h_u}^{\mathcal S}-N_u^{\mathcal S})\bigr).
\]
When $D_u\ge0$, Corollary~\ref{cor:positive-optimal-increment} supplies the
conditional moments of the first parenthesis; when $D_u<0$,
Lemma~\ref{lem:stopped-optimal-set-increment} supplies its stopped
positive-part moments.  Corollary~\ref{cor:state-dependent-horizon} controls
the positive part of the second parenthesis.  The choice of the strip, the
moving optimal-set minimum, and the state-dependent horizon are explained in
Sections~\ref{subsec:controlled-strip},
\ref{subsec:moving-optimal-minimum}, and
\ref{subsec:state-dependent-blocks}, respectively.

Fix the constant $C_\alpha$ in
\eq{eq:sketch-local-contraction}.  Choose
$\gamma\in(0,1]$, depending only on $\alpha$, so that
$4\gamma\le C_\alpha^{-1}$, and put
$\mathscr G_j:=\mathscr F_{s_j-1}$ and
$H_j:=s_j^\gamma g_{s_j}(D_{s_j})$.
Appendix~\ref{sec:parameter-verification} gives $h_{s_j}/s_j\le1/8$, so
\eq{eq:sketch-local-contraction} gives, on
$\{D_{s_j}\ge-\beta s_j\}$,
\begin{equation}
\label{eq:sketch-weighted-supermartingale}
\E[H_{j+1}\mid\mathscr G_j]
\le
\left(1+\frac{h_{s_j}}{s_j}\right)^\gamma
\left(1-4\gamma\frac{h_{s_j}}{s_j}\right)H_j
\le H_j.
\end{equation}
For a fixed $t\ge N$, using
\eq{eq:sketch-weighted-supermartingale} and applying
Lemma~\ref{lem:finite-horizon-stopping} to $H_j$ on successive stretches of
skeleton nodes satisfying $D_{s_j}\ge-\beta s_j$, we first control the last
skeleton node before $t$, namely $s_{J_t}$ with
$J_t:=\max\{j:s_j\le t\}$, as follows:
\begin{align}
\label{eq:sketch-skeleton-bound}
&\E\left[
g_{s_{J_t}}(D_{s_{J_t}})
\1_{\{D_{s_{J_t}}\ge-\beta s_{J_t}\}}
\right]\nonumber\\
&\quad\le
C_\alpha\left(\frac Nt\right)^\gamma
\E[g_N(D_N)]
+C_\alpha\E\left[
\sum_{k:N<S_k\le t}
\left(\frac{S_k}{t}\right)^\gamma
g_{S_k}(D_{S_k})
\right].
\end{align}
Here $S_1<S_2<\cdots$ denote the calendar-time reentries into the controlled
strip, namely the times $u\ge N$ at which
$D_{u-1}<-\beta(u-1)$ and $D_u\ge-\beta u$.  A skeleton reentry is a
transition from $D_{s_j}<-\beta s_j$ to
$D_{s_{j+1}}\ge-\beta s_{j+1}$; since
\eq{eq:sketch-state-dependent-horizon} then gives $s_{j+1}=s_j+1$, every
skeleton reentry is one of the times $S_k$.  Enlarging the resulting
nonnegative reentry sum to all $S_k$ and comparing $s_{J_t}$ with $t$ gives
\eq{eq:sketch-skeleton-bound}, which is
Lemma~\ref{lem:entrance-decomposition}.

For the initial term, choose any $i_*\in\mathcal O$.  The inequality
$D_N^+\le\underline{\hat L}_{N,i_*}$,
Lemmas~\ref{lem:fixed-arm-maximal} and
\ref{lem:lyapunov-branch-comparison}(3), and the parameter choices in
Appendix~\ref{sec:parameters} give $\E[g_N(D_N)]\le C_\alpha N$.
The second term in \eq{eq:sketch-skeleton-bound} is controlled by
Lemma~\ref{lem:entrance-budget}.  Since $(N/t)^\gamma\le1$, its right-hand side
is therefore at most
$C_\alpha\{N+\beta^{-2}(1+|\mathcal S|\beta^{-2})\}$.
The role of $\gamma>0$ in the entrance discount $(S_k/t)^\gamma$ is explained
in Section~\ref{subsec:weighted-entrances}.

Lemma~\ref{lem:one-block-maximal} states that, for every $u\ge N$ with
$D_u\ge-\beta u$,
\[
\E\left[
\max_{u\le v\le u+h_u}g_v(D_v)
\,\middle|\,\mathscr F_{u-1}
\right]
\le C_\alpha g_u(D_u).
\]
Finally, on $\{D_t\ge-\beta t\}$,
Lemma~\ref{lem:one-block-maximal} transfers the skeleton-node bound in
\eq{eq:sketch-skeleton-bound} from $s_{J_t}$ to $t$.  On
$\{D_t<-\beta t\}$, the negative branch is bounded directly by
\begin{equation}
\label{eq:sketch-below-strip}
g_t(D_t)\le g_t(-\beta t)
\le\frac{4}{\alpha^2\beta^2}.
\end{equation}
Together, \eq{eq:sketch-skeleton-bound},
Lemma~\ref{lem:one-block-maximal}, and \eq{eq:sketch-below-strip} prove
Theorem~\ref{thm:uniform-moment}.  The parameter
simplification in Appendix~\ref{sec:parameter-verification} and the
pre-burn-in fixed-arm bound give
Theorem~\ref{thm:coarse-uniform-moment}, and hence
Theorem~\ref{thm:main}.

\subsection{Structural changes in the Lyapunov argument}
\label{subsec:structural-changes}

The proof sketch above records what is used.  We now explain why its four
main ingredients differ from the unique-optimum analysis of
\citet{zhan2026does}.

\subsubsection{The scale of the controlled strip}
\label{sec:lyapunov-overview}
\label{subsec:controlled-strip}

This subsection explains why the present inverse-square argument uses the
linear controlled strip $D_t\ge-\beta t$, rather than the square-root strip
$D_t\ge-\sqrt{dt}/\alpha$ used by \citet{zhan2026does}.  The boundary must be
deep enough to keep the Lyapunov cost of reentries uniformly bounded, but not
so deep that the local one-block estimate loses its supermartingale drift.

We first explain why the strip cannot be too shallow.  For the entrance
budget used here to remain uniformly summable, the value of the Lyapunov
function at the boundary must at least be uniformly bounded in time.  The
empirical-recommendation analysis of \citet{zhan2026does} uses an exponential
negative branch.  On $\{D_t<-\sqrt{dt}/\alpha\}$, monotonicity gives the bound
$\Delta^2t^q\exp(-\lambda\sqrt{dt}/\alpha)$ for a fixed $\lambda>0$, which is
uniform in $t$.  However, for our Lyapunov function $g_t$ defined in
\eq{eq:potential}, the value at the same square-root boundary is
$g_t(-\sqrt{dt}/\alpha)=4t/(2+\sqrt d)^2$, which grows with $t$.  More
generally, at a cutoff of depth $b_t$ the boundary value of our $g_t$ is
$4t^2/(2\sqrt t+\alpha b_t)^2$, so a sublinear depth does not give a
time-uniform bound.

We next explain why the controlled strip cannot be made wider by moving its
lower boundary substantially deeper.  If $D_t<0$ and, only for a mean-field
heuristic, the random endpoint over a short block of $h$ rounds is replaced
by $D_t-\Delta h$, then for
$-D_t\gg\sqrt t$,
$h/t\ll1$, and $\Delta h/(-D_t)\ll1$,
\[
\frac{g_{t+h}(D_t-\Delta h)}{g_t(D_t)}
\approx
\left(\frac{1+h/t}{1-\Delta h/D_t}\right)^2
\approx
1+2h\left(\frac1t+\frac{\Delta}{D_t}\right).
\]
Thus, at the heuristic level, if $-D_t$ grows superlinearly in $t$, the
explicit time growth of our $g_t$ dominates the nominal gap drift.  Our $g_t$
then loses the supermartingale contraction that motivates its use as a
Lyapunov function; correspondingly, with the present $g_t$ and $h_t$, an
estimate of the form stated in \eq{eq:sketch-local-contraction} cannot hold uniformly
on a strip containing such states.

\subsubsection{Controlling the increment of
\texorpdfstring{$M_t^{\mathcal O}$}{the optimal-set minimum}}
\label{subsec:moving-optimal-minimum}

The local contraction estimate requires conditional moment and stopped
maximal bounds for the change of $M_t^{\mathcal O}$ over a predictable block
of $L$ rounds.  Under the unique-optimum assumption of
\citet{zhan2026does}, $M_t^{\mathcal O}$ is the cumulative estimate of one
fixed optimal arm.  With multiple optimal arms, however,
$M_t^{\mathcal O}=\min_{a\in\mathcal O}\hat L_{t,a}$, and the arm attaining
this minimum may change during the block.  Fixing an arm $a\in\mathcal O$
does not resolve this issue, because
\[
\underline{\hat L}_{u,a}
=D_u^++\bigl(\hat L_{u,a}-M_u^{\mathcal O}\bigr).
\]
Once $a$ ceases to attain $M_u^{\mathcal O}$, the
shifted loss $\underline{\hat L}_{u,a}$ entering \eq{pti}
is no longer determined by $D_u$ alone.

We therefore control the change of $M_t^{\mathcal O}$ one round at a time.
Before observing the round-$u$ loss, choose
$i_u^{\mathcal O}\in\argmin_{a\in\mathcal O}\hat L_{u,a}$ using fixed
tie-breaking.  Then $i_u^{\mathcal O}$ is $\mathscr F_{u-1}$-measurable and
\[
\begin{gathered}
\underline{\hat L}_{u,i_u^{\mathcal O}}
=M_u^{\mathcal O}-M_u=D_u^+,\\
0\le M_{u+1}^{\mathcal O}-M_u^{\mathcal O}
\le\hat\ell_{u,i_u^{\mathcal O}},
\qquad
M_{t+L}^{\mathcal O}-M_t^{\mathcal O}
\le\sum_{u=t}^{t+L-1}\hat\ell_{u,i_u^{\mathcal O}}.
\end{gathered}
\]
The first identity provides the shifted loss needed to apply
Lemma~\ref{4-2} and control the inverse sampling probability.  Because every predictably selected arm is
optimal, centering the selected loss increments by $\mu_*$ produces
martingale differences.  Appendix~\ref{sec:optimal-minimum} combines these
comparisons with the monotonicity of $M_u^{\mathcal O}$ to derive the
conditional moment and stopped maximal inputs used to prove
\eq{eq:sketch-local-contraction}.  With these replacements, the ensuing
scalar estimates follow the organization of
\citet[Appendix~F]{zhan2026does}, without a union bound over $\mathcal O$.

\subsubsection{State-dependent blocks for the growth of
\texorpdfstring{$N_t^{\mathcal S}$}{the suboptimal-set minimum}}
\label{subsec:state-dependent-blocks}

The proof of the local contraction in \eq{eq:sketch-local-contraction}
requires $N_{t+h_t}^{\mathcal S}-N_t^{\mathcal S}$ not to fall substantially
below $(\mu_*+\Delta)h_t$, so that the suboptimal-set minimum supplies the gap
drift over the block.  This subsection explains why obtaining this growth
leads to the state-dependent block length $h_t$ in
\eq{eq:sketch-state-dependent-horizon}.  For $t\ge N$ with
$D_t\ge-\beta t$, fix the numerical constant $\varepsilon\in(0,1)$ used in
the local proof.  Corollary~\ref{cor:state-dependent-horizon} provides the
needed positive-shortfall bound:
\[
 \E\left[
  \left(
   (\mu_*+\Delta)h_t
   -(N_{t+h_t}^{\mathcal S}-N_t^{\mathcal S})
  \right)^+
  \middle|\mathscr F_{t-1}
 \right]
 \le \varepsilon\Delta h_t,
\]
together with a corresponding second-moment bound.  In particular,
$\E[N_{t+h_t}^{\mathcal S}-N_t^{\mathcal S}\mid\mathscr F_{t-1}]
\ge(\mu_*+(1-\varepsilon)\Delta)h_t$.  Intuitively, every arm that can attain
the block-end minimum must be sampled often enough for this growth to hold,
so the block length should be at least of the order of its reciprocal
sampling probability.  At the block start, the lower sampling-probability
bound in Lemma~\ref{4-2} gives, for an arm
$i$ attaining $N_t^{\mathcal S}$,
\[
 p_{t,i}^{-1}
 \le\left(\sqrt d+\frac{\alpha(-D_t)^+}{2\sqrt t}\right)^2,
 \qquad
 h_t=\left\lceil
 A\left(\sqrt d+\frac{\alpha(-D_t)^+}{2\sqrt t}\right)^2
 \right\rceil,
\]
where $A$ supplies the concentration budget.  On the square-root strip used
by \citet{zhan2026does}, their short-block argument gives a time-uniform
lower bound of order $1/d$ for the relevant sampling probabilities, so one
deterministic block length suffices.  On the wider strip
$D_t\ge-\beta t$, the reciprocal-probability bound can instead grow with
$(-D_t)^+$, which leads to the state-dependent horizon above and hence to
the adaptive skeleton in \eq{eq:sketch-skeleton}.

Choosing the state-dependent horizon $h_t$ does not by itself prove the
required growth of $N_t^{\mathcal S}$.  On the wider linear strip, the
available reciprocal-probability bound can be of order $t$, so the variance
bound for an untruncated $L$-round increment can deteriorate from
$\mathcal O(dL)$ in \citet{zhan2026does} to $\mathcal O(Lt)$.
Appendix~\ref{sec:suboptimal-minimum} therefore stops each relevant arm's
increment when its estimate reaches $N_t^{\mathcal S}+L$.  This stopped
argument yields the required first- and second-moment shortfall bounds,
extending the corresponding part of
\citet[Lemma~E.3(1)]{zhan2026does} to the wider strip.

\subsubsection{A strengthened supermartingale condition}
\label{subsec:weighted-entrances}

This subsection explains why we need the stronger contraction condition in
\eq{eq:sketch-local-contraction}, whereas \citet{zhan2026does} only needs
the usual supermartingale inequality
$\E[g_{t+L}(D_{t+L})\mid\mathscr F_{t-1}]\le g_t(D_t)$.  The difference lies
in the entrance budget.  In their argument, the Lyapunov value at a reentry
is bounded by $\Delta^2s^q\exp(-\lambda\sqrt{ds}/(2\alpha))$, which is
summable in $s$.  In our argument, every entrance only satisfies
$g_{S_k}(D_{S_k})\le16/(\alpha^2\beta^2)$, a constant-order bound with no
decay in $S_k$.

If we retained only the usual supermartingale condition and applied the
stopping decomposition of \citet[Appendix~G, Lemma~G.2]{zhan2026does}, or its
stopped-time version in Lemma~\ref{lem:finite-horizon-stopping}, the
corresponding entrance budget would be
$\E[\sum_{k:N<S_k\le t}g_{S_k}(D_{S_k})]$.  Intuitively, every reentry means
that a suboptimal arm is selected, so this unweighted entrance budget is on
the scale of cumulative regret, namely $\mathcal O(\log t)$ rather than
$\mathcal O(1)$.  It therefore cannot yield the uniform Lyapunov moment bound
required here.

The strengthened contraction condition in \eq{eq:sketch-local-contraction}
removes this logarithmic loss.  For a sufficiently small fixed $\gamma>0$,
$H_j=s_j^\gamma g_{s_j}(D_{s_j})$ satisfies the supermartingale condition in
\eq{eq:sketch-weighted-supermartingale}.  Applying
Lemma~\ref{lem:finite-horizon-stopping} to $H_j$ gives the weighted entrance
budget
$\E[\sum_{k:N<S_k\le t}(S_k/t)^\gamma g_{S_k}(D_{S_k})]$ in
\eq{eq:sketch-skeleton-bound}.  Lemma~\ref{lem:entrance-budget} reduces its
control to counting the entrance times $S_k$ and proves the uniform bound
$C_\alpha\beta^{-2}(1+|\mathcal S|\beta^{-2})$.  Thus the strengthened
condition turns the regret-scale entrance budget into the $\mathcal O(1)$
bound needed for the Lyapunov moment estimate.

\section{Proof of Theorem \ref{rate}}\label{finalpf}
In this section, we provide the proof of Theorem~\ref{rate}. To apply the self-bounding technique \citep{pmlr-v89-zimmert19a}, we take expectations on both sides of the inequality and rearrange the terms as follows:
\begin{equation}\label{expect}
    \begin{aligned}
        \operatorname{Reg}^{\mathrm{sim}}_t\leq{}&\left(2\E[\mathbf{\uppercase\expandafter{\romannumeral1}]}-\operatorname{Reg}^{\mathrm{sim}}_t\right)+2\E[\mathbf{\uppercase\expandafter{\romannumeral2}]}.
    \end{aligned}
\end{equation}
Our proof is based on a direct term-by-term analysis of \eq{expect}. 

\textbf{Left-hand side}\quad
First, we focus on the left-hand side. 
In the classical self-bounding technique~\citep{zimmert2021tsallis}, one requires a lower bound on the simple regret $\operatorname{Reg}^{\mathrm{sim}}_t$ of the form
$\E[\sum_{i \neq i_*} \Delta_i' p_{t,i}] - C$.
In the stochastic bandit setting, this condition is satisfied with $\Delta_i' = \Delta_i$ and $C = 0$, and the equality holds.
However, for our new decomposition in \eq{decom}, the term \textbf{\uppercase\expandafter{\romannumeral2}} involves a Bregman divergence, which renders the original lower bound inapplicable. This makes it necessary to derive a new lower bound that is more closely tied, in form, to the Bregman divergence.

Intuitively, based on \eq{simpleregret} and \eq{2sqrt}, one may conjecture that the magnitude of $\operatorname{Reg}^{\mathrm{sim}}_t$ is comparable to the second moment of $D_{\Psi}(e_{i_*}, p_t)$. In fact, we have:
\begin{Lem}\label{simlow}
    Assume that $i_*$ is unique. For Algorithm \ref{alg: FTRL}, if $0<\alpha<1$, then
    \[
    \operatorname{Reg}^{\mathrm{sim}}_t\ge \frac{C(1-\alpha^2\mu_{i_*})\Delta}{d\,e^{\alpha^2\mu_{i_*}}} \left(\E[D_{\Psi}(e_{i_*},p_t)]\right)^2,
    \]
    where $C$ is a positive constant.
\end{Lem}
Its proof will be deferred to Appendix \ref{pfsimlow}.

\textbf{Right-hand side}\quad
For the right-hand side, the stability term $\mathbf{\uppercase\expandafter{\romannumeral1}}$ is controlled by the self-bounding technique, while the penalty term $\mathbf{\uppercase\expandafter{\romannumeral2}}$ is given explicitly by \eq{decom}.
For the stability term $\mathbf{\uppercase\expandafter{\romannumeral1}}$, we can directly borrow the result from~\citet{pmlr-v89-zimmert19a} and there exists $C>0$ such that
\begin{equation}\label{31}
    2\E[\mathbf{\uppercase\expandafter{\romannumeral1}]}-\operatorname{Reg}^{\mathrm{sim}}_t\leq \frac{Cd}{t\Delta}.
\end{equation}
First, we need the following result:
\begin{Lem}\label{stab}
    For any $t\ge 1$, there exists $C>0$ such that
    \(
    \E_{t-1}[\mathbf{\uppercase\expandafter{\romannumeral1}}]
    \leq C\left(\sum_{i\ne i_*} \sqrt{\frac{p_{t,i}}{t}}+\frac{1}{t}\right).
    \)
\end{Lem}
\noindent This is a direct corollary from \citet{pmlr-v89-zimmert19a} and we leave the proof in Appendix \ref{proof1}.

Recall that, by \eq{simpleregret}, we have
$2\E[\mathbf{\uppercase\expandafter{\romannumeral1}}]-\operatorname{Reg}^{\mathrm{sim}}_t=2\E\left[\mathbf{\uppercase\expandafter{\romannumeral1}}-\sum_{i\ne i_*}^d p_{t,i}\Delta_i/2\right].$
Then we will compute the conditional expectation of the right-hand side with respect to $\mathscr{F}_{t-1}$. By Lemma \ref{stab}, we have
\begin{equation}\label{self1}
    \begin{aligned}
    \E_{t-1}\left[\mathbf{\uppercase\expandafter{\romannumeral1}}-\sum_{i\ne i_*} p_{t,i}\Delta_i/2\right]\leq \sum_{i\ne i_*}\left(C\sqrt{\frac{p_{t,i}}{t}}-p_{t,i}\Delta_i/2\right)+\frac{C}{t}\leq \frac{C'd}{t\Delta},
\end{aligned}
\end{equation}
where $C'=\frac{C^2}{2}+C$ and we applied $\max_{x\ge 0}\left(a\sqrt{x}-b x\right)\leq \frac{a^2}{4b}$ for any $a,b>0$ and $\Delta=\min\limits_{i\ne i_*} \Delta_i$. Multiplying \eq{self1} by $2$ and absorbing this universal factor into $C$ completes our proof.

\textbf{Combining the two sides}\quad
By the definition of $\mathbf{\uppercase\expandafter{\romannumeral2}}$ and $\eta_t=\alpha/\sqrt{t}$,
\[
\begin{aligned}
2\E[\mathbf{\uppercase\expandafter{\romannumeral2}}]
=\frac{2\sqrt{t}}{\alpha}\left[
\E[D_{\Psi}(e_{i_*},p_t)]
-\left(2-\sqrt{\frac{t+1}{t}}\right)
\E[D_{\Psi}(e_{i_*},p_{t+1})]
\right].
\end{aligned}
\]
Combining the two sides, we have
\begin{equation}\label{final}
    \begin{aligned}
    &\frac{(1-\alpha^2\mu_{i_*})\Delta}{d\,e^{\alpha^2\mu_{i_*}}} \left(\E[D_{\Psi}(e_{i_*},p_t)]\right)^2\\
    &\leq \frac{Cd}{t\Delta}+C'\alpha^{-1}\sqrt{t}\left[\E[D_{\Psi}(e_{i_*},p_t)]-\left(2-\sqrt{\frac{t+1}{t}}\right)\E[D_{\Psi}(e_{i_*},p_{t+1})]\right],
\end{aligned}
\end{equation}
where $C$ and $C'$ are positive constants. This gives rise to an iterative inequality for \(\E[D_{\Psi}(e_{i_*},p_t)]\).  
Intuitively, if we assume \(\E[D_{\Psi}(e_{i_*},p_t)] \sim t^{-k}\), then for sufficiently large \(t\),  
the left-hand side scales as \(t^{-2k}\), while the right-hand side scales as \(t^{-\min(k+\frac12,1)} \).  
This implies that $2k\ge \min(k+\frac12,1)$, leading to $k\ge\frac12$ and  
\(\E[D_{\Psi}(e_{i_*},p_t)] = \mathcal{O}(t^{-\frac12})\).  
A rigorous proof can be obtained directly by applying Lemma~\ref{ratelemma}.

\section{Concluding Remarks}\label{cr}
In summary, we study the last-iterate behavior of the unmodified BOBW FTRL algorithm $1/2$-Tsallis-INF in stochastic bandits. We prove that its sampling simple regret decays at rate $\mathcal{O}(t^{-1})$ without requiring the optimal arm to be unique. As a complementary result, under a unique optimal arm, we prove that $\E[D_{\Psi}(e_{i_*},p_t)]$ decays at rate $\mathcal{O}(t^{-1/2})$. The lower bounds in Theorem~\ref{thm:lower} show that both exponents of $t$ are tight under the same uniqueness condition. Our analysis leaves the algorithm unchanged and hence preserves its BOBW property.

For the sampling simple-regret bound, we develop a Lyapunov analysis that handles the moving empirical minimum over the optimal arms, uses stopped increments to control the larger importance-weighted fluctuations, employs an adaptive skeleton with state-dependent block lengths, and retains a strict local contraction. For the Bregman-divergence bound, we combine an FTRL decomposition with a refined self-bounding argument and a fixed-arm second-moment estimate. Improving the dependence of the upper bounds on the dimension and the gaps, extending the Bregman-divergence guarantee beyond the unique-optimal-arm setting, and developing suitable last-iterate guarantees beyond stochastic bandits are interesting directions for future work.


\newpage
\bibliography{ref}
\bibliographystyle{plainnat}
\appendix
\newpage
\renewcommand{\appendixpagename}{\centering \LARGE Appendix}
\appendixpage

\startcontents[section]
\printcontents[section]{l}{1}{\setcounter{tocdepth}{2}}
\newpage

\section{Experiments}\label{Experiments}
In this section, we present experimental results to validate our theoretical findings. We conduct experiments on a specific stochastic bandit instance. The loss distributions are Bernoulli, with $d=5$ arms and mean loss vector $\mu=(0.1, 0.2, 0.3, 0.4, 0.5)$. We run the algorithm for $10^6$ rounds and repeat the experiment $5000$ times.

\begin{figure}[ht]
    \centering\includegraphics[width=0.8\linewidth]{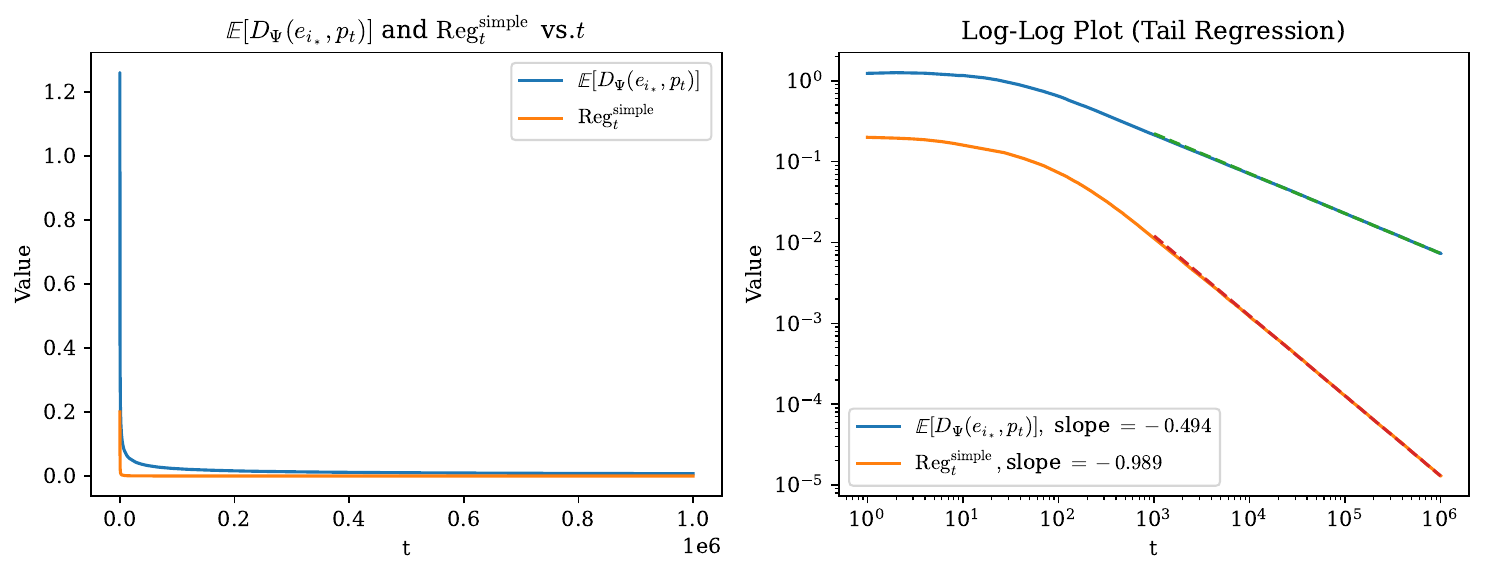} 
    \caption{The left plot show $\mathbb{E}[D_{\Psi}(e_{i_*}, p_t)]$ and $\operatorname{Reg}^{\mathrm{sim}}_t$ versus $t$, and the right plot uses log--log scales. The slopes for the linear fit are $-0.494$ and $-0.989$, respectively.}
    \label{exp}
\end{figure}

We plot $\mathbb{E}[D_{\Psi}(e_{i_*}, p_t)]$ and $\operatorname{Reg}^{\mathrm{sim}}_t$ as functions of $t$ (see Figure~\ref{exp}). The left panel shows the original plots, while the right panel uses log--log scales on both axes. We estimate the convergence rates by performing linear regression on the log--log plots, discarding data points before $t=1000$, since the algorithm has not yet converged in this early phase. The experimental results show that the slope corresponding to $\mathbb{E}[D_{\Psi}(e_{i_*}, p_t)]$ is close to $-0.5$, indicating a rate of $t^{-1/2}$, which is consistent with our theoretical result in Theorem~\ref{rate}. In contrast, the slope corresponding to $\operatorname{Reg}^{\mathrm{sim}}_t$ is close to $-1$, which is consistent with the $t^{-1}$ upper and lower bounds in Theorems~\ref{thm:main} and~\ref{thm:lower}, respectively.

Our experiments are conducted on a server with 4 NVIDIA RTX 4090 GPUs and Intel(R) Xeon(R) Gold 6132 CPU @ 2.60GHz.

\section{Important Facts}\label{important}
This appendix first records structural properties of the
Tsallis-INF FTRL map and then collects the Lyapunov comparisons, moving-gap
control, and counting estimates used in the simple-regret analysis.
\subsection{\texorpdfstring{Structural properties of the
Tsallis-INF FTRL map}{Structural properties of the Tsallis-INF FTRL map}}
This subsection records identities for the Tsallis-INF FTRL map
and the resulting upper and lower bounds on its sampling probabilities.
\begin{Lem}\label{exist}
Suppose \( f \colon \mathbb{R}_+=[0,+\infty) \mapsto  \mathbb{R} \) is a strictly convex function and \(f'\) is continuously differentiable on \( (0,+\infty) \), satisfying \(\lim_{x \to +\infty} f'(x) = 0\) and \(\lim_{x \to 0} f'(x) = -\infty\).
Define $\Psi \colon \R_{+}^{d} \mapsto \R$ and $\phi \colon \R_{+}^{d} \mapsto \Delta^{d-1}$ as $\Psi(\lambda)=\sum_{i=1}^df(\lambda_i) $ and
\[
\phi(\lambda):=\argmax_{p\in\Delta^{d-1}}\agp{p,\lambda}-\Psi(p),
\]
respectively.
For any $\lambda\in\R^d$,
there exists a unique $\nu(\lambda)<-\max\limits_{1\leq i\leq d}\lambda_i$ such that
the $i$-th component of $\phi$ is
\begin{equation}\label{compu}
    \phi_i(\lambda)=(f')^{-1}(\lambda_i+\nu(\lambda)), \quad i=1, \ldots, d,
\end{equation}
with $\nu(\lambda)$ satisfying \begin{equation}\label{constr}
    \sum_{i=1}^d \phi_i(\lambda)=1.
\end{equation}
Moreover, $\nabla \Psi(\phi(\lambda)) = \lambda + \nu(\lambda) \mathbf{1}$.
    \begin{proof}
    First, we prove that for any $\lambda \in \R_{+}^{d}$, there exists a unique solution to the equation $\sum_{i=1}^d  (f')^{-1} (\lambda_i + \nu) = 1$.
    By the condition, $\operatorname{dom}((f')^{-1})\supseteq(-\infty,0)$. For any $\nu<-\max\limits_{1\leq i\leq d}\lambda_i$, let $g(\nu)=\sum\limits_{i=1}^d(f')^{-1}(\lambda_i+\nu)$. Then  $g(\nu)$ is clearly increasing in $\nu$ and we also have
        \[
            \lim_{\nu \to -\infty} g(\nu) = 0, \qquad \lim_{\nu \to -\max\limits_{1\leq i\leq d}\lambda_i} g(\nu) = +\infty.
        \]
        Then by the continuity of $f'$, there exists a unique $\nu(\lambda)<-\max\limits_{1\leq i\leq d}\lambda_i$ such that $g(\nu(\lambda))=1$.

        Now we come back to the simplex-constrained optimization problem for a fixed $\lambda$.
        Suppose that $\tilde{p}$ is a solution.
        From the KKT condition,  $\tilde{p}$ must satisfy $\nabla \Psi(\tilde{p}) = \lambda + z \mathbf{1}$ and $\sum_{i=1}^d \tilde{p}_i=1$, with $z$ the Lagrangian multiplier. From the definition of $\Psi$, this condition implies that $\sum_{i=1}^d (f')^{-1}(\lambda_i+z) = 1$, which has the unique solution $z=\nu(\lambda)$.
        Hence, the solution to the simplex-constrained optimization problem is also unique and satisfies $\tilde{p}_i = (f')^{-1}(\lambda_i+\nu(\lambda))$,
        implying that $\phi(\lambda)$ is well-defined and equal to $(f')^{-1}(\lambda_i+\nu(\lambda))$.
        It follows that $\nabla \Psi(\phi(\lambda)) = \lambda + \nu(\lambda) \mathbf{1}$.
    \end{proof}
\end{Lem}
\begin{Lem}\label{+1}
    For any $a\in\R$ and $\lambda\in\R^d$, we have
    \[
    \phi(\lambda)=\phi(\lambda+a\mathbf{1}).
    \]
    \begin{proof}
        Note that for any $p\in\Delta^{d-1}$, we have
        \[
        \agp{p,\lambda+a\mathbf{1}}-\Psi(p)=\agp{p,\lambda}-\Psi(p)+a,
        \]
        thus the result follows from the definition of $\phi$.
    \end{proof}
\end{Lem}

\begin{Lem}\label{pti-2}
    For any $t\ge 1$, there exists $\nu>-\min\limits_{1\leq i\leq d}\hat{L}_{t, i}$ such that
    \[
    p_{t, i}=4\left(\eta_t\left(\hat{L}_{t, i}+\nu\right)\right)^{-2}=4\left(\eta_t\underline{\hat{L}}_{t, i}+2p_{t,i_{t}^*}^{-\frac{1}{2}}\right)^{-2}.
    \]
    \begin{proof}
        By the definition and Lemma~\ref{+1}, we have $p_t=\phi(-\eta_t\underline{\hat{L}}_t)$, then by Lemma \ref{exist} (with $f(x)=-4\sqrt{x}$), there exists $\nu>-\min\limits_{1\leq i\leq d}\hat{L}_{t, i}$ such that
        \[
        p_{t,i}=\phi_i(-\eta_t\underline{\hat{L}}_t)=4\left(\eta_t\left(\hat{L}_{t, i}+\nu\right)\right)^{-2}.
        \]
        From this, we also have
$
\nu=-\hat{L}_{t,i_{t}^*}+2\eta_t^{-1}p_{t,i_{t}^*}^{-\frac{1}{2}},
$
so for any $1\leq i\leq d$, we have
\[
    p_{t,i}=4\left(\eta_t\left(\hat{L}_{t, i}-\hat{L}_{t,i_{t}^*}\right)+2p_{t,i_{t}^*}^{-\frac{1}{2}}\right)^{-2}=4\left(\eta_t\underline{\hat{L}}_{t, i}+2p_{t,i_{t}^*}^{-\frac{1}{2}}\right)^{-2}.
\]
    \end{proof}
\end{Lem}

\begin{Lem}\label{4-2}
    For any $t\ge 1$, we have
    \[
    4\left(\eta_t\underline{\hat{L}}_{t,i}+2\sqrt{d}\right)^{-2}\leq p_{t,i}\leq 4\left(\eta_t\underline{\hat{L}}_{t,i}\right)^{-2}.
    \]
    \begin{proof}
        By \eq{pti}, for any $1\leq i\leq d$, we have
\[
    p_{t,i}=4\left(\eta_t\underline{\hat{L}}_{t, i}+2p_{t,i_{t}^*}^{-\frac{1}{2}}\right)^{-2}\leq 4\left(\eta_t\underline{\hat{L}}_{t,i}\right)^{-2}.
\]
Similarly, since $p_{t,i_{t}^*}=\max\limits_{1\leq i'\leq d}p_{t,i'}\ge 1/d$, then we have
\[
p_{t,i}\ge 4\left(\eta_t\underline{\hat{L}}_{t,i}+2\sqrt{d}\right)^{-2}.
\]
    \end{proof}
\end{Lem}

\subsection{Auxiliary estimates for the Lyapunov analysis}
\label{sec:auxiliary-estimates}

This subsection collects three types of estimates used in the Lyapunov analysis.
The first compares the two branches of the Lyapunov function,
the second controls the motion of the two empirical minima.
The third converts uniform mass bounds on doubling intervals into power-sum
bounds.

\subsubsection{Comparison between the two Lyapunov branches}

As in \citet[Appendix~G]{zhan2026does}, the local proofs sometimes compare the
two formulas defining the Lyapunov function outside their respective
half-lines.  The following inverse-square analogue of their Lemma~G.1 records
these comparisons together with a uniform upper bound used below.

\begin{Lem}[Comparison between the two branches of $g_t$]
\label{lem:lyapunov-branch-comparison}
Suppose that $t\ge1$.  Then
\begin{enumerate}[label=(\arabic*)]
\item if $\alpha\Delta\sqrt t\ge2$, then, for every
$0\le y<2\sqrt t/\alpha$,
\[
\frac{(y+\Delta t)^2}{\Delta^2t}
\le\frac{4t^2}{(2\sqrt t-\alpha y)^2};
\]
\item if $\alpha\Delta\sqrt t\ge2$, then, for every
$2\sqrt t/\alpha-\Delta t\le y\le0$,
\[
\frac{4t^2}{(2\sqrt t-\alpha y)^2}
\le\frac{(y+\Delta t)^2}{\Delta^2t}.
\]
\item for every $y\in\mathbb R$,
\[
g_t(y)\le2t+\frac{2(y^+)^2}{\Delta^2t}.
\]
\end{enumerate}
\end{Lem}

\begin{proof}
Both comparisons follow from
\[
(y+\Delta t)(2\sqrt t-\alpha y)-2\Delta t^{3/2}
=y(2\sqrt t-\alpha y-\alpha\Delta t).
\]
For the first claim, $y+\Delta t>0$ and $2\sqrt t-\alpha y>0$, while
$2\sqrt t-\alpha y-\alpha\Delta t\le0$.  The right-hand side is therefore
nonpositive.  For the second claim, the same two factors are positive, and
$y\ge2\sqrt t/\alpha-\Delta t$ again gives
$2\sqrt t-\alpha y-\alpha\Delta t\le0$.  Since now $y\le0$, the right-hand
side is nonnegative.  Dividing by the positive factors and squaring gives the
two stated inequalities.

For the third claim, if $y<0$, then
$2\sqrt t-\alpha y\ge2\sqrt t$, so $g_t(y)\le t$.  If $y\ge0$, then
$(y+\Delta t)^2\le2y^2+2\Delta^2t^2$.  These two bounds prove the stated
uniform inequality.
\end{proof}

\subsubsection{Deterministic control of the moving gap}

The sampling-probability identity in \eq{pti} and the probability
bounds in Lemma~\ref{4-2} are exactly those in
\citet[Appendix~D, Lemmas~D.1--D.2]{zhan2026does}.  The following lemma
records the additional deterministic facts needed when both empirical minima
can move.

\begin{Lem}[Moving-minimum gap increments]
\label{lem:gap-increments}
For every $t\ge1$,
\[
M_t=\min\{M_t^{\mathcal O},N_t^{\mathcal S}\},
\qquad
N_t^{\mathcal S}-M_t=(-D_t)^+,
\]
and
\[
0\le M_{t+1}^{\mathcal O}-M_t^{\mathcal O}
\le\hat\ell_{t,i_t^{\mathcal O}},
\qquad
-(N_{t+1}^{\mathcal S}-N_t^{\mathcal S})
\le D_{t+1}-D_t
\le\hat\ell_{t,i_t^{\mathcal O}}.
\]
If $D_t<0$, then $p_{t,i_t^{\mathcal O}}\ge1/d$ and
$0\le\hat\ell_{t,i_t^{\mathcal O}}\le d$.

Moreover, if $t\ge(12dh_t)^2$ and $D_t\ge-\sqrt t$, then for every integer
$0\le r<h_t$,
\[
N_{t+r+1}^{\mathcal S}-N_{t+r}^{\mathcal S}\le 4d.
\]
Consequently, for every integer $0\le r\le h_t$,
\[
D_{t+r}\ge D_t-4dr.
\]
If additionally $D_t\le0$, then
$D_{t+r}\le D_t+4dr$ for every integer $0\le r\le h_t$.
\end{Lem}

\begin{proof}
The two identities follow immediately from the definitions.  Taking $L=1$
in the moving-minimum comparison given in \eq{eq:moving-minimum-comparison} gives
the upper bound on $M_{t+1}^{\mathcal O}-M_t^{\mathcal O}$, and monotonicity
of $N_t^{\mathcal S}$ gives the two one-step bounds for $D_t$.  If $D_t<0$,
$i_t^{\mathcal O}$ is a global empirical minimizer, so
$p_{t,i_t^{\mathcal O}}\ge1/d$ and
$\hat\ell_{t,i_t^{\mathcal O}}\le d$.

For the block bounds, $t\ge(12dh_t)^2$ implies
$4dh_t\le\sqrt{dt}/(4\alpha)$, while $D_t\ge-\sqrt t$ implies
$D_t\ge-\sqrt{dt}/\alpha$.  The induction in
\citet[Appendix~D, Lemma~D.4]{zhan2026does} then applies verbatim, using the
one-step moving-minimum comparison above in place of its fixed-optimal-arm
counterpart.
\end{proof}

\subsubsection{Dyadic counts and power sums}

To see the underlying equivalence in its simplest form, fix a lower cutoff
and place points at integer locations above it.  The sum of their
$\gamma$-th powers up to $t$ is $O(t^\gamma)$, uniformly in $t$, if and only if
every doubling interval above the cutoff contains a uniformly bounded number
of points.  The following lemma records the slightly more general version in
which each integer carries a nonnegative mass, as needed for the entrance
budget.

\begin{Lem}[Dyadic counts and power sums]
\label{lem:dyadic-power-sum}
Let $\gamma>0$ and $n_0\ge1$, and let $(a_r)_{r>n_0}$ be a nonnegative
sequence indexed by the integers.  Then the mass of every doubling interval
is uniformly bounded if and only if the corresponding normalized
$\gamma$-power sum is uniformly bounded.  More precisely,
\[
\begin{aligned}
2^{-\gamma}\sup_{x\ge n_0}\sum_{x<r\le2x}a_r
&\le
\sup_{t\ge n_0}\frac1{t^\gamma}
\sum_{n_0<r\le t}r^\gamma a_r\\
&\le
\frac1{1-2^{-\gamma}}
\sup_{x\ge n_0}\sum_{x<r\le2x}a_r,
\end{aligned}
\]
where the sums are over integer values of $r$ and the suprema are over real
values of $x$ and $t$.
\end{Lem}

\begin{proof}
We first derive the power-sum bound from the doubling-interval mass bound.
Let
\[
A:=\sup_{x\ge n_0}\sum_{x<r\le2x}a_r.
\]
The upper bound is immediate when $t=n_0$.  For $t>n_0$, choose $m\ge0$ so
that $2^m n_0<t\le2^{m+1}n_0$.  For $0\le j<m$, the interval
$(2^{-j-1}t,2^{-j}t]$ has total mass at most $A$, and its contribution after
division by $t^\gamma$ is at most $2^{-j\gamma}A$.  The remaining interval
$(n_0,2^{-m}t]$ is contained in $(n_0,2n_0]$ and contributes at most
$2^{-m\gamma}A$.  Hence
\[
\frac1{t^\gamma}\sum_{n_0<r\le t}r^\gamma a_r
\le A\sum_{j=0}^{m}2^{-j\gamma}
\le\frac{A}{1-2^{-\gamma}}.
\]

Conversely, fix $x\ge n_0$ and use $t=2x$.  Since
$r^\gamma\ge x^\gamma$ on $(x,2x]$,
\[
\frac1{(2x)^\gamma}\sum_{n_0<r\le2x}r^\gamma a_r
\ge2^{-\gamma}\sum_{x<r\le2x}a_r.
\]
Taking the supremum over $x$ proves the other inequality.
\end{proof}

\section{\texorpdfstring{Fixed-Arm Second-Moment and Maximal Bounds}{Fixed-Arm Second-Moment and Maximal Bounds}}
\label{sec:fixed-arm-moments}

\begingroup

This appendix controls the shifted cumulative loss of a fixed optimal arm
over a future interval.  Specifically, for an $\mathscr F_{s-1}$-measurable
arm $i\in\mathcal O$ and a finite integer-valued
$\mathscr F_{s-1}$-measurable endpoint $t\ge s$, our goal is to bound
\[
\E\left[
\max_{s\le u\le t}\underline{\hat L}_{u,i}^{\,2}
\,\middle|\,\mathscr F_{s-1}
\right].
\]
The conditional running-maximum bound is used in the simple-regret analysis
to control the fixed-optimal-arm fluctuations over adaptive blocks, including
blocks restarted after a predictable hitting time.  Its specialization to
$s=1$ also retains the explicit dependence on $\alpha$ and $\mu_*$ required
by the fixed-time second-moment estimate in the Bregman-divergence argument.

The main difficulty is that standard concentration inequalities do not
provide the required control.  Although $\hat L_{t,j}$ is an unbiased
estimator of the cumulative loss of arm~$j$, importance weighting can make
its variance large.  Indeed,
\[
\hat\ell_{u,j}=\frac{\ell_{u,j}A_{u,j}}{p_{u,j}},
\qquad
A_{u,j}:=\1_{\{I_u=j\}},
\]
so when a suboptimal arm is sampled with probability of order $u^{-1}$, its
one-step conditional second moment can be of order $u$.  The accumulated
variance up to time $t$ can therefore be of order $t^2$, making the standard
deviation of the cumulative estimator comparable to its mean.

A further difficulty comes from the data-dependent empirical minimizer.  Let
$i_u^*\in\argmin_{j\in\mathcal I}\hat L_{u,j}$.  Since $i_u^*$ depends on the
same estimated losses that we wish to control, the two terms in
\[
\underline{\hat L}_{u,i}=\hat L_{u,i}-\hat L_{u,i_u^*}
\]
cannot be decoupled directly by conditioning.  We address this difficulty
through the interval decomposition
\[
\underline{\hat L}_{u,i}=U_u^{(s)}+\widehat R_u^{(s)},
\]
where
\[
U_u^{(s)}
:=\sum_{r=s}^{u-1}\langle e_i-p_r,\hat\ell_r\rangle,
\qquad
\widehat R_u^{(s)}
:=\sum_{r=s}^{u-1}\langle p_r-e_{i_u^*},\hat\ell_r\rangle
+\hat L_{s,i}-\hat L_{s,i_u^*}.
\]
This separates the fixed-arm term $U_u^{(s)}$, which can be handled by
conditional second-moment and martingale maximal estimates, from the
estimated-regret term $\widehat R_u^{(s)}$, which can be handled by the
standard pathwise regret analysis for FTRL.  The first lemma below establishes
the pathwise interval estimated-regret bound used to control
$\widehat R_u^{(s)}$.  We then control $U_u^{(s)}$ and combine the two
estimates to obtain the desired conditional maximal bound.
\endgroup

\begin{Lem}[Interval estimated-regret bound]
\label{lem:interval-regret-bound}
For every pair of integers $1\le s\le u$ and every $j\in\mathcal I$,
pathwise,
\begin{equation}
\label{eq:interval-regret-bound}
\sum_{r=s}^{u-1}\langle p_r-e_j,\hat\ell_r\rangle
\le \underline{\hat L}_{s,j}
+\frac{4(\sqrt d-1)\sqrt u}{\alpha}
+\frac\alpha2\sum_{r=s}^{u-1}\frac1{\sqrt{r p_{r,I_r}}}.
\end{equation}
\end{Lem}

\begin{proof}
Set $\Psi(q):=-4\sum_{j\in\mathcal I}\sqrt{q_j}$ and
$D_\Psi(q,p):=\Psi(q)-\Psi(p)-\langle\nabla\Psi(p),q-p\rangle$.
Fix $s<u$ and set $n=u-s$.  In
\citet[Exercise~28.12(a)]{lattimore2020bandit}, make the substitutions
\[
a_k=p_{s+k-1},\qquad
y_k=\hat\ell_{s+k-1},\qquad
F_k(q)=\langle q,\hat L_s\rangle
      +\eta_{s+k-1}^{-1}\Psi(q),
\qquad 1\le k\le n+1.
\]
Indeed,
\[
F_k(q)+\sum_{v=1}^{k-1}\langle q,y_v\rangle
=\langle q,\hat L_{s+k-1}\rangle
+\eta_{s+k-1}^{-1}\Psi(q),
\]
so the minimizer denoted by $a_k$ in that exercise is $p_{s+k-1}$, and
$D_{F_k}=\eta_{s+k-1}^{-1}D_\Psi$.  Part~(a) therefore gives
\[
\begin{aligned}
\sum_{r=s}^{u-1}\langle p_r-e_j,\hat\ell_r\rangle
\le{}&\hat L_{s,j}-\langle p_s,\hat L_s\rangle\\
&+\sum_{r=s}^{u-1}\left(
\langle p_r-p_{r+1},\hat\ell_r\rangle
-\eta_r^{-1}D_\Psi(p_{r+1},p_r)
\right)\\
&+\eta_s^{-1}\bigl(\Psi(e_j)-\Psi(p_s)\bigr)\\
&+\sum_{r=s}^{u-1}(\eta_{r+1}^{-1}-\eta_r^{-1})
\bigl(\Psi(e_j)-\Psi(p_{r+1})\bigr).
\end{aligned}
\]
The first term is at most
$\hat L_{s,j}-M_s=\underline{\hat L}_{s,j}$.  Since
$\eta_r^{-1}$ is increasing and
$0\le\Psi(e_j)-\Psi(p_r)\le4(\sqrt d-1)$, the last two lines are at most
\[
4(\sqrt d-1)\left(
\eta_s^{-1}+\sum_{r=s}^{u-1}
(\eta_{r+1}^{-1}-\eta_r^{-1})
\right)
=\frac{4(\sqrt d-1)\sqrt u}{\alpha}.
\]
Finally, Lemma~\ref{x1} and \eq{bri}, followed by
Lemma~\ref{111}, give
\[
\langle p_r-p_{r+1},\hat\ell_r\rangle
-\eta_r^{-1}D_\Psi(p_{r+1},p_r)
\le\frac{\alpha}{2\sqrt{r p_{r,I_r}}},
\qquad s\le r<u.
\]
Substitution proves \eq{eq:interval-regret-bound} for $s<u$; for $s=u$,
the inequality is immediate.
\end{proof}

Equation~\eqref{eq:interval-regret-bound} controls the estimated regret against
every comparator arm.  We apply it below to an empirical minimizer at each
endpoint.

\begin{Lem}[Conditional maximal estimate for a fixed optimal arm]
\label{lem:fixed-arm-maximal}
Suppose that $\alpha\in(0,1)$.  Let $s\ge1$, let $i\in\mathcal O$ be
$\mathscr F_{s-1}$-measurable, and
let $t\ge s$ be a finite integer-valued $\mathscr F_{s-1}$-measurable random
variable.  Then
\[
{\begin{aligned}
&
\E\left[
\max_{s\le u\le t}\underline{\hat L}_{u,i}^{2}
\,\middle|\,\mathscr F_{s-1}
\right]\\
&\quad\le
{
C\Bigg\{
\frac{e^{\alpha^2\mu_*}}{1-\alpha^2\mu_*}
\left[
\left(\frac ts\right)^{\alpha^2\mu_*}
\underline{\hat L}_{s,i}^{2}
+(1+d\mu_*)(t-s)
\right]
+\min\left\{\frac{dt}{\alpha^2},(t-s)^2\right\}
\Bigg\}}
.
\end{aligned}}
\]
Here $C$ is a universal positive constant.
\end{Lem}

\begin{proof}
Conditioning on $\mathscr F_{s-1}$ fixes $i$ and $t$.  For $s\le u\le t$,
recall that $i_u^*\in\argmin_{j\in\mathcal I}\hat L_{u,j}$, and define
\[
\begin{aligned}
U_u^{(s)}
&:=\sum_{r=s}^{u-1}\langle e_i-p_r,\hat\ell_r\rangle,\\
\widehat R_u^{(s)}
&:=\sum_{r=s}^{u-1}\langle p_r-e_{i_u^*},\hat\ell_r\rangle
+\hat L_{s,i}-\hat L_{s,i_u^*}.
\end{aligned}
\]
Direct expansion gives
\begin{equation}
\label{eq:fixed-arm-decomposition}
\underline{\hat L}_{u,i}=U_u^{(s)}+\widehat R_u^{(s)}.
\end{equation}
We control the two terms separately, first $\widehat R_u^{(s)}$ and then
$U_u^{(s)}$.

\paragraph{Control of $\widehat R_u^{(s)}$.}
Since \eq{eq:interval-regret-bound} holds pathwise for every comparator
arm, we may apply it with $j=i_u^*$.  The identity
$\underline{\hat L}_{s,i_u^*}+\hat L_{s,i}-\hat L_{s,i_u^*}
=\underline{\hat L}_{s,i}$ gives
\[
\max_{s\le u\le t}(\widehat R_u^{(s)})^+
\le \underline{\hat L}_{s,i}+\frac{4\sqrt{dt}}\alpha
+\frac\alpha2\sum_{r=s}^{t-1}\frac1{\sqrt{r p_{r,I_r}}}.
\]
Here the case $u=s$ follows directly from
$\widehat R_s^{(s)}=\underline{\hat L}_{s,i}$.  Conditional Minkowski's
inequality and
$\E[(r p_{r,I_r})^{-1}\mid\mathscr F_{r-1}]=d/r$ give
\[
\E\left[
\left(\sum_{r=s}^{t-1}\frac1{\sqrt{r p_{r,I_r}}}\right)^2
\,\middle|\,\mathscr F_{s-1}\right]^{1/2}
\le\sqrt d\sum_{r=s}^{t-1}\frac1{\sqrt r}
\le C\sqrt{d(t-s)}.
\]
Consequently, the preceding pathwise bound implies
\[
{
\E\left[
\max_{s\le u\le t}\bigl((\widehat R_u^{(s)})^+\bigr)^2
\,\middle|\,\mathscr F_{s-1}\right]
\le C\left(\underline{\hat L}_{s,i}^2+\frac{dt}{\alpha^2}\right).}
\]
There is also a pathwise bound that retains the interval length.  Since the
estimated losses are nonnegative and
$\hat L_{s,i}-\hat L_{s,i_u^*}\le\underline{\hat L}_{s,i}$, while
$\langle p_r,\hat\ell_r\rangle=\ell_{r,I_r}\le1$, the definition of
$\widehat R_u^{(s)}$ gives, for every $s\le u\le t$,
\[
\widehat R_u^{(s)}
\le\sum_{r=s}^{u-1}\langle p_r,\hat\ell_r\rangle
+\underline{\hat L}_{s,i}
\le u-s+\underline{\hat L}_{s,i}.
\]
Consequently,
\[
\max_{s\le u\le t}(\widehat R_u^{(s)})^+
\le\underline{\hat L}_{s,i}+t-s.
\]
Combining the last two bounds according to whether
$dt/\alpha^2$ or $(t-s)^2$ is smaller gives
\begin{equation}
\label{eq:fixed-arm-regret-moment}
{
\E\left[
\max_{s\le u\le t}\bigl((\widehat R_u^{(s)})^+\bigr)^2
\,\middle|\,\mathscr F_{s-1}\right]
\le C\left\{
\underline{\hat L}_{s,i}^2
+\min\left\{\frac{dt}{\alpha^2},(t-s)^2\right\}
\right\}.}
\end{equation}

\paragraph{Control of $U_u^{(s)}$.}
We first derive an endpoint estimate.  Since
$\mu_i-\langle p_u,\mu\rangle\le0$ for every $i\in\mathcal O$,
write
$Z_u:=\langle e_i-p_u,\hat\ell_u\rangle$.  If $U_u^{(s)}<0$, then
$((U_{u+1}^{(s)})^+)^2\le \hat\ell_{u,i}^2$, whose conditional expectation
is at most $\mu_*/p_{u,i}$.  If $U_u^{(s)}\ge0$, expanding the square and
using
$\E[Z_u\mid\mathscr F_{u-1}]=\mu_*-\langle p_u,\mu\rangle\le0$ and
$\E[Z_u^2\mid\mathscr F_{u-1}]\le1+\mu_*/p_{u,i}$ gives,
for $s\le u<t$,
\[
{
\E\left[((U_{u+1}^{(s)})^+)^2\mid\mathscr F_{u-1}\right]
\le ((U_u^{(s)})^+)^2+1+\frac{\mu_*}{p_{u,i}}.}
\]
The lower sampling-probability bound in Lemma~\ref{4-2}
and \eq{eq:fixed-arm-decomposition} imply
\[
p_{u,i}^{-1}
\le2d+\frac{\alpha^2\underline{\hat L}_{u,i}^2}{2u},
\qquad
\frac{\underline{\hat L}_{u,i}^2}{2}
\le ((U_u^{(s)})^+)^2+((\widehat R_u^{(s)})^+)^2.
\]
Set $a:=\alpha^2\mu_*$; then $0\le a<1$.
Taking conditional expectations and applying
\eq{eq:fixed-arm-regret-moment} on $[s,u]$ gives
\[
{
\begin{aligned}
\E[((U_{u+1}^{(s)})^+)^2\mid\mathscr F_{s-1}]
&\le\left(1+\frac{a}{u}\right)
\E[((U_u^{(s)})^+)^2\mid\mathscr F_{s-1}]\\
&\quad+C(1+d\mu_*)
+\frac{Ca\underline{\hat L}_{s,i}^2}{u}.
\end{aligned}}
\]
Here we used
$\frac{a}{u}\min\{du/\alpha^2,(u-s)^2\}\le d\mu_*$.
Starting from $U_s^{(s)}=0$,
iteration of this recurrence gives
\[
{
\begin{aligned}
\E[((U_u^{(s)})^+)^2\mid\mathscr F_{s-1}]
&\le C\sum_{v=s}^{u-1}
\left(1+d\mu_*+\frac{a\underline{\hat L}_{s,i}^2}{v}\right)
\prod_{r=v+1}^{u-1}\left(1+\frac{a}{r}\right)\\
&\le C\sum_{v=s}^{u-1}
\left(1+d\mu_*+\frac{a\underline{\hat L}_{s,i}^2}{v}\right)
\left(\frac uv\right)^a\\
&\le \frac{Ce^a}{1-a}\left\{
(1+d\mu_*)(u-s)+\left(\frac us\right)^a
\underline{\hat L}_{s,i}^2
\right\}.
\end{aligned}}
\]
Together with \eq{eq:fixed-arm-decomposition} and
\eq{eq:fixed-arm-regret-moment} on $[s,u]$, this gives
\begin{equation}
\label{eq:fixed-arm-endpoint-moment}
{
\begin{aligned}
&\E[\underline{\hat L}_{u,i}^2\mid\mathscr F_{s-1}]\\
&\quad\le C\Bigg\{
\frac{e^a}{1-a}\left[
\left(\frac us\right)^a\underline{\hat L}_{s,i}^2
+(1+d\mu_*)(u-s)
\right]
+\min\left\{\frac{du}{\alpha^2},(u-s)^2\right\}
\Bigg\}.
\end{aligned}}
\end{equation}

To control the running maximum, we apply Doob's inequality to the martingale
obtained by centering the increments of $U_u^{(s)}$.  Define
\[
\widetilde U_u^{(s)}:=\sum_{r=s}^{u-1}\left(
\langle e_i-p_r,\hat\ell_r\rangle
-\E[\langle e_i-p_r,\hat\ell_r\rangle
\mid\mathscr F_{r-1}]
\right).
\]
Because $i\in\mathcal O$,
direct conditional expectation gives
$\mu_i-\langle p_r,\mu\rangle\le0$ for every $r\ge s$, and hence
$U_u^{(s)}\le\widetilde U_u^{(s)}$.
The corresponding conditional second-moment calculation,
together with the inverse-probability consequence of
Lemma~\ref{4-2}, gives
\[
{
\begin{aligned}
\operatorname{Var}\left(
\langle e_i-p_u,\hat\ell_u\rangle
\,\middle|\,\mathscr F_{u-1}\right)
&\le1+\frac{\mu_*}{p_{u,i}}\\
&\le1+2d\mu_*
+\frac{a\underline{\hat L}_{u,i}^2}{2u}.
\end{aligned}}
\]
Martingale orthogonality, the preceding variance bound, and
\eq{eq:fixed-arm-endpoint-moment} give
\[
{
\begin{aligned}
\E[(\widetilde U_t^{(s)})^2\mid\mathscr F_{s-1}]
&=\sum_{u=s}^{t-1}\E\left[
\operatorname{Var}\left(
\langle e_i-p_u,\hat\ell_u\rangle
\,\middle|\,\mathscr F_{u-1}\right)
\,\middle|\,\mathscr F_{s-1}\right]\\
&\le C(1+d\mu_*)(t-s)
+\frac a2\sum_{u=s}^{t-1}
\frac{\E[\underline{\hat L}_{u,i}^2\mid\mathscr F_{s-1}]}{u}\\
&\le \frac{Ce^a}{1-a}\left\{
\left(\frac ts\right)^a\underline{\hat L}_{s,i}^2
+(1+d\mu_*)(t-s)
\right\}.
\end{aligned}}
\]
Since $(U_u^{(s)})^+\le|\widetilde U_u^{(s)}|$, Doob's inequality now gives
\[
{
\E\left[
\max_{s\le u\le t}((U_u^{(s)})^+)^2
\,\middle|\,\mathscr F_{s-1}\right]
\le \frac{Ce^a}{1-a}\left\{
\left(\frac ts\right)^a\underline{\hat L}_{s,i}^2
+(1+d\mu_*)(t-s)
\right\}.}
\]
Finally,
$\underline{\hat L}_{u,i}\le(U_u^{(s)})^++(\widehat R_u^{(s)})^+$.
Combining the last display with \eq{eq:fixed-arm-regret-moment} proves the
lemma.
\end{proof}

\begingroup

Taking $s=1$, a deterministic optimal arm $i$, and a deterministic endpoint
$t$ in Lemma~\ref{lem:fixed-arm-maximal}, and using
$\underline{\hat L}_{1,i}=0$, gives
\[
\E[\underline{\hat L}_{t,i}^2]
\le C\left[
\frac d{\alpha^2}
+(1+d\mu_*)\frac{e^{\alpha^2\mu_*}}
{1-\alpha^2\mu_*}
\right]t.
\]
Thus the same lemma provides the fixed-time parameter dependence used by the
Bregman argument and the conditional interval maximum used by the Lyapunov
argument.
\endgroup

\section{Block and parameter setup}
\label{sec:parameters}

This appendix fixes the block parameters used throughout the proof.  Their
compatibility with all local and global estimates is verified in
Appendix~\ref{sec:parameter-verification}.  Assume that
$\mathcal S\ne\varnothing$ and recall
$\Delta=\min_{i\in\mathcal S}\Delta_i$. Choose
\begin{align*}
A&=\left\lceil C\Delta^{-2}\log(2|\mathcal S|)\right\rceil,
&
\beta&=\frac{1}{C_\alpha A},
\end{align*}
where the universal constant in $A$ and the constant depending only on
$\alpha$ in the denominator defining $\beta$ are fixed sufficiently large.
Define the predictable positive-integer step
\[
h_t:=
\begin{cases}
\displaystyle
\left\lceil
A\left(\sqrt d+\frac{\alpha(-D_t)^+}{2\sqrt t}\right)^2
\right\rceil,
&D_t\ge-\beta t,\\[2mm]
1,&D_t< -\beta t.
\end{cases}
\]
The first branch is the planned block length in the local estimates, while
the second makes the global skeleton advance one round at a time outside the
strip.  Thus $A$ sets the local block scale and $\beta$ specifies the strip
$D_t\ge-\beta t$.  Finally, let $N$ be the least integer strictly larger than
\[
C'_\alpha\max\{A^2d^4,\beta^{-2}\},
\]
where this last constant is chosen after the constants in $A$ and $\beta$.
The parameter $N$ is the common burn-in time.  Appendix~
\ref{sec:parameter-verification} checks the conditions imposed on these
choices and simplifies the resulting dependence on $d$ and $\Delta$.

\section{Proofs of the global estimates}
\label{sec:global-proofs}

This appendix proves the parameter-form Lyapunov estimate and then derives
the all-time moment bound stated in the main text.

\begin{Thm}[Uniform Lyapunov moment]
\label{thm:uniform-moment}
For every $t\ge N$,
\begin{equation}
\label{eq:uniform-moment}
\E[g_t(D_t)]
\le C_\alpha\bigl\{N+\beta^{-2}
(1+|\mathcal S|\beta^{-2})\bigr\}.
\end{equation}
\end{Thm}

Theorem~\ref{thm:uniform-moment} is the sharper parameter-form input for
Theorem~\ref{thm:coarse-uniform-moment}.  After proving the parameter-form
estimate, we combine it with the parameter simplification and a short
small-time argument to obtain the all-time statement.

The proof follows the stopping-time framework of
\citet[Appendix~G]{zhan2026does}.  We first control the Lyapunov potential at
the last adaptive-skeleton node before a prescribed time $t$.  The resulting
bound retains a weighted sum over the entrances into the controlled strip;
the entrance budget then closes this sum.  Only after obtaining the
skeleton-node bound do we use Lemma~\ref{lem:one-block-maximal} to cross the
remaining incomplete block and reach time $t$.

Starting from $s_1:=N$, define the adaptive skeleton and its last node before
$t\ge N$ by
\[
s_{j+1}:=s_j+h_{s_j},
\qquad
J_t:=\max\{j:s_j\le t\}.
\]
Since $h_r\ge1$, the index $J_t$ is well defined.  Also write
$S_1<S_2<\cdots$ for the successive calendar times $u\ge N$ such that
\[
D_{u-1}< -\beta(u-1)
\qquad\text{and}\qquad
D_u\ge-\beta u.
\]

The first global input is the promised skeleton-node decomposition.  Its
proof uses Lemma~\ref{lem:one-block-contraction} to verify the one-step
supermartingale inequality and then applies the bounded-stopping-time
extension of \citet[Appendix~G, Lemma~G.2]{zhan2026does} recorded in
Lemma~\ref{lem:finite-horizon-stopping}.

\begin{Lem}[Skeleton-node entrance decomposition]
\label{lem:entrance-decomposition}
There exist $\gamma\in(0,1]$ and $C_\alpha>0$, both depending only on $\alpha$,
such that, for every $t\ge N$,
\begin{align}
\label{eq:excursion-decomposition-bound}
&\E\left[
g_{s_{J_t}}(D_{s_{J_t}})
\1_{\{D_{s_{J_t}}\ge-\beta s_{J_t}\}}
\right]\nonumber\\
&\quad\le
C_\alpha\left(\frac Nt\right)^\gamma
\E[g_N(D_N)\1_{\{D_N\ge-\beta N\}}]
+C_\alpha\E\left[
\sum_{k:N<S_k\le t}
\left(\frac{S_k}{t}\right)^\gamma
g_{S_k}(D_{S_k})
\right].
\end{align}
\end{Lem}

Fix the exponent $\gamma$ supplied by
Lemma~\ref{lem:entrance-decomposition}.  To control the entrance term in
that lemma, we use the following entrance budget.

\begin{Lem}[Entrance budget]
\label{lem:entrance-budget}
For every $t\ge N$,
\begin{equation}
\label{eq:weighted-entrance-potential}
\E\left[
\sum_{k:N<S_k\le t}
\left(\frac{S_k}{t}\right)^\gamma
g_{S_k}(D_{S_k})
\right]
\le C_\alpha\beta^{-2}
(1+|\mathcal S|\beta^{-2}).
\end{equation}
\end{Lem}

We first combine these two global inputs with the local maximal estimate to
prove Theorem~\ref{thm:uniform-moment}.  The proofs of
Lemmas~\ref{lem:entrance-decomposition} and~\ref{lem:entrance-budget} follow
the theorem proof.

\subsection{Proof of the uniform Lyapunov theorem}

The proof first combines the skeleton-node decomposition with the entrance
budget.  It then uses the local maximal estimate only for the final incomplete
block and adds the direct inverse-square bound below the strip.

\begin{proof}[Proof of Theorem~\ref{thm:uniform-moment}]
We first control the single initial time $N$.  Fix any deterministic
$i_*\in\mathcal O$.  Apply Lemma~\ref{lem:fixed-arm-maximal} to $i_*$ on
$[1,N]$.  Since $\underline{\hat L}_{1,i_*}=0$, it gives
\begin{equation}
\label{eq:initial-second-moment}
\E[\underline{\hat L}_{N,i_*}^2]
\le C_\alpha\left\{d(N-1)+\min\{dN,(N-1)^2\}\right\}
\le C_\alpha d N.
\end{equation}
Moreover,
\[
D_N^+\le\underline{\hat L}_{N,i_*}.
\]
Indeed, when $D_N>0$ we have $M_N=N_N^{\mathcal S}$ and
$M_N^{\mathcal O}\le\hat L_{N,i_*}$; when $D_N\le0$ the claim is immediate.
Lemma~\ref{lem:lyapunov-branch-comparison}(3), applied with $t=N$ and
$y=D_N$, gives
\[
g_N(D_N)
\le\frac{2(D_N^+)^2}{\Delta^2N}+2N.
\]
Using \eq{eq:initial-second-moment} and the consequence
$N\ge d\Delta^{-2}$ of the parameter choices, we obtain
\begin{equation}
\label{eq:initial-potential}
\E[g_N(D_N)\1_{\{D_N\ge-\beta N\}}]
\le C_\alpha N.
\end{equation}

Lemma~\ref{lem:entrance-budget} now closes the entrance term in
Lemma~\ref{lem:entrance-decomposition}.  Substituting
\eq{eq:initial-potential} and \eq{eq:weighted-entrance-potential} into
\eq{eq:excursion-decomposition-bound}, and using
$(N/t)^\gamma\le1$, gives
\begin{equation}
\label{eq:last-skeleton-moment}
\E\left[
g_{s_{J_t}}(D_{s_{J_t}})
\1_{\{D_{s_{J_t}}\ge-\beta s_{J_t}\}}
\right]
\le C_\alpha\bigl\{N+\beta^{-2}
(1+|\mathcal S|\beta^{-2})\bigr\}.
\end{equation}

It remains to pass from the last skeleton node to the prescribed time $t$.
On $\{D_t\ge-\beta t\}$, the node $s_{J_t}$ also lies in the strip.  Indeed,
if $s_{J_t}<t$ and $D_{s_{J_t}}< -\beta s_{J_t}$, then the second branch in
the definition of $h_{s_{J_t}}$ gives $s_{J_t+1}=s_{J_t}+1\le t$,
contradicting the definition of $J_t$; the case $s_{J_t}=t$ is immediate.
We therefore have the pathwise bound
\[
g_t(D_t)\1_{\{D_t\ge-\beta t\}}
\le
\1_{\{D_{s_{J_t}}\ge-\beta s_{J_t}\}}
\max_{s_{J_t}\le u\le s_{J_t+1}}g_u(D_u).
\]
For each $j$, the event
\[
\{J_t=j,\ D_{s_j}\ge-\beta s_j\}
\]
is $\mathscr F_{s_j-1}$-measurable because $s_{j+1}$ is known at time
$s_j-1$.  Partitioning by these events and then by $\{s_j=r\}$, and applying
Lemma~\ref{lem:one-block-maximal} conditionally at each deterministic $r$,
gives
\[
\E[g_t(D_t)\1_{\{D_t\ge-\beta t\}}]
\le C_\alpha
\E\left[
g_{s_{J_t}}(D_{s_{J_t}})
\1_{\{D_{s_{J_t}}\ge-\beta s_{J_t}\}}
\right].
\]
Combining this inequality with \eq{eq:last-skeleton-moment} controls the
strip-restricted moment.

Finally, on $\{D_t< -\beta t\}$,
\[
g_t(D_t)
\le\frac{4t^2}{(2\sqrt t+\alpha\beta t)^2}
\le\frac{4}{\alpha^2\beta^2}
\le C_\alpha\beta^{-2}.
\]
Adding the strip-restricted and complementary-region bounds proves
\eq{eq:uniform-moment}.
\end{proof}

\begin{proof}[Proof of Theorem~\ref{thm:coarse-uniform-moment}]
Suppose first that $t\ge N$.  Theorem~\ref{thm:uniform-moment} and the
parameter simplification in Appendix~\ref{sec:parameter-verification} give
\[
\E[g_t(D_t)]
\le C_\alpha d^4\Delta^{-8}\log^4(2d).
\]

It remains to consider $1\le t<N$.  Fix any deterministic
$i_*\in\mathcal O$.  If $D_t>0$, then $M_t=N_t^{\mathcal S}$ and
$M_t^{\mathcal O}\le\hat L_{t,i_*}$, so
$D_t\le\underline{\hat L}_{t,i_*}$; if $D_t\le0$, the inequality
$D_t^+\le\underline{\hat L}_{t,i_*}$ is immediate.  Lemma~
\ref{lem:lyapunov-branch-comparison}(3), followed by
Lemma~\ref{lem:fixed-arm-maximal} with initial time $1$ and endpoint $t$,
therefore yields
\[
\E[g_t(D_t)]
\le 2t+\frac{2\E[(D_t^+)^2]}{\Delta^2t}
\le 2t+C_\alpha d\Delta^{-2}.
\]
Since $t<N$ and Appendix~\ref{sec:parameter-verification} gives
$N\le C_\alpha d^4\Delta^{-4}\log^2(2d)$, while
$d\ge2$ and $\Delta\le1$, the last display is at most
$C_\alpha d^4\Delta^{-8}\log^4(2d)$.  This proves the all-time bound.
\end{proof}

\subsection{Proof of the skeleton-node decomposition}

We now prove Lemma~\ref{lem:entrance-decomposition}.  The one-block
contraction verifies the hypothesis of the finite-horizon stopping lemma on
the adaptive skeleton, after which the skeleton reentries are mapped to the
calendar-time entrances $S_k$.

\begin{proof}[Proof of Lemma~\ref{lem:entrance-decomposition}]
Because $D_r$ and $h_r$ are $\mathscr F_{r-1}$-measurable, each $s_j$ is a
predictable integer-valued stopping time, $s_{j+1}$ is known at time
$s_j-1$, and $(\mathscr G_j)_{j\ge1}$ defined by
\[
\mathscr G_j:=\mathscr F_{s_j-1}
\]
is a filtration.  To apply Lemma~\ref{lem:finite-horizon-stopping}, define,
only for this proof,
\[
X_j:=D_{s_j}+\beta s_j,
\qquad
\mathcal M_j:=[0,\infty),
\qquad
H_j:=s_j^\gamma g_{s_j}(D_{s_j}).
\]
The processes $(X_j)$ and $(H_j)$ are adapted to $(\mathscr G_j)$, and
$H_j$ is nonnegative.

On $\{X_j\in\mathcal M_j\}$, the skeleton node lies in the strip, so the
definition of $h_{s_j}$ and Appendix~\ref{sec:parameter-verification} give
\begin{equation}
\label{eq:skeleton-step-size}
1\le h_{s_j}\le s_j/8.
\end{equation}
Let $C_\alpha$ be the constant in
Lemma~\ref{lem:one-block-contraction}, and fix
$\gamma:=\min\{1,(4C_\alpha)^{-1}\}$.  Because the contraction constant
depends only on $\alpha$, the resulting $\gamma\in(0,1]$ does as well, and it
satisfies $4\gamma\le C_\alpha^{-1}$.  The one-block contraction then gives
\[
\E[g_{s_{j+1}}(D_{s_{j+1}})\mid\mathscr G_j]
\le
\left(1-4\gamma\frac{h_{s_j}}{s_j}\right)
g_{s_j}(D_{s_j})
\qquad\text{on }\{X_j\in\mathcal M_j\}.
\]
This random-time conditional inequality follows by applying the
deterministic-time statement on each event $\{s_j=r\}$.  Since
$(1+x)^\gamma\le1+\gamma x$ for $x\ge0$, we have
\[
(1+x)^\gamma(1-4\gamma x)\le1,
\qquad 0\le x\le1/8.
\]
Consequently,
\[
\E[H_{j+1}\mid\mathscr G_j]\le H_j
\qquad\text{on }\{X_j\in\mathcal M_j\}.
\]
Let $(\sigma_k,\tau_k)_{k\ge0}$ be the entrance and exit indices associated
with $(X_j,\mathcal M_j)$ in Lemma~\ref{lem:finite-horizon-stopping}.  The
lower bound $h_r\ge1$ gives $J_t\le t-N+1$, while
$\{J_t=j\}=\{s_j\le t<s_{j+1}\}\in\mathscr G_j$; hence $J_t$ is a bounded
stopping index for the skeleton filtration.  Applying
Lemma~\ref{lem:finite-horizon-stopping} with $T=J_t$ gives
\begin{align*}
\E[H_{J_t}\1_{\{X_{J_t}\in\mathcal M_{J_t}\}}]
&\le
\E[H_1\1_{\{X_1\in\mathcal M_1\}}]\\
&\quad+
\sum_{k\ge1}
\E[H_{\sigma_k}\1_{\{\sigma_k\le J_t\}}].
\end{align*}

We next express the skeleton reentries through the calendar-time entrances
$S_k$.  If $k\ge1$ and $\sigma_k<\infty$, then the skeleton node immediately
before $s_{\sigma_k}$ is outside the strip.  The second branch in the
definition of $h_t$ gives $s_{\sigma_k}=s_{\sigma_k-1}+1$, so
$s_{\sigma_k}$ is one of the entrance times $S_m$.  Distinct skeleton
reentries give distinct entrance times, and $\sigma_k\le J_t$ implies
$N<s_{\sigma_k}\le t$.  Thus the skeleton reentries form a subset of the
entrance times appearing in \eq{eq:excursion-decomposition-bound}.
\begin{samepage}
Substituting $H_1=N^\gamma g_N(D_N)$ and
$H_{\sigma_k}=s_{\sigma_k}^\gamma
g_{s_{\sigma_k}}(D_{s_{\sigma_k}})$ into the finite-horizon skeleton bound
and enlarging its nonnegative reentry sum to all $S_m$ satisfying
$N<S_m\le t$ gives
\begin{align*}
&t^{-\gamma}
\E[H_{J_t}\1_{\{X_{J_t}\in\mathcal M_{J_t}\}}]\\
&\quad\le
\left(\frac Nt\right)^\gamma
\E[g_N(D_N)\1_{\{D_N\ge-\beta N\}}]
+\E\left[
\sum_{m:N<S_m\le t}
\left(\frac{S_m}{t}\right)^\gamma
g_{S_m}(D_{S_m})
\right].
\end{align*}
\end{samepage}
On $\{X_{J_t}\in\mathcal M_{J_t}\}$, either $s_{J_t}=t$ or
\eq{eq:skeleton-step-size} and $t<s_{J_t+1}$ give
$s_{J_t}\ge8t/9$.  Therefore,
\[
\E\left[
g_{s_{J_t}}(D_{s_{J_t}})
\1_{\{D_{s_{J_t}}\ge-\beta s_{J_t}\}}
\right]
\le
\left(\frac98\right)^\gamma t^{-\gamma}
\E[H_{J_t}\1_{\{X_{J_t}\in\mathcal M_{J_t}\}}].
\]
Absorbing $(9/8)^\gamma$ into $C_\alpha$ proves
\eq{eq:excursion-decomposition-bound}.
\end{proof}

\subsection{Proof of the entrance budget}

The entrance budget closes the term left by
Lemma~\ref{lem:entrance-decomposition} by separating the size of each
entrance contribution from the frequency of the entrance times.

\begin{proof}[Proof of Lemma~\ref{lem:entrance-budget}]
The weighted entrance term involves both the potential at each entrance and
the distribution of the entrance times.  We first bound the potential at an
entrance, leaving only a normalized power sum of the entrance times.  We then
encode their expected frequency as a deterministic mass sequence.
Lemma~\ref{lem:dyadic-power-sum} says that the resulting power-sum bound is
equivalent to a uniform mass bound on every doubling interval.  Finally, we
pair each entrance after the first with an earlier exit from the strip and
prove a one-step exit probability of order $1/u$; this order is exactly
sufficient because $\sum_{x\le u\le2x}u^{-1}$ is uniformly bounded.

\paragraph{Entrance size.}
We first bound $g_{S_k}(D_{S_k})$ uniformly over all entrances.  At time
$S_k$, the path enters the strip in one round, so the one-step increment bound
controls $D_{S_k}+\beta S_k$.  By definition,
$D_{S_k-1}< -\beta(S_k-1)$ and $D_{S_k}\ge-\beta S_k$.  The upper increment
bound in Lemma~\ref{lem:gap-increments}, together with $D_{S_k-1}<0$, gives
\[
D_{S_k}-D_{S_k-1}
\le\hat\ell_{S_k-1,i_{S_k-1}^{\mathcal O}}\le d.
\]
Consequently,
\[
\begin{aligned}
0\le D_{S_k}+\beta S_k
&=(D_{S_k}-D_{S_k-1})
 +(D_{S_k-1}+\beta(S_k-1))+\beta\\
&\le d+\beta.
\end{aligned}
\]
The burn-in choice implies
$\beta S_k\ge\beta N\ge2(d+\beta)$, and hence
\[
D_{S_k}\le-\beta S_k+d+\beta
\le-\frac{\beta S_k}{2}<0.
\]
Substitution into the negative branch of $g_{S_k}$ gives
\[
g_{S_k}(D_{S_k})
=\frac{4S_k^2}{(2\sqrt{S_k}-\alpha D_{S_k})^2}
\le\frac{4S_k^2}{(\alpha\beta S_k/2)^2}
=\frac{16}{\alpha^2\beta^2}.
\]
Thus every entrance potential contributes the same factor of order
$\beta^{-2}$, and the remaining task is to bound the normalized power sum of
the entrance times.  For every $t\ge N$,
\[
\begin{aligned}
&\E\left[
\sum_{k:N<S_k\le t}
\left(\frac{S_k}{t}\right)^\gamma
g_{S_k}(D_{S_k})
\right]\\
&\qquad\le
\frac{16}{\alpha^2\beta^2}
\E\left[
\sum_{k:N<S_k\le t}
\left(\frac{S_k}{t}\right)^\gamma
\right].
\end{aligned}
\]

\paragraph{Reduction to dyadic entrance counts.}
The preceding pointwise bound leaves only the distribution of the entrance
times.  To apply the deterministic Lemma~\ref{lem:dyadic-power-sum}, we encode
the random entrance configuration by its expected mass at each calendar
time.  For each integer $r>N$, define
\[
a_r:=\E\left[\sum_{k\ge1}\1_{\{S_k=r\}}\right].
\]
Thus $a_r$ is the expected number of entrances at time $r$; since the
entrance times are distinct, it is also the probability that $r$ is an
entrance time.  Tonelli's theorem gives, for $x,t\ge N$,
\[
\sum_{x<r\le2x}a_r
=\E\#\{k:x<S_k\le2x\},
\]
and
\[
\frac1{t^\gamma}\sum_{N<r\le t}r^\gamma a_r
=\E\left[
\sum_{k:N<S_k\le t}\left(\frac{S_k}{t}\right)^\gamma
\right].
\]
The first identity is the mass in the doubling interval $(x,2x]$, whereas
the second is exactly the normalized power sum left by the entrance-size
reduction.  The upper bound in Lemma~\ref{lem:dyadic-power-sum}, with
$n_0=N$, therefore reduces the power-sum estimate to proving, uniformly for
$x\ge N$,
\[
\E\#\{k:x<S_k\le2x\}
\le C_\alpha(1+|\mathcal S|\beta^{-2}).
\]

\paragraph{Entrance frequency.}
We now reduce the required doubling-interval entrance count to one-step exits
from the strip.  Apart from the first entrance in $(x,2x]$, every entrance is
preceded by a distinct exit.  More precisely, if the entrances are
$r_1<\cdots<r_m$, then for every $q<m$ the path is inside the strip at $r_q$
and outside it at $r_{q+1}-1$.  Hence there is an exit at some time in
$\{r_q,\ldots,r_{q+1}-2\}$, and these time intervals are disjoint.  Therefore,
pathwise,
\[
\#\{k:x<S_k\le2x\}
\le1+\sum_{u=\lfloor x\rfloor}^{\lceil2x\rceil}
\1_{\{D_u\ge-\beta u,\ D_{u+1}< -\beta(u+1)\}}.
\]
Thus it is enough to prove that, for every $u\ge N$,
\[
\P\left(
D_u\ge-\beta u,\ D_{u+1}< -\beta(u+1)
\,\middle|\,\mathscr F_{u-1}
\right)
\le\frac{C_\alpha|\mathcal S|\beta^{-2}}{u}.
\]
Once established, this conditional bound, the tower property, and
$\sum_{u=\lfloor x\rfloor}^{\lceil2x\rceil}u^{-1}\le C$ give the required
uniform entrance count.

\paragraph{One-step exit probability.}
The preceding paragraph identifies the only probabilistic input still
needed.  We now prove this one-step estimate.  For $u\ge N$, define
\[
E_u^{\mathrm{out}}
:=\{D_u\ge-\beta u,\ D_{u+1}< -\beta(u+1)\}.
\]
Condition on $\mathscr F_{u-1}$, so that $D_u$ and $p_u$ are fixed.  On
$E_u^{\mathrm{out}}$, the sampled arm at round $u$ must be suboptimal.
Indeed, if $I_u\in\mathcal O$, then
$N_{u+1}^{\mathcal S}=N_u^{\mathcal S}$ and
$M_{u+1}^{\mathcal O}\ge M_u^{\mathcal O}$, whence
$D_{u+1}\ge D_u\ge-\beta u> -\beta(u+1)$.

We use different estimates depending on the distance from zero.  Far below
zero, the sampling-probability identity in \eq{pti} makes sampling any suboptimal arm unlikely; closer to
the boundary, an exit requires an unusually large one-round increase of
$N_u^{\mathcal S}$.

First suppose that $D_u\le-\beta u/2$.  Then $D_u<0$, so an optimal arm is a
global minimizer.  Hence, for every $i\in\mathcal S$,
\[
\underline{\hat L}_{u,i}
\ge N_u^{\mathcal S}-M_u^{\mathcal O}
=-D_u\ge\frac{\beta u}{2}.
\]
Substituting this bound and $\eta_u=\alpha/\sqrt u$ into
\eq{pti} gives
\[
p_{u,i}
\le4\left(\frac{\alpha}{\sqrt u}\frac{\beta u}{2}\right)^{-2}
=\frac{16}{\alpha^2\beta^2u}.
\]
Since an exit requires $I_u\in\mathcal S$, summing this bound over
$i\in\mathcal S$ controls the exit probability in this case.

Now suppose that $D_u>-\beta u/2$.  On $E_u^{\mathrm{out}}$,
\[
D_{u+1}-D_u
< -\beta(u+1)+\frac{\beta u}{2}
< -\frac{\beta u}{2}.
\]
The lower increment bound in Lemma~\ref{lem:gap-increments} therefore gives
$N_{u+1}^{\mathcal S}-N_u^{\mathcal S}>\beta u/2$.  Moreover,
\[
0\le N_{u+1}^{\mathcal S}-N_u^{\mathcal S}
\le\sum_{i\in\mathcal S}\hat\ell_{u,i},
\qquad
\E\left[\sum_{i\in\mathcal S}\hat\ell_{u,i}
\,\middle|\,\mathscr F_{u-1}\right]
=\sum_{i\in\mathcal S}\mu_i\le |\mathcal S|.
\]
Markov's inequality gives a conditional probability at most
$2|\mathcal S|/(\beta u)$.  Since $\beta\le\Delta\le1$, the two cases prove
\[
\P(E_u^{\mathrm{out}}\mid\mathscr F_{u-1})
\le\frac{C_\alpha|\mathcal S|\beta^{-2}}{u}.
\]

\paragraph{Completion.}
We now trace the preceding reductions backwards.  Substituting the one-step
exit probability into the pathwise entrance--exit comparison and using the
tower property gives
\[
\E\#\{k:x<S_k\le2x\}
\le C_\alpha(1+|\mathcal S|\beta^{-2}).
\]
The upper bound in Lemma~\ref{lem:dyadic-power-sum} and the two Tonelli
identities above then give
\[
\E\left[
\sum_{k:N<S_k\le t}\left(\frac{S_k}{t}\right)^\gamma
\right]
\le\frac{C_\alpha(1+|\mathcal S|\beta^{-2})}{1-2^{-\gamma}}.
\]
Since the fixed $\gamma>0$ depends only on $\alpha$, the factor
$(1-2^{-\gamma})^{-1}$ is absorbed into $C_\alpha$ and introduces no additional
dependence on $t$, $d$, $\Delta$, or $|\mathcal S|$.  Combining the resulting
bound with the pointwise entrance-size estimate proves
\eq{eq:weighted-entrance-potential}.
\end{proof}

\section{Local Lyapunov Estimates}
\label{sec:local-estimates}

This appendix proves the two one-block estimates used by the global Lyapunov
argument.  The first contracts the potential at the planned endpoint and is
iterated along the skeleton times.  The second controls the maximum over the
complete planned block and is used to interpolate to an arbitrary deterministic
time.  Each proof treats positive and negative starting points separately, but
the resulting statements hold on the entire strip $D_t\ge-\beta t$.

The parameters and block length are fixed in Appendix~\ref{sec:parameters}.
The estimates for $N_t^{\mathcal S}$ and $M_t^{\mathcal O}$ are proved in
Appendices~\ref{sec:suboptimal-minimum} and~\ref{sec:optimal-minimum}, and the
additional algebraic and maximal estimates are collected in
Appendix~\ref{sec:auxiliary-estimates}.  All constraints on the constants are
verified in Appendix~\ref{sec:parameter-verification}.
\subsection{One-block contraction}

The endpoint estimate is obtained by controlling the Lyapunov ratio over one
planned block.  We first state the estimate for every starting point satisfying
$D_t\ge-\beta t$ and then prove it separately for positive and negative
starting points.

The restriction to this strip is motivated by the mean-field calculation in
Section~\ref{sec:lyapunov-overview}.  Suppose that $D_t<0$ and temporarily
replace the random endpoint after the planned block by its nominal gap drift
$D_t-\Delta h_t$.  The negative branch gives the benchmark ratio
\[
\frac{g_{t+h_t}(D_t-\Delta h_t)}{g_t(D_t)}
=\left(1+\frac{h_t}{t}\right)^2
\left(
\frac{2\sqrt t-\alpha D_t}
{2\sqrt{t+h_t}-\alpha D_t+\alpha\Delta h_t}
\right)^2.
\]
When $-D_t\gg\sqrt t$ and $h_t$ is short relative to both $t$ and
$-D_t/\Delta$, its first-order form is
\[
\frac{g_{t+h_t}(D_t-\Delta h_t)}{g_t(D_t)}
\approx 1+2h_t\left(\frac1t+\frac{\Delta}{D_t}\right).
\]
Because $D_t<0$, this calculation predicts a decrease only above the balance
point $D_t=-\Delta t$.  The parameter choices make $\beta/\Delta$ sufficiently
small, so the formal strip lies safely above this mean-field threshold.  The
proof must still retain the square-root term and control the two moving minima,
possible crossings of zero, and the importance-weighted fluctuations; the
same parameter conditions keep the planned block short enough to absorb these
effects into the gap drift.

\begin{Lem}[One-block contraction]
\label{lem:one-block-contraction}
For every $t\ge N$ such that $D_t\ge-\beta t$,
\begin{equation}
\E\left[
g_{t+h_t}(D_{t+h_t})
\,\middle|\,\mathscr F_{t-1}
\right]
\le\left(1-\frac{h_t}{C_\alpha t}\right)g_t(D_t).
\label{eq:one-block-contraction}
\end{equation}
\end{Lem}

\begin{proof}
We control the endpoint ratio first for $D_t\ge0$ and then for
$D_t\in[-\beta t,0)$.  To express the endpoint increment in both cases,
use the same algebraic block decomposition as \citet{zhan2026does} and define
\begin{align*}
R_t&:=M_{t+h_t}^{\mathcal O}-M_t^{\mathcal O}-\mu_*h_t,\\
T_t&:=(\mu_*+\Delta)h_t-
\bigl(N_{t+h_t}^{\mathcal S}-N_t^{\mathcal S}\bigr),\\
U_t&:=\bigl[\Delta(t+h_t)+D_{t+h_t}\bigr]
-\bigl[\Delta t+D_t\bigr]=R_t+T_t.
\end{align*}
Thus $R_t$ is the moving optimal-set minimum increment relative to
$\mu_*h_t$, $T_t$ is the signed suboptimal-side shortfall, and $U_t$ is the
exact increment of $D_t+\Delta t$ over the block.

\subsubsection{Positive starting points}
Suppose that $D_t\ge0$.
\medskip\par\noindent\textbf{Reduction to the ratio.}\quad
We first control
$g_{t+h_t}(D_{t+h_t})/g_t(D_t)$.  Although $D_t\ge0$, the endpoint
$D_{t+h_t}$ may be negative, so we first bound the endpoint potential by the
same quadratic expression on both branches.  Put $x=D_t+\Delta t$.
The definitions give the pathwise comparison
\begin{equation}
\label{eq:positive-pathwise}
D_{t+h_t}+\Delta(t+h_t)
=x+U_t=x+R_t+T_t\le x+R_t+T_t^+.
\end{equation}
Here $D_t\ge0$ gives $h_t=\lceil Ad\rceil$.  To control the endpoint branch,
we first use a moving-gap lower bound, then place that lower bound in the
range of Lemma~\ref{lem:lyapunov-branch-comparison}.  Appendix~
\ref{sec:parameter-verification} verifies the three corresponding conditions:
\begin{equation}
\label{eq:positive-parameter-conditions}
t\ge(12dh_t)^2,
\qquad
4dh_t+\frac{2\sqrt t}{\alpha}\le\Delta t,
\qquad
\alpha\Delta\sqrt t\ge2.
\end{equation}
Since $D_t\ge0$, the bound $t\ge(12dh_t)^2$ and
Lemma~\ref{lem:gap-increments} give $D_{t+h_t}\ge-4dh_t$.  Since
$4dh_t+2\sqrt t/\alpha\le\Delta t$ and
$u\mapsto\Delta u-2\sqrt u/\alpha$ is increasing for $u\ge t$, it follows
that
\[
D_{t+h_t}\ge \frac{2\sqrt{t+h_t}}{\alpha}-\Delta(t+h_t).
\]
If $D_{t+h_t}\ge0$, the quadratic branch is the definition of
$g_{t+h_t}(D_{t+h_t})$.  If $D_{t+h_t}<0$, then
$\alpha\Delta\sqrt{t+h_t}\ge2$ because
$\alpha\Delta\sqrt t\ge2$, so the endpoint lower bound and
Lemma~\ref{lem:lyapunov-branch-comparison}(2), applied at time $t+h_t$,
give
\[
g_{t+h_t}(D_{t+h_t})
\le\frac{(D_{t+h_t}+\Delta(t+h_t))^2}{\Delta^2(t+h_t)}.
\]
Moreover,
\[
x+U_t=D_{t+h_t}+\Delta(t+h_t)
\ge\Delta(t+h_t)-4dh_t>0.
\]
Since $T_t^+\ge T_t$, \eq{eq:positive-pathwise} and the
positivity just established imply
\[
g_{t+h_t}(D_{t+h_t})
\le\frac{(x+R_t+T_t^+)^2}{\Delta^2(t+h_t)}.
\]
Because $g_t(D_t)=x^2/(\Delta^2t)$ and $x$ is
$\mathscr F_{t-1}$-measurable, expanding the square gives the analogue of
\citet[Eq.~(18)]{zhan2026does}:
\begin{samepage}
\begin{align}
&\frac{\E[g_{t+h_t}(D_{t+h_t})\mid\mathscr F_{t-1}]}{g_t(D_t)}
\notag\\
&\quad\le
\left(1+\frac{h_t}{t}\right)^{-1}
\Bigg[
1+\frac{2\E[R_t\mid\mathscr F_{t-1}]
+2\E[T_t^+\mid\mathscr F_{t-1}]}{x}
\notag\\
&\qquad\qquad+
\frac{\E[R_t^2\mid\mathscr F_{t-1}]
+\E[(T_t^+)^2\mid\mathscr F_{t-1}]
+2\E[R_tT_t^+\mid\mathscr F_{t-1}]}{x^2}
\Bigg].
\label{eq:positive-ratio-calculation}
\end{align}
\end{samepage}
Thus the ratio requires the first moments of $R_t$ and $T_t^+$, their
second moments, and the mixed term $\E[R_tT_t^+\mid\mathscr F_{t-1}]$.
We now bound the five terms displayed in
\eq{eq:positive-ratio-calculation}.

\medskip\par\noindent\textbf{Moment inputs.}\quad
To absorb the contribution of $\E[R_t^2\mid\mathscr F_{t-1}]$ in the
quadratic expansion, choose a sufficiently small fixed $\rho>0$.
Appendix~\ref{sec:parameter-verification} verifies
$t\ge A^2d$, $t\ge C_\alpha dh_t/\Delta^2$, and
$h_t\le\min\{t/8,\Delta t\}$, which include the parameter conditions used
in the following moment estimates.  By
Corollary~\ref{cor:positive-optimal-increment} and
Corollary~\ref{cor:state-dependent-horizon} with $\varepsilon=1/16$,
\[
\begin{aligned}
\E[R_t\mid\mathscr F_{t-1}]&\le0,
&\qquad
\E[R_t^2\mid\mathscr F_{t-1}]
&\le\mu_*\left(\frac{\alpha^2}{4}+\frac\rho4\right)
\frac{h_tx^2}{t},\\
\E[T_t^+\mid\mathscr F_{t-1}]&\le\frac{\Delta h_t}{16},
&
\E[(T_t^+)^2\mid\mathscr F_{t-1}]
&\le \frac{C\log(2|\mathcal S|)}{A}h_t^2.
\end{aligned}
\]

\medskip\par\noindent\textbf{Contraction.}\quad
Since $x\ge\Delta t$ and
$\E[R_t\mid\mathscr F_{t-1}]\le0$,
\[
\frac{2\E[R_t+T_t^+\mid\mathscr F_{t-1}]}{x}
\le\frac{2\E[T_t^+\mid\mathscr F_{t-1}]}{x}
\le\frac{h_t}{8t}.
\]
Choose also a fixed $\delta>0$, decreasing $\rho$ if necessary, so that
$(1+\delta)\mu_*(\alpha^2/4+\rho/4)\le1/3$.  Weighted Young's inequality gives
\[
\begin{aligned}
\frac{\E[(R_t+T_t^+)^2\mid\mathscr F_{t-1}]}{x^2}
&\le(1+\delta)
\frac{\E[R_t^2\mid\mathscr F_{t-1}]}{x^2}
+(1+\delta^{-1})
\frac{\E[(T_t^+)^2\mid\mathscr F_{t-1}]}{x^2}\\
&\le\frac{h_t}{3t}
+\frac{C\log(2|\mathcal S|)}{A}
\frac{h_t^2}{\Delta^2t^2}.
\end{aligned}
\]
Appendix~\ref{sec:parameter-verification} verifies
\[
\frac{C\log(2|\mathcal S|)}{A}
\frac{h_t^2}{\Delta^2t^2}\le\frac{h_t}{24t}.
\]
Substituting these first- and second-moment bounds into
\eq{eq:positive-ratio-calculation} yields
\[
\frac{\E[g_{t+h_t}(D_{t+h_t})\mid\mathscr F_{t-1}]}{g_t(D_t)}
\le\frac{1+h_t/(2t)}{1+h_t/t}
=1-\frac{h_t}{2(t+h_t)}
\le1-\frac{h_t}{16t}.
\]
After increasing $C_\alpha$ if necessary, the last display proves
\eq{eq:one-block-contraction} for $D_t\ge0$.

\subsubsection{Negative starting points}
Suppose that $D_t\in[-\beta t,0)$.  Appendix~
\ref{sec:parameter-verification} verifies
\begin{equation}
\label{eq:adaptive-block-scales}
Ad\le h_t\le C_\alpha A(d+\beta^2t)\le\frac t8.
\end{equation}

\medskip\par\noindent\textbf{Reduction to the ratio and path decomposition.}\quad
We begin by controlling
$g_{t+h_t}(D_{t+h_t})/g_t(D_t)$.  Since $D_t<0$, the denominator is given by
the inverse-square branch.  The endpoint $D_{t+h_t}$ may lie on either
branch, so we first obtain an inverse-square upper bound for the endpoint
potential that covers both possibilities.  Conditioning on
$\mathscr F_{t-1}$ fixes the $\mathscr F_{t-1}$-measurable endpoint
$t+h_t$.

The block decomposition gives
\[
D_{t+h_t}-D_t=U_t-\Delta h_t=R_t+T_t-\Delta h_t.
\]
Although $D_t<0$, the path $D_u$ may enter the positive half-line during
$[t,t+h_t]$.  Define the first time at which $D_u$ reaches the positive value
$\sqrt t/\alpha$ by
\[
\kappa_t:=\inf\left\{u\ge t:D_u\ge\frac{\sqrt t}{\alpha}\right\}.
\]
Before the stopping time,
\(D_u<\sqrt t/\alpha<2\sqrt u/\alpha\).  If $D_u\ge0$, then
$2\sqrt u-\alpha D_u\ge\sqrt t>0$ and Appendix~
\ref{sec:parameter-verification} gives
\(\alpha\Delta\sqrt u\ge2\), so
Lemma~\ref{lem:lyapunov-branch-comparison}(1) applies.  If $D_u<0$, the
identity $g_u(D_u)=4u^2/(2\sqrt u-\alpha D_u)^2$ holds by definition.
Hence
\begin{equation}
\label{eq:positive-threshold-branch-comparison}
g_u(D_u)\le\frac{4u^2}{(2\sqrt u-\alpha D_u)^2},
\qquad t\le u\le t+h_t,\quad u<\kappa_t.
\end{equation}
On every path with $\kappa_t>t+h_t$,
$D_{t+h_t}<\sqrt t/\alpha$, and hence
$2\sqrt{t+h_t}-\alpha D_{t+h_t}>\sqrt t>0$.  Since $D_t<0$, the
inverse-square definition also gives
\[
2\sqrt t-\alpha D_t=\frac{2t}{\sqrt{g_t(D_t)}}.
\]
The endpoint denominator therefore satisfies
\[
\begin{aligned}
2\sqrt{t+h_t}-\alpha D_{t+h_t}
&=\frac{2t}{\sqrt{g_t(D_t)}}
+2(\sqrt{t+h_t}-\sqrt t)
-\alpha(D_{t+h_t}-D_t).
\end{aligned}
\]
Using \eq{eq:positive-threshold-branch-comparison} at the endpoint gives
\begin{equation}
\label{eq:negative-ratio-calculation}
\frac{g_{t+h_t}(D_{t+h_t})}{g_t(D_t)}
\le(1+h_t/t)^2
\left(
1-\frac{\sqrt{g_t(D_t)}}{2t}
\left[
\alpha(D_{t+h_t}-D_t)
-2(\sqrt{t+h_t}-\sqrt t)
\right]
\right)^{-2}.
\end{equation}
To control the inverse-square factor in \eq{eq:negative-ratio-calculation},
we use the elementary inequality
\[
(1-v)^{-2}\le1+2v+8v^2,
\qquad v\le\frac14.
\]
The normalized change in \eq{eq:negative-ratio-calculation} is increasing
in $D_{t+h_t}-D_t$.  Since
$D_{t+h_t}-D_t=U_t-\Delta h_t$,
\[
D_{t+h_t}-D_t
\le U_t^+-\Delta h_t
\le R_t^++T_t^+-\Delta h_t.
\]
We therefore define the following upper bound on the normalized change:
\[
v_t:=\frac{\sqrt{g_t(D_t)}}{2t}
\left[
\alpha(U_t^+-\Delta h_t)
-2(\sqrt{t+h_t}-\sqrt t)
\right].
\]
On paths with $\kappa_t>t+h_t$ and $v_t\le1/4$, the elementary inequality
controls the ratio.  When $\kappa_t>t+h_t$ and $v_t>1/4$, we instead use a
one-sided tail bound.  When $\kappa_t\le t+h_t$, the inverse-square ratio is
replaced by a fixed-arm continuation from the first time that $D_u$ reaches
$\sqrt t/\alpha$.  These alternatives give the following three disjoint
classes:
\[
\begin{aligned}
&\kappa_t>t+h_t,\qquad v_t\le\frac14;\\
&\kappa_t>t+h_t,\qquad v_t>\frac14;\\
&\kappa_t\le t+h_t.
\end{aligned}
\]
We bound the contribution of each class and then add the three bounds.

\medskip\par\noindent\textbf{Paths with $\kappa_t>t+h_t$ and $v_t\le1/4$.}\quad
The upper bound on $D_{t+h_t}-D_t$ gives
\[
\begin{aligned}
2\sqrt{t+h_t}-\alpha D_{t+h_t}
&\ge\frac{2t}{\sqrt{g_t(D_t)}}
\left(1-v_t\right)
\ge\frac{3t}{2\sqrt{g_t(D_t)}}>0.
\end{aligned}
\]
Since $v\mapsto(1-v)^{-2}$ is increasing for $v<1$,
\eq{eq:negative-ratio-calculation} and
$D_{t+h_t}-D_t\le U_t^+-\Delta h_t$ give
\[
\frac{g_{t+h_t}(D_{t+h_t})}{g_t(D_t)}
\le(1+h_t/t)^2
\left(1-v_t\right)^{-2}.
\]
After multiplying this ratio bound by
$\1_{\{\kappa_t>t+h_t,\ v_t\le1/4\}}$ and taking conditional expectations,
the inequality $(1-v_t)^{-2}\le1+2v_t+8v_t^2$ shows that it remains to control
\[
\E\left[v_t\1_{\{\kappa_t>t+h_t,\ v_t\le1/4\}}
\,\middle|\,\mathscr F_{t-1}\right]
\quad\text{and}\quad
\E\left[v_t^2\1_{\{\kappa_t>t+h_t,\ v_t\le1/4\}}
\,\middle|\,\mathscr F_{t-1}\right].
\]
By the definition of $v_t$, these two quantities require a negative first
moment and a second-moment bound for $U_t^+-\Delta h_t$ on
$\{\kappa_t>t+h_t\}$.  The pathwise inequality
$U_t^+\le R_t^++T_t^+$ gives
\[
\begin{aligned}
&\E\left[(U_t^+-\Delta h_t)\1_{\{\kappa_t>t+h_t\}}
\,\middle|\,\mathscr F_{t-1}\right]\\
&\quad\le
\E\left[R_t^+\1_{\{\kappa_t>t+h_t\}}
\,\middle|\,\mathscr F_{t-1}\right]
+\E[T_t^+\mid\mathscr F_{t-1}]\\
&\qquad
-\Delta h_t\P(\kappa_t>t+h_t\mid\mathscr F_{t-1}).
\end{aligned}
\]
Thus the linear term requires first-moment bounds for $R_t^+$ and $T_t^+$
and a lower bound on the probability that $D_u$ does not reach
$\sqrt t/\alpha$.  The quadratic term requires the corresponding second
moments.  We now establish these inputs before substituting them into the
ratio.

\smallskip\noindent\emph{Linear input.}
Corollary~\ref{cor:state-dependent-horizon}, applied with
\(\varepsilon=1/4\), gives
\[
0\le T_t^+\le h_t,
\qquad
\E[T_t^+\mid\mathscr F_{t-1}]\le\frac{\Delta h_t}{4}.
\]

To retain the negative term
$-\Delta h_t\P(\kappa_t>t+h_t\mid\mathscr F_{t-1})$, we need the
probability in this term to be bounded away from zero.  The
negative-starting-point proof of Lemma~\ref{lem:one-block-maximal} below
establishes \eq{eq:positive-threshold-hitting-probability}, and
Appendix~\ref{sec:parameter-verification} verifies that its exponential
upper bound is at most $1/4$.  Hence,
\[
\P(\kappa_t>t+h_t\mid\mathscr F_{t-1})\ge\frac34.
\]

With $L=h_t$, Lemma~\ref{lem:stopped-optimal-set-increment} gives the
required first moment of the endpoint increment $R_t$ on
$\{\kappa_t>t+h_t\}$:
\[
\E\left[R_t^+\1_{\{\kappa_t>t+h_t\}}
\,\middle|\,\mathscr F_{t-1}\right]\le\sqrt{8dh_t}.
\]
Appendix~\ref{sec:parameter-verification} also gives
\[
\sqrt{8dh_t}\le\frac{\Delta h_t}{4}.
\]

In the upper bound for
$\E[(U_t^+-\Delta h_t)\1_{\{\kappa_t>t+h_t\}}\mid\mathscr F_{t-1}]$,
the $R_t^+$ and $T_t^+$ terms are each at most $\Delta h_t/4$, while
the preceding probability bound gives
$\P(\kappa_t>t+h_t\mid\mathscr F_{t-1})\ge3/4$.  Therefore,
\begin{align*}
&\E\left[(U_t^+-\Delta h_t)\1_{\{\kappa_t>t+h_t\}}
\,\middle|\,\mathscr F_{t-1}\right]\\
&\quad\le
\E\left[R_t^+\1_{\{\kappa_t>t+h_t\}}
\,\middle|\,\mathscr F_{t-1}\right]
+\E[T_t^+\mid\mathscr F_{t-1}]\\
&\qquad
-\Delta h_t\P(\kappa_t>t+h_t\mid\mathscr F_{t-1})
\le-\frac{\Delta h_t}{4}.
\end{align*}
On $\{\kappa_t>t+h_t,\ v_t>1/4\}$, the variable $v_t$ is positive.
Therefore, the first-moment bound for $U_t^+-\Delta h_t$ gives
\[
\begin{aligned}
&\E\left[
v_t\1_{\{\kappa_t>t+h_t,\ v_t\le1/4\}}
\,\middle|\,\mathscr F_{t-1}
\right]\\
&\quad\le
\E\left[v_t\1_{\{\kappa_t>t+h_t\}}
\,\middle|\,\mathscr F_{t-1}\right]\\
&\quad\le
\frac{\alpha\sqrt{g_t(D_t)}}{2t}
\E\left[(U_t^+-\Delta h_t)\1_{\{\kappa_t>t+h_t\}}
\,\middle|\,\mathscr F_{t-1}\right]
\le-\frac{\alpha\Delta h_t\sqrt{g_t(D_t)}}{8t}.
\end{aligned}
\]
The second inequality discards the nonpositive term
\[
-\frac{\sqrt{g_t(D_t)}}{t}(\sqrt{t+h_t}-\sqrt t)
\P(\kappa_t>t+h_t\mid\mathscr F_{t-1})
\]
from the conditional expectation of $v_t$.

\smallskip\noindent\emph{Quadratic input.}
Corollary~\ref{cor:state-dependent-horizon} and
Lemma~\ref{lem:stopped-optimal-set-increment}, again with
$\varepsilon=1/4$ and $L=h_t$, give
\[
\E[(T_t^+)^2\mid\mathscr F_{t-1}]
\le\frac{C\log(2|\mathcal S|)}A h_t^2,
\qquad
\E\left[(R_t^+)^2\1_{\{\kappa_t>t+h_t\}}
\,\middle|\,\mathscr F_{t-1}\right]\le8dh_t.
\]
Since
\[
(U_t^+-\Delta h_t)^2
\le4(R_t^+)^2+4(T_t^+)^2+2\Delta^2h_t^2,
\]
these two estimates imply
\[
\E\left[(U_t^+-\Delta h_t)^2
\1_{\{\kappa_t>t+h_t\}}
\,\middle|\,\mathscr F_{t-1}\right]
\le C\left(
dh_t+\frac{\log(2|\mathcal S|)}A h_t^2+\Delta^2h_t^2
\right).
\]
Moreover, $D_t<0$ gives $g_t(D_t)\le t$.  The conditional second-moment bound
for $U_t^+-\Delta h_t$, the inequality
$\sqrt{t+h_t}-\sqrt t\le h_t/(2\sqrt t)$, and
$h_t\le t/8$ from \eq{eq:adaptive-block-scales} imply
\[
\begin{aligned}
&\E\left[v_t^2\1_{\{\kappa_t>t+h_t\}}
\,\middle|\,\mathscr F_{t-1}\right]\\
&\qquad\le C_\alpha\left[
\frac{h_t}{t}+
\frac{g_t(D_t)}{t^2}\left(
dh_t+\frac{\log(2|\mathcal S|)}A h_t^2+\Delta^2h_t^2
\right)
\right].
\end{aligned}
\]
Insert the linear and quadratic conditional-moment bounds into
$(1-v_t)^{-2}\le1+2v_t+8v_t^2$, and then use
\eq{eq:adaptive-block-scales} to bound
$(1+h_t/t)^2\le1+Ch_t/t$ in the ratio.  This gives
\[
\begin{aligned}
&\frac{1}{g_t(D_t)}\E\left[
g_{t+h_t}(D_{t+h_t})
\1_{\{\kappa_t>t+h_t,\ v_t\le1/4\}}
\,\middle|\,\mathscr F_{t-1}
\right]\\
&\quad\le1-
\frac{\alpha\Delta h_t\sqrt{g_t(D_t)}}{4t}
+C_\alpha\left[
\frac{h_t}{t}+
\frac{g_t(D_t)}{t^2}\left(
dh_t+\frac{\log(2|\mathcal S|)}A h_t^2+\Delta^2h_t^2
\right)
\right].
\end{aligned}
\]
Appendix~\ref{sec:parameter-verification} verifies
\begin{equation}
\label{eq:negative-normalized-remainders}
C_\alpha\left[
\frac{h_t}{t}+
\frac{g_t(D_t)}{t^2}\left(
dh_t+\frac{\log(2|\mathcal S|)}A h_t^2+\Delta^2h_t^2
\right)
\right]
\le\frac{\alpha\Delta h_t\sqrt{g_t(D_t)}}{8t}.
\end{equation}
The right-hand side of \eq{eq:negative-normalized-remainders} absorbs the
$C_\alpha[\cdots]$ remainder in the conditional ratio bound for
$\{\kappa_t>t+h_t,\ v_t\le1/4\}$.  Using also
$\Delta\sqrt{g_t(D_t)}\ge2/3$ from
Appendix~\ref{sec:parameter-verification} gives
\begin{align}
\begin{aligned}
&\E\left[
g_{t+h_t}(D_{t+h_t})
\1_{\{\kappa_t>t+h_t,\ v_t\le1/4\}}
\,\middle|\,\mathscr F_{t-1}\right]\\
&\qquad\le
\left(1-\frac{\alpha\Delta h_t\sqrt{g_t(D_t)}}{8t}\right)g_t(D_t)
\le\left(1-\frac{2h_t}{C_\alpha t}\right)g_t(D_t).
\end{aligned}
\label{eq:negative-no-hit-good}
\end{align}

\medskip\par\noindent\textbf{Paths with $\kappa_t>t+h_t$ and $v_t>1/4$.}\quad
The bound $(1-v)^{-2}\le1+2v+8v^2$ used above only applies when
$v\le1/4$, so it does not control paths with $v_t>1/4$.  We show instead
that this event forces a large positive value of $R_t$; a one-sided tail
bound then shows that these paths have small conditional probability.
Since \(U_t^+\le R_t^++T_t^+\), \(T_t^+\le h_t\), and Appendix~
\ref{sec:parameter-verification} gives
$h_t\sqrt{g_t(D_t)}/t\le1/8$, the inequality $v_t>1/4$ implies
\[
R_t^+\ge
\frac{t}{2\alpha\sqrt{g_t(D_t)}}-(1-\Delta)h_t
\ge\frac{3t}{8\sqrt{g_t(D_t)}}.
\]
The maximal Freedman bound in Lemma~\ref{lem:negative-persistence} supplies
the required upper tail: for every \(x\ge0\),
\begin{equation}
\label{eq:negative-R-tail}
\P\left(
R_t^+\ge x,\ \kappa_t>t+h_t
\,\middle|\,\mathscr F_{t-1}\right)
\le\exp\left\{-\frac{x^2}{4dh_t+4dx/3}\right\}.
\end{equation}
Applying \eq{eq:negative-R-tail} with
$x=3t/(8\sqrt{g_t(D_t)})$ gives
\[
\P\left(
\kappa_t>t+h_t,\ v_t>\frac14
\,\middle|\,\mathscr F_{t-1}\right)
\le\exp\left\{-\frac{t}
{C_\alpha d\sqrt{g_t(D_t)}}\right\}.
\]
On every path with $\kappa_t>t+h_t$, Eqs.~
\eqref{eq:adaptive-block-scales} and~
\eqref{eq:positive-threshold-branch-comparison} give
\[
g_{t+h_t}(D_{t+h_t})
\le \frac{4(t+h_t)^2}
{(2\sqrt{t+h_t}-\alpha D_{t+h_t})^2}
\le \frac{4(t+h_t)^2}{t}
\le C t.
\]
Thus the contribution of the paths
with $\kappa_t>t+h_t$ and $v_t>1/4$ is at most
\[
C_\alpha t\exp\left\{-\frac{t}
{C_\alpha d\sqrt{g_t(D_t)}}\right\}.
\]
Appendix~\ref{sec:parameter-verification} converts this exponential payment
to the scale of the one-block contraction:
\begin{equation}
\label{eq:exceptional-payment-absorption}
C_\alpha t\exp\left\{-\frac{t}
{C_\alpha d\sqrt{g_t(D_t)}}\right\}
\le\frac{h_t}{8C_\alpha t}g_t(D_t).
\end{equation}
Hence the paths with $\kappa_t>t+h_t$ and $v_t>1/4$ contribute at most
$h_tg_t(D_t)/(8C_\alpha t)$.

\begin{samepage}
\medskip\par\noindent\textbf{Paths on which $D_u$ reaches $\sqrt t/\alpha$.}\quad
On $\{\kappa_t\le t+h_t\}$,
\eq{eq:positive-threshold-branch-comparison} only controls
$g_u(D_u)$ before $\kappa_t$, so it cannot be applied directly at
$t+h_t$.  We instead use the post-hitting maximal estimate in
\eq{eq:post-positive-threshold-payment}, proved below in the
negative-starting-point case of Lemma~\ref{lem:one-block-maximal}.
That estimate combines the hitting-probability bound in
\eq{eq:positive-threshold-hitting-probability} with
Lemma~\ref{lem:fixed-arm-maximal} and gives
\[
\begin{aligned}
&\E\left[
g_{t+h_t}(D_{t+h_t})
\1_{\{\kappa_t\le t+h_t\}}
\,\middle|\,\mathscr F_{t-1}\right]\\
&\quad\le
\E\left[
\1_{\{\kappa_t\le t+h_t\}}
\max_{\kappa_t\le u\le t+h_t}g_u(D_u)
\,\middle|\,\mathscr F_{t-1}\right]\\
&\quad\le
C_\alpha t\exp\left\{-\frac{t}
{C_\alpha d\sqrt{g_t(D_t)}}\right\}.
\end{aligned}
\]
The first inequality holds because $t+h_t\in[\kappa_t,t+h_t]$ on
$\{\kappa_t\le t+h_t\}$.  Equation~
\eqref{eq:exceptional-payment-absorption} therefore bounds this contribution
by $h_tg_t(D_t)/(8C_\alpha t)$.
\end{samepage}

\medskip\par\noindent\textbf{Combining the three path classes.}\quad
The events $\{\kappa_t>t+h_t,\ v_t\le1/4\}$,
$\{\kappa_t>t+h_t,\ v_t>1/4\}$, and $\{\kappa_t\le t+h_t\}$ are disjoint and
exhaust all possibilities.  Equation~\eqref{eq:negative-no-hit-good} controls
the first event, while the conclusions of the preceding two paragraphs bound
each of the other two contributions by
$h_tg_t(D_t)/(8C_\alpha t)$.  Adding the three contributions gives
\[
\begin{aligned}
&\E\left[g_{t+h_t}(D_{t+h_t})\mid\mathscr F_{t-1}\right]\\
&\quad\le
\left(1-\frac{2h_t}{C_\alpha t}\right)g_t(D_t)
+2\frac{h_t}{8C_\alpha t}g_t(D_t)
\le
\left(1-\frac{h_t}{C_\alpha t}\right)g_t(D_t).
\end{aligned}
\]
This proves the negative-starting-point case of
\eq{eq:one-block-contraction} and completes the proof.
\end{proof}

\subsection{One-block maximal estimate}

The global argument also needs to control the potential between consecutive
skeleton times.  The next estimate covers the complete planned block for every
starting point in the strip.

\begin{Lem}[One-block maximal estimate]
\label{lem:one-block-maximal}
For every $t\ge N$ such that $D_t\ge-\beta t$,
\begin{equation}
\E\left[
\max_{t\le u\le t+h_t}g_u(D_u)
\,\middle|\,\mathscr F_{t-1}
\right]
\le C_\alpha g_t(D_t).
\label{eq:one-block-maximal}
\end{equation}
\end{Lem}

\begin{proof}
To prove \eq{eq:one-block-maximal}, we first establish a common estimate
that controls the maximum of $g_u(D_u)$ over an interval starting from a
nonnegative state.  We then distinguish the cases $D_t\ge0$ and $D_t<0$.

\subsubsection{A common estimate from a nonnegative state}

Let $s\ge1$, let $v\ge s$ be a finite integer-valued
$\mathscr F_{s-1}$-measurable random variable, and use the predictable
minimizer
\[
i_s^{\mathcal O}\in\argmin_{a\in\mathcal O}\hat L_{s,a}
\]
as a single fixed comparator on $[s,v]$.
On $\{D_s\ge0\}$,
\[
\underline{\hat L}_{s,i_s^{\mathcal O}}=D_s,
\qquad
D_u^+\le\underline{\hat L}_{u,i_s^{\mathcal O}},
\quad s\le u\le v.
\]
Indeed, the second inequality is immediate when $D_u\le0$; when $D_u>0$,
$M_u=N_u^{\mathcal S}$ and
$M_u^{\mathcal O}\le\hat L_{u,i_s^{\mathcal O}}$.
Lemma~\ref{lem:lyapunov-branch-comparison}(3), applied at time $u$, gives
\[
g_u(y)\le2u+\frac{2(y^+)^2}{\Delta^2u}.
\]
Combining this inequality with the bound on $D_u^+$ gives, on
$\{D_s\ge0\}$ and for every
$s\le u\le v$,
\[
g_u(D_u)\le
2v+\frac{2}{\Delta^2s}
\underline{\hat L}_{u,i_s^{\mathcal O}}^2.
\]
Taking the maximum over $u$, then taking the conditional expectation and
applying Lemma~\ref{lem:fixed-arm-maximal}, gives
\begin{equation}
\label{eq:nonnegative-state-maximal}
\begin{aligned}
&\1_{\{D_s\ge0\}}\E\left[
\max_{s\le u\le v}g_u(D_u)
\,\middle|\,\mathscr F_{s-1}
\right]\\
&\quad\le C_\alpha\1_{\{D_s\ge0\}}\left[
v+\frac1{\Delta^2s}\left\{
\left(\frac vs\right)^{\alpha^2}D_s^2
+d(v-s)+\min\{dv,(v-s)^2\}
\right\}
\right].
\end{aligned}
\end{equation}

\subsubsection{Positive starting points}
Suppose that $D_t\ge0$.  Apply
\eq{eq:nonnegative-state-maximal} with $s=t$ and $v=t+h_t$.
Since $h_t\le t/8$ and $h_t\ge d$, it gives
\[
\E\left[\max_{t\le u\le t+h_t}g_u(D_u)
\,\middle|\,\mathscr F_{t-1}\right]
\le C_\alpha\left(
t+\frac{D_t^2}{\Delta^2t}
+\frac{h_t^2}{\Delta^2t}
\right).
\]
Finally, $h_t\le\Delta t$ and $D_t\ge0$ give
$h_t^2/(\Delta^2t)\le t$ and
$g_t(D_t)\ge t+D_t^2/(\Delta^2t)$.  This proves
\eq{eq:one-block-maximal} for $D_t\ge0$.

\subsubsection{Negative starting points}
Suppose that $D_t\in[-\beta t,0)$.  To control the maximum over the block,
as in the proof of Lemma~\ref{lem:one-block-contraction}, we split the path
when $D_u$ first reaches $\sqrt t/\alpha$.  Before that
time, \eq{eq:positive-threshold-branch-comparison} reduces the potential
to an inverse-square expression whose denominator is controlled by a stopped
optimal-set fluctuation.  After that time, the fixed-comparator estimate in
\eq{eq:nonnegative-state-maximal} controls the remaining part of the
block.  Accordingly, recall that
\[
\kappa_t:=\inf\left\{u\ge t:D_u\ge\frac{\sqrt t}{\alpha}\right\}.
\]

\medskip\par\noindent\textbf{After $D_u$ reaches $\sqrt t/\alpha$.}\quad
As noted above, we first use $\kappa_t$ as the starting time in
\eq{eq:nonnegative-state-maximal} to control $g_u(D_u)$ after
$D_u$ reaches $\sqrt t/\alpha$.  For each
$s\in\{t+1,\ldots,t+h_t\}$, the event $\{\kappa_t=s\}$ belongs to
$\mathscr F_{s-1}$.  On this event, $D_{s-1}<\sqrt t/\alpha$, and the
lower sampling-probability bound in Lemma~\ref{4-2} gives
$p_{s-1,i_{s-1}^{\mathcal O}}^{-1}\le(\sqrt d+1/2)^2\le2d$.
Lemma~\ref{lem:gap-increments} therefore gives
\[
D_s-D_{s-1}
\le\hat\ell_{s-1,i_{s-1}^{\mathcal O}}
\le p_{s-1,i_{s-1}^{\mathcal O}}^{-1}
\le2d,
\]
and hence
\[
\frac{\sqrt t}{\alpha}\le D_s
\le\frac{\sqrt t}{\alpha}+2d.
\]
Apply \eq{eq:nonnegative-state-maximal} with starting time $s$ and
endpoint $v=t+h_t$, and multiply it by $\1_{\{\kappa_t=s\}}$.
The endpoint $t+h_t$ is
$\mathscr F_{s-1}$-measurable and is at least $s$; moreover,
$\{\kappa_t=s\}\subseteq\{D_s\ge0\}$.  Therefore,
\[
\begin{aligned}
&\1_{\{\kappa_t=s\}}\E\left[
\max_{s\le u\le t+h_t}g_u(D_u)
\,\middle|\,\mathscr F_{s-1}\right]\\
&\quad\le C_\alpha\1_{\{\kappa_t=s\}}(t+h_t)\\
&\qquad+\frac{C_\alpha\1_{\{\kappa_t=s\}}}{\Delta^2s}\left\{
\left(\frac{t+h_t}{s}\right)^{\alpha^2}D_s^2
+d(t+h_t-s)
+\min\{d(t+h_t),(t+h_t-s)^2\}
\right\}
\end{aligned}
\]
Here $(t+h_t)/s\le9/8$ and $D_s^2\le C_\alpha(t+d^2)$.  Since
$s\ge t$, $0\le t+h_t-s\le h_t\le t/8$, and
$t\ge N\ge d/\Delta^2$, the right-hand side is at most
$C_\alpha t\,\1_{\{\kappa_t=s\}}$.  Thus,
\[
\1_{\{\kappa_t=s\}}\E\left[
\max_{s\le u\le t+h_t}g_u(D_u)
\,\middle|\,\mathscr F_{s-1}\right]
\le C_\alpha t\,\1_{\{\kappa_t=s\}}.
\]
Applying the tower property and summing over $s$ yields
\[
\begin{aligned}
&\E\left[
\1_{\{\kappa_t\le t+h_t\}}
\max_{\kappa_t\le u\le t+h_t}g_u(D_u)
\,\middle|\,\mathscr F_{t-1}\right]\\
&\qquad\le
C_\alpha t\,\P(\kappa_t\le t+h_t\mid\mathscr F_{t-1}).
\end{aligned}
\]

It remains to show that the probability of $\kappa_t\le t+h_t$ is
sufficiently small.  Appendix~\ref{sec:parameter-verification} verifies
\[
h_t\le\frac14\left(\frac{\sqrt t}{\alpha}-D_t\right).
\]
Since $\mu_*\le1$, this verifies the condition in
Lemma~\ref{lem:negative-persistence} with $L=h_t$.  Consequently,
\begin{align}
\P(\kappa_t\le t+h_t\mid\mathscr F_{t-1})
&\le
\exp\left\{
-\frac{(\sqrt t/\alpha-D_t)^2}
{8\bigl(dh_t+d(\sqrt t/\alpha-D_t)\bigr)}
\right\}\notag\\
&\le
\exp\left\{-\frac{\sqrt t/\alpha-D_t}{10d}\right\}
\le
\exp\left\{-\frac{t}{C_\alpha d\sqrt{g_t(D_t)}}\right\}.
\label{eq:positive-threshold-hitting-probability}
\end{align}
Here the second inequality uses
$h_t\le(\sqrt t/\alpha-D_t)/4$, whereas the third follows from
\[
\frac{\sqrt t}{\alpha}-D_t
\ge\frac{t}{\alpha\sqrt{g_t(D_t)}}.
\]
Substituting \eq{eq:positive-threshold-hitting-probability} into the
preceding post-hitting estimate gives
\begin{align}
&\E\left[
\1_{\{\kappa_t\le t+h_t\}}
\max_{\kappa_t\le u\le t+h_t}g_u(D_u)
\,\middle|\,\mathscr F_{t-1}\right]\notag\\
&\qquad\le C_\alpha t\exp\left\{-\frac{t}
{C_\alpha d\sqrt{g_t(D_t)}}\right\}.
\label{eq:post-positive-threshold-payment}
\end{align}

\medskip\par\noindent\textbf{Before $D_u$ reaches $\sqrt t/\alpha$.}\quad
As in the treatment of $v_t$ in the proof of
Lemma~\ref{lem:one-block-contraction}, we identify an event, shown below to
have high conditional probability, on which $g_u(D_u)$ is controlled
uniformly for $t\le u\le t+h_t$ before $D_u$ reaches $\sqrt t/\alpha$.
For this purpose, consider the stopped running increment
\[
\max_{t\le u\le(t+h_t)\wedge\kappa_t}
\bigl(M_u^{\mathcal O}-M_t^{\mathcal O}-\mu_*(u-t)\bigr).
\]
In Lemma~\ref{lem:one-block-contraction}, the event $v_t>1/4$ was reduced to
a large value of the terminal optimal-set increment $R_t^+$, whose tail is
recorded in \eq{eq:negative-R-tail}.  The corresponding exceptional event
here is that the displayed stopped running increment is large.  Both tail
bounds follow from the maximal-tail conclusion of
Lemma~\ref{lem:negative-persistence}.  On the complementary event,
\eq{eq:positive-threshold-branch-comparison} reduces the uniform
control of $g_u(D_u)$, for $t\le u\le t+h_t$ with $u<\kappa_t$, to a lower
bound on $2\sqrt u-\alpha D_u$.  We therefore split according to whether the
displayed stopped running increment is smaller than
$t/(8\sqrt{g_t(D_t)})$.

Apply Lemma~\ref{lem:negative-persistence} with $L=h_t$ and
$x=t/(8\sqrt{g_t(D_t)})$.  Since Appendix~
\ref{sec:parameter-verification} gives
$h_t\sqrt{g_t(D_t)}/t\le1/8$, its maximal-tail conclusion yields
\[
\begin{aligned}
&\P\left(
\max_{t\le u\le(t+h_t)\wedge\kappa_t}
\bigl(M_u^{\mathcal O}-M_t^{\mathcal O}-\mu_*(u-t)\bigr)
\ge\frac{t}{8\sqrt{g_t(D_t)}}
\,\middle|\,\mathscr F_{t-1}\right)\\
&\qquad\le\exp\left\{-\frac{t}
{C_\alpha d\sqrt{g_t(D_t)}}\right\}.
\end{aligned}
\]
On the smaller-fluctuation event, the monotonicity of
\(N_u^{\mathcal S}\) gives
\[
D_u-D_t\le\mu_*(u-t)+\frac{t}{8\sqrt{g_t(D_t)}},
\qquad
t\le u\le t+h_t,\quad u<\kappa_t.
\]
Consequently,
\[
\begin{aligned}
2\sqrt u-\alpha D_u
&=\frac{2t}{\sqrt{g_t(D_t)}}
+2(\sqrt u-\sqrt t)-\alpha(D_u-D_t)\\
&\ge\frac{2t}{\sqrt{g_t(D_t)}}
-\alpha h_t-\frac{\alpha t}{8\sqrt{g_t(D_t)}}
\ge\frac{7t}{4\sqrt{g_t(D_t)}}.
\end{aligned}
\]
Combining this denominator bound with
\eq{eq:positive-threshold-branch-comparison} and using
$u\le t+h_t\le9t/8$, where the second inequality follows from
\eq{eq:adaptive-block-scales}, gives
\[
\max_{\substack{t\le u\le t+h_t\\u<\kappa_t}}
g_u(D_u)\le C_\alpha g_t(D_t)
\]
on the smaller-fluctuation event.  On the complementary event,
$u<\kappa_t$ implies $2\sqrt u-\alpha D_u\ge\sqrt t$; hence
\eq{eq:positive-threshold-branch-comparison} gives
\[
\max_{\substack{t\le u\le t+h_t\\u<\kappa_t}}
g_u(D_u)\le C_\alpha t.
\]
Combining these two pathwise bounds with the maximal-tail estimate from
Lemma~\ref{lem:negative-persistence} gives
\[
\begin{aligned}
&\E\left[
\max_{\substack{t\le u\le t+h_t\\u<\kappa_t}}g_u(D_u)
\,\middle|\,\mathscr F_{t-1}\right]\\
&\qquad\le C_\alpha g_t(D_t)
+C_\alpha t\exp\left\{-\frac{t}
{C_\alpha d\sqrt{g_t(D_t)}}\right\}.
\end{aligned}
\]
The pathwise decomposition
\[
\max_{t\le u\le t+h_t}g_u(D_u)
\le
\max_{\substack{t\le u\le t+h_t\\u<\kappa_t}}g_u(D_u)
+\1_{\{\kappa_t\le t+h_t\}}
\max_{\kappa_t\le u\le t+h_t}g_u(D_u)
\]
separates the maximum before and after $\kappa_t$.  Taking conditional
expectations, the conditional estimate for the maximum over $u<\kappa_t$
controls the first term, and
\eq{eq:post-positive-threshold-payment} controls the second term.  Hence,
after changing $C_\alpha$ by a universal factor,
\[
\E\left[\max_{t\le u\le t+h_t}g_u(D_u)
\,\middle|\,\mathscr F_{t-1}\right]
\le C_\alpha g_t(D_t)
+C_\alpha t\exp\left\{-\frac{t}
{C_\alpha d\sqrt{g_t(D_t)}}\right\}.
\]
After enlarging $C_\alpha$ if necessary,
\eq{eq:exceptional-payment-absorption} and $h_t\le t/8$ absorb the
exponential term into $C_\alpha g_t(D_t)$.
This proves the negative-starting-point case of
\eq{eq:one-block-maximal} and completes the proof.
\end{proof}

\section{Growth of \texorpdfstring{$N_t^{\mathcal S}$}{the suboptimal-set minimum}}
\label{sec:suboptimal-minimum}

Throughout this appendix, $C$ or $C'$ denotes a generic positive universal
constant whose value may change from line to line and is determined by the
context; $C_\alpha$ denotes a positive constant that depends only on $\alpha$.

In the nontrivial case $\mathcal S\ne\varnothing$, our goal is to derive lower bounds for
\[
 N_t^{\mathcal S}:=\min_{i\in\mathcal S}\hat L_{t,i}.
\]
Together with the estimates for $M_t^{\mathcal O}$ in
Appendix~\ref{sec:optimal-minimum}, lower
bounds for $N_{t+L}^{\mathcal S}-N_t^{\mathcal S}$ are used to compute
\[
 \frac{\E[g_{t+L}(D_{t+L})\mid\mathscr F_{t-1}]}{g_t(D_t)}.
\]
The required estimates must control both the mean growth and the downward
fluctuation of $N_t^{\mathcal S}$ over rounds $t,\ldots,t+L-1$.

\subsection{From all \texorpdfstring{$L$}{L} rounds to stopped increments}

We first explain why the fixed-block argument of \citet{zhan2026does} is
insufficient for controlling the growth of $N_t^{\mathcal S}$ throughout the
region required by our Lyapunov analysis.  We then introduce a
stopped-increment argument that remains valid over this entire region.

\citet[Appendix~E]{zhan2026does} derives mean-growth and
downward-fluctuation estimates for $N_{t+L}-N_t$ under
$D_t\ge-\sqrt{dt}/\alpha$ and $4dL\le\sqrt{dt}/(4\alpha)$.  However, our Lyapunov
analysis requires mean-growth and downward-fluctuation estimates for
$N_{t+L}^{\mathcal S}-N_t^{\mathcal S}$ throughout the wider range
$D_t\ge-\beta t$.  The condition $D_t\ge-\sqrt{dt}/\alpha$ in
\citet{zhan2026does} is used to obtain a uniform lower bound on the sampling
probabilities.  In their notation, $i_*$ is the unique optimal arm and
$N_t=\min_{i\ne i_*}\hat L_{t,i}$.  Lemma D.4 of
\citet{zhan2026does} first shows
that $N_{r+1}-N_r\le4d$ and $D_{r+1}-D_r\ge-4d$ for
$r=t,\ldots,t+L-1$.  The bound $N_{r+1}-N_r\le4d$ implies that every arm
that can attain $N_{t+L}$ belongs to
$\mathcal A_t:=\{i\ne i_*:\hat L_{t,i}-N_t\le4dL\}$.  For every
$i\in\mathcal A_t$, \citet{zhan2026does} defines
\[
 W_i
 :=\hat L_{t+L,i}-\hat L_{t,i}-L\mu_i
 =\sum_{r=t}^{t+L-1}(\hat\ell_{r,i}-\mu_i).
\]
The sum defining $W_i$ contains every round from $t$ to $t+L-1$.
Lemma D.4 also gives $p_{r,i}^{-1}\le4d$ for every
$r=t,\ldots,t+L-1$, so the one-sided Freedman inequality controls
$\P(-W_i\ge x\mid\mathscr F_{t-1})$ uniformly over $i\in\mathcal A_t$.

The uniform estimate $p_{r,i}^{-1}\le4d$ is precisely the step that fails when
$D_t$ is allowed to be as small as $-\beta t$.  For a suboptimal arm whose
estimate at time $t$ is within $L$ of $N_t^{\mathcal S}$, the lower
sampling-probability bound in Lemma~\ref{4-2} only gives
\[
 p_{t,i}^{-1}
 \le
 \left(
  \sqrt d+\frac{\alpha(L+(-D_t)^+)}{2\sqrt t}
 \right)^2
 \le C\left(d+\frac{(L+(-D_t)^+)^2}{t}\right).
\]
Under $D_t\ge-\sqrt{dt}/\alpha$ and
$4dL\le\sqrt{dt}/(4\alpha)$, the upper bound on $p_{t,i}^{-1}$ is of order
$d$.  By contrast, when $D_t=-\Theta(t)$, the upper bound on $p_{t,i}^{-1}$ can be
of order $t$.  The resulting upper bound on the predictable quadratic
variation of $-W_i$ therefore deteriorates from order $dL$ to order $Lt$, and
the Freedman bound for
$\P(-W_i\ge x\mid\mathscr F_{t-1})$ is no longer uniform over $D_t\ge-\beta t$.

The conditional-variance bound of order $Lt$ motivates replacing $W_i$ with
a centered increment that stops when $\hat L_{r,i}$ first reaches
$N_t^{\mathcal S}+L$.  While $\hat L_{r,i}<N_t^{\mathcal S}+L$,
Lemma~\ref{4-2} gives a lower bound on $p_{r,i}$.  After
$\hat L_{r,i}\ge N_t^{\mathcal S}+L$, monotonicity preserves
$\hat L_{t+L,i}\ge N_t^{\mathcal S}+L$.  We therefore define $\mathcal B_t$
using distance $L$, rather than the distance $4dL$ in $\mathcal A_t$, and
define $U_i$ by summing only until $\hat L_{r,i}$ first reaches
$N_t^{\mathcal S}+L$, rather than through round $t+L-1$ as in $W_i$.  The
lower bound on $p_{r,i}$ while $\hat L_{r,i}<N_t^{\mathcal S}+L$ is valid for
every value of $D_t$.  Thus the stopped-increment argument extends the moment
bounds in \citet[Lemma~E.3(1)]{zhan2026does} to $D_t\ge-\beta t$.  The
specialization at the end of this appendix shows how the same argument
recovers \citet[Lemma~E.3(1)]{zhan2026does} under
$D_t\ge-\sqrt{dt}/\alpha$ and $4dL\le\sqrt{dt}/(4\alpha)$.

Fix a positive $\mathscr F_{t-1}$-measurable integer $L$.  Define
\begin{equation}
 \mathcal B_t
 :=\left\{
  i\in\mathcal S:
  \hat L_{t,i}-N_t^{\mathcal S}<L
 \right\}.
 \label{eq:barrier-active-set}
\end{equation}
For every $i\in\mathcal B_t$, let
\begin{equation}
 \zeta_{t,i}
 :=\inf\left\{
  u\ge t:
  \hat L_{u,i}\ge N_t^{\mathcal S}+L
 \right\}\wedge(t+L),
 \label{eq:barrier-stopping-time}
\end{equation}
where $\inf\varnothing=\infty$, and define the stopped centered increment
\begin{equation}
 U_i
 :=\sum_{r=t}^{\zeta_{t,i}-1}
 (\hat\ell_{r,i}-\mu_i).
 \label{eq:barrier-deviation}
\end{equation}
The stopping time $\zeta_{t,i}$ is taken with respect to
$(\mathscr F_{r-1})_{r\ge t}$ because $L$ and $N_t^{\mathcal S}$ are
$\mathscr F_{t-1}$-measurable and $\hat L_{r,i}$ is $\mathscr F_{r-1}$-measurable.

\begin{Lem}[Lower bound for the suboptimal-set minimum]
\label{lem:barrier-decomposition}
Pathwise,
\begin{equation}
 0
 \le
 \left(
  (\mu_*+\Delta)L
  -(N_{t+L}^{\mathcal S}-N_t^{\mathcal S})
 \right)^+
 \le
 \max_{i\in\mathcal B_t}(-U_i)^+
 \le L.
 \label{eq:barrier-minimum-bridge}
\end{equation}
\end{Lem}

\begin{proof}
If $i\notin\mathcal B_t$, then
$\hat L_{t,i}\ge N_t^{\mathcal S}+L$, and monotonicity gives
\[
 \hat L_{t+L,i}
 \ge N_t^{\mathcal S}+L
 \ge N_t^{\mathcal S}+(\mu_*+\Delta)L.
\]
If $i\in\mathcal B_t$ and $\zeta_{t,i}<t+L$, then
$\hat L_{\zeta_{t,i},i}\ge N_t^{\mathcal S}+L$ and monotonicity again gives
$\hat L_{t+L,i}\ge N_t^{\mathcal S}+(\mu_*+\Delta)L$.  If
$i\in\mathcal B_t$ and
$\zeta_{t,i}=t+L$, then
\[
 \hat L_{t+L,i}-N_t^{\mathcal S}
 \ge(\mu_*+\Delta)L+U_i.
\]
Since $\max_{j\in\mathcal B_t}(-U_j)^+\ge0$, the first two cases also satisfy
\[
 \hat L_{t+L,i}-N_t^{\mathcal S}
 \ge(\mu_*+\Delta)L-\max_{j\in\mathcal B_t}(-U_j)^+.
\]
In the third case,
$U_i\ge-(-U_i)^+\ge-\max_{j\in\mathcal B_t}(-U_j)^+$, so the same bound holds.
Taking the minimum over $i\in\mathcal S$, rearranging, and taking the positive
part proves the middle inequality in
\eq{eq:barrier-minimum-bridge}.  The positive part is nonnegative.  For
every $i\in\mathcal B_t$,
$-U_i\le\mu_i(\zeta_{t,i}-t)\le L$, which proves the upper bound by $L$ in
\eq{eq:barrier-minimum-bridge}.
\end{proof}

\subsection{Concentration and growth of
\texorpdfstring{$N_t^{\mathcal S}$}{the suboptimal-set minimum}}

We next derive concentration and moment bounds from the stopped-increment
argument.  These bounds determine a common block length for both half-lines
within the controlled strip and recover the block scale of
\citet{zhan2026does} in their setting.

Lemma~\ref{lem:barrier-decomposition} reduces the first- and second-moment
estimates for the positive shortfall in
\eq{eq:barrier-minimum-bridge} to estimates for
$\max_{i\in\mathcal B_t}(-U_i)^+$.  We first lower-bound $p_{r,i}$ for
$t\le r<\zeta_{t,i}$.

Before $\zeta_{t,i}$, monotonicity of $M_r$ and the identity
$N_t^{\mathcal S}-M_t=(-D_t)^+$ give
\[
 \underline{\hat L}_{r,i}
 =\hat L_{r,i}-M_r
 \le N_t^{\mathcal S}+L-M_t
 =L+(-D_t)^+.
\]
For convenience, define
\begin{equation}
 P_t
 :=\left(
  \sqrt d+\frac{\alpha(L+(-D_t)^+)}{2\sqrt t}
 \right)^{-2}.
 \label{eq:probability-lower-bound}
\end{equation}
Lemma~\ref{4-2} therefore gives
\begin{equation}
 p_{r,i}\ge P_t,
 \qquad t\le r<\zeta_{t,i}.
 \label{eq:active-probability-lower-bound}
\end{equation}
Thus $P_t$ is a common lower bound on the sampling probabilities of all
arms while their stopped increments are being accumulated.

\begin{Lem}[Concentration of the stopped increments]
\label{lem:stopped-concentration}
For every $i\in\mathcal B_t$ and $x\ge0$,
\begin{equation}
 \P(-U_i\ge x\mid\mathscr F_{t-1})
 \le
 \exp\left(
  -\frac{x^2}{2(L/P_t+x/3)}
 \right).
 \label{eq:barrier-single-arm-tail}
\end{equation}
Moreover,
\begin{align}
 &\E\left[
  \left(
   (\mu_*+\Delta)L
   -(N_{t+L}^{\mathcal S}-N_t^{\mathcal S})
  \right)^+
  \middle|\mathscr F_{t-1}
 \right]
 &\le
 C\left(
  \sqrt{\frac{L\log(2|\mathcal S|)}{P_t}}
  +\log(2|\mathcal S|)
 \right),
 \label{eq:fixed-barrier-first-moment}\\
 &\E\left[
  \left[
   \left(
    (\mu_*+\Delta)L
    -(N_{t+L}^{\mathcal S}-N_t^{\mathcal S})
   \right)^+
  \right]^2
  \middle|\mathscr F_{t-1}
 \right]
 &\le
 C\left(
  \frac{L\log(2|\mathcal S|)}{P_t}
  +\log^2(2|\mathcal S|)
 \right).
 \label{eq:fixed-barrier-second-moment}
\end{align}
\end{Lem}

\begin{proof}
Because $L$ is $\mathscr F_{t-1}$-measurable, conditioning on $\mathscr F_{t-1}$ fixes $L$
and all quantities defined at time $t$.  For $t\le r<t+L$, set
\[
 X_{r,i}:=\1_{\{r<\zeta_{t,i}\}}(\mu_i-\hat\ell_{r,i}).
\]
The variables $X_{r,i}$ are martingale differences, $X_{r,i}\le1$, and
\[
 \E[X_{r,i}^2\mid\mathscr F_{r-1}]
 \le \1_{\{r<\zeta_{t,i}\}}p_{r,i}^{-1}
 \le P_t^{-1}.
\]
Since $-U_i=\sum_{r=t}^{t+L-1}X_{r,i}$, its predictable quadratic variation
is at most $L/P_t$.  The one-sided Freedman inequality therefore gives
\eq{eq:barrier-single-arm-tail}.

A union bound over at most $|\mathcal S|$ arms and integration over $x$ show
that the first and second moments of
$\max_{i\in\mathcal B_t}(-U_i)^+$ are bounded by the right-hand sides of
\eq{eq:fixed-barrier-first-moment} and
\eq{eq:fixed-barrier-second-moment}, respectively.
Lemma~\ref{lem:barrier-decomposition} gives
\[
 \left(
  (\mu_*+\Delta)L
  -(N_{t+L}^{\mathcal S}-N_t^{\mathcal S})
 \right)^+
 \le\max_{i\in\mathcal B_t}(-U_i)^+,
\]
which proves \eq{eq:fixed-barrier-first-moment} and
\eq{eq:fixed-barrier-second-moment}.
\end{proof}

For the drift term in the Lyapunov supermartingale inequality to be negative,
the suboptimal-set minimum $N_t^{\mathcal S}$ must grow faster than the
optimal-set minimum $M_t^{\mathcal O}$.
Appendix~\ref{sec:optimal-minimum} controls
$M_{t+L}^{\mathcal O}-M_t^{\mathcal O}$ by $\mu_*L$ plus a centered
fluctuation.  Fix $\varepsilon\in(0,1)$.  We seek a horizon $L$ such that
\[
 \E[
 N_{t+L}^{\mathcal S}-N_t^{\mathcal S}
 \mid\mathscr F_{t-1}]
 \ge(\mu_*+(1-\varepsilon)\Delta)L.
\]
By Lemma~\ref{lem:stopped-concentration}, a sufficient condition is
\begin{equation}
 LP_t\ge C\varepsilon^{-2}\Delta^{-2}\log(2|\mathcal S|).
 \label{eq:effective-sampling-condition}
\end{equation}
Indeed, after increasing the universal constant, this condition makes the
right-hand side of \eq{eq:fixed-barrier-first-moment} at most
$\varepsilon\Delta L$.

Since $P_t$ depends on $L$, the condition in
\eq{eq:effective-sampling-condition} determines an interval of feasible
horizons.  Let $L_-<L_+$ be the two solutions obtained by replacing the
inequality with equality.  Under the parameter conditions imposed below, the
smaller solution satisfies
\[
 L_-
 \asymp
 \varepsilon^{-2}\Delta^{-2}\log(2|\mathcal S|)
 \left(
  \sqrt d+\frac{\alpha(-D_t)^+}{2\sqrt t}
 \right)^2.
\]
Thus, when $A$ is chosen at a sufficiently large constant multiple of
$\varepsilon^{-2}\Delta^{-2}\log(2|\mathcal S|)$, the horizon
\[
 A\left(
  \sqrt d+\frac{\alpha(-D_t)^+}{2\sqrt t}
 \right)^2
\]
has the same order as $L_-$.  Once the required effective sampling condition
is reached, taking a larger feasible value of $L$ does not improve the final
problem-parameter dependence in the Lyapunov analysis, so we use this shorter
scale.

This scale also has a direct sampling interpretation.  For any arm
$i\in\mathcal S$ attaining $N_t^{\mathcal S}$, Lemma~\ref{4-2} gives
\[
 p_{t,i}^{-1}
 \le
 \left(
  \sqrt d+\frac{\alpha(-D_t)^+}{2\sqrt t}
 \right)^2.
\]
Hence the proposed horizon is, up to the concentration factor $A$, the
reciprocal sampling-probability scale of an arm currently attaining
$N_t^{\mathcal S}$.  A block of this order gives such an arm enough
opportunities to be selected.  By contrast, if one of the arms attaining
$N_t^{\mathcal S}$ is not selected during the block, its cumulative estimate
does not change, and $N_t^{\mathcal S}$ cannot increase.  The uniform
within-block lower bound $P_t$ makes this intuition rigorous: the choice below
will satisfy $h_tP_t\ge A/C$.

For fixed $\varepsilon$, this scale also recovers the order of the block
length in \citet{zhan2026does}.  Indeed, under
$D_t\ge-\sqrt{dt}/\alpha$, the squared factor above is of order $d$.  In their
unique-optimal-arm setting, $|\mathcal S|=d-1$, and hence
$h_t\asymp d\Delta^{-2}\log(2d)$.

These considerations motivate a single local block length for both half-lines
within the controlled strip.  It is fixed when $D_t\ge0$ and adapts to
$(-D_t)^+$ when $-\beta t\le D_t<0$.

\begin{Cor}[Choice of $h_t$]
\label{cor:state-dependent-horizon}
Suppose that $D_t\ge-\beta t$, fix $\varepsilon\in(0,1)$, and set
\begin{equation}
 h_t
 :=\left\lceil
  A\left(
   \sqrt d+\frac{\alpha(-D_t)^+}{2\sqrt t}
  \right)^2
 \right\rceil.
 \label{eq:state-dependent-horizon}
\end{equation}
Assume that $A\ge1$,
$A\ge C\varepsilon^{-2}\Delta^{-2}\log(2|\mathcal S|)$,
$A\beta\le1$, and
$t\ge A^2d$.  Evaluate
$\mathcal B_t$, $\zeta_{t,i}$, $U_i$, and $P_t$ at $L=h_t$.  Then
\begin{align}
 &\E\left[
  \left(
   (\mu_*+\Delta)h_t
   -(N_{t+h_t}^{\mathcal S}-N_t^{\mathcal S})
  \right)^+
  \middle|\mathscr F_{t-1}
 \right]
 \le C h_t\sqrt{\frac{\log(2|\mathcal S|)}{A}}
 \le\varepsilon\Delta h_t,
 \label{eq:state-dependent-first-moment}\\
 &\E\left[
  \left[
   \left(
    (\mu_*+\Delta)h_t
    -(N_{t+h_t}^{\mathcal S}-N_t^{\mathcal S})
   \right)^+
  \right]^2
  \middle|\mathscr F_{t-1}
 \right]
 \le\frac{C\log(2|\mathcal S|)}{A}h_t^2.
 \label{eq:state-dependent-second-moment}
\end{align}
\end{Cor}

\begin{proof}
Since $A\ge1$ and $d\ge1$,
\[
 A\left(
  \sqrt d+\frac{\alpha(-D_t)^+}{2\sqrt t}
 \right)^2
 \le h_t
 \le
 2A\left(
  \sqrt d+\frac{\alpha(-D_t)^+}{2\sqrt t}
 \right)^2.
\]
Moreover, $D_t\ge-\beta t$, $t\ge A^2d$, $A\beta\le1$, and
$\alpha\in(0,1)$ imply
\[
 \frac{A\alpha}{\sqrt t}
 \left(
  \sqrt d+\frac{\alpha(-D_t)^+}{2\sqrt t}
 \right)
 \le
 1+\frac12.
\]
Consequently, with $P_t$ evaluated at $L=h_t$,
\[
 P_t^{-1}
 =
 \left(
  \sqrt d+
  \frac{\alpha((-D_t)^++h_t)}{2\sqrt t}
 \right)^2
 \le
 C\left(
  \sqrt d+\frac{\alpha(-D_t)^+}{2\sqrt t}
 \right)^2
 \le C\frac{h_t}{A}.
\]
In particular, $h_tP_t\ge A/C$.  Substituting this bound into
Lemma~\ref{lem:stopped-concentration} with $L=h_t$ yields
\begin{align*}
 &\E\left[
  \left(
   (\mu_*+\Delta)h_t
   -(N_{t+h_t}^{\mathcal S}-N_t^{\mathcal S})
  \right)^+
  \middle|\mathscr F_{t-1}
 \right]
 \le C\left(
  h_t\sqrt{\frac{\log(2|\mathcal S|)}{A}}
  +\log(2|\mathcal S|)
 \right),\\
 &\E\left[
  \left[
   \left(
    (\mu_*+\Delta)h_t
    -(N_{t+h_t}^{\mathcal S}-N_t^{\mathcal S})
   \right)^+
  \right]^2
  \middle|\mathscr F_{t-1}
 \right]
 \le C\left(
  \frac{\log(2|\mathcal S|)}{A}h_t^2
  +\log^2(2|\mathcal S|)
 \right).
\end{align*}
Finally, $h_t\ge A$ and a sufficiently large universal constant in
$A\ge C\varepsilon^{-2}\Delta^{-2}\log(2|\mathcal S|)$ absorb the additive
logarithmic terms and give \eq{eq:state-dependent-first-moment} and
\eq{eq:state-dependent-second-moment}.  When $D_t\ge0$, this choice reduces
to $h_t=\lceil Ad\rceil$.
\end{proof}

For comparison with \citet[Appendix~E]{zhan2026does}, specialize
Lemma~\ref{lem:stopped-concentration} to
$D_t\ge-\sqrt{dt}/\alpha$ and $4dL\le\sqrt{dt}/(4\alpha)$, the range studied by
\citet{zhan2026does}.  The first condition gives
$\alpha(-D_t)^+/(2\sqrt t)\le\sqrt d/2$, while the second gives
$\alpha L/(2\sqrt t)\le1/(32\sqrt d)$.  Hence
\[
 P_t^{-1/2}
 =\sqrt d+\frac{\alpha(L+(-D_t)^+)}{2\sqrt t}
 \le2\sqrt d,
 \qquad
 P_t^{-1}\le4d.
\]
Therefore \eq{eq:barrier-single-arm-tail} has the same Freedman form as
\citet[Lemma~E.2]{zhan2026does}, and Lemma~\ref{lem:stopped-concentration}
gives
\begin{align}
 &\E\left[
  \left(
   (\mu_*+\Delta)L
   -(N_{t+L}^{\mathcal S}-N_t^{\mathcal S})
  \right)^+
  \middle|\mathscr F_{t-1}
 \right]
 &\le
 C\left(
  \sqrt{4dL\log(2|\mathcal S|)}+\log(2|\mathcal S|)
 \right),
 \label{eq:fixed-first-moment}\\
 &\E\left[
  \left[
   \left(
    (\mu_*+\Delta)L
    -(N_{t+L}^{\mathcal S}-N_t^{\mathcal S})
   \right)^+
  \right]^2
  \middle|\mathscr F_{t-1}
 \right]
 &\le
 C\left(
  4dL\log(2|\mathcal S|)+\log^2(2|\mathcal S|)
 \right).
 \label{eq:fixed-second-moment}
\end{align}
These are the first- and second-moment bounds of
\citet[Lemma~E.3(1)]{zhan2026does}; the first also implies
\citet[Lemma~5.1]{zhan2026does}.

\section{Computation related to
\texorpdfstring{$M_{t+L}^{\mathcal O}-M_t^{\mathcal O}$}{the optimal-set minimum}}
\label{sec:optimal-minimum}

In this appendix, we give estimates for
\(M_{t+L}^{\mathcal O}-M_t^{\mathcal O}\), which are needed to compute
\[
 \frac{\E[g_{t+L}(D_{t+L})\mid\mathscr F_{t-1}]}{g_t(D_t)}
\]
in the local Lyapunov argument.  Throughout this appendix, \(L\) may be a
positive, finite, integer-valued, \(\mathscr F_{t-1}\)-measurable random variable.
Conditional on \(\mathscr F_{t-1}\), its value is fixed, so the arguments below apply
conditionally.  All conditions involving \(L\) are understood almost surely.
A substantive difference from
\citet[Appendix~F]{zhan2026does} is that their unique optimal arm makes the
optimal-side increment the growth of one fixed coordinate, whereas here
\[
 M_t^{\mathcal O}=\min_{a\in\mathcal O}\hat L_{t,a}.
\]
The arm attaining this minimum may change from round to round, so
\(M_{t+L}^{\mathcal O}-M_t^{\mathcal O}\) need not be the increment of one arm
fixed before the block.  We handle this change one round at a time.  Before
round \(r\), choose
\[
 i_r^{\mathcal O}\in\arg\min_{a\in\mathcal O}\hat L_{r,a}
\]
using a fixed tie-breaking rule.  This arm is \(\mathscr F_{r-1}\)-measurable, and
\begin{equation}
 \begin{aligned}
  M_{r+1}^{\mathcal O}-M_r^{\mathcal O}
  &\le \hat\ell_{r,i_r^{\mathcal O}},\\
  M_{t+L}^{\mathcal O}-M_t^{\mathcal O}
  &\le\sum_{r=t}^{t+L-1}\hat\ell_{r,i_r^{\mathcal O}}.
 \end{aligned}
 \label{eq:moving-minimum-comparison}
\end{equation}
Indeed, since \(i_r^{\mathcal O}\) attains the minimum at time \(r\),
\[
 M_{r+1}^{\mathcal O}
 =\min_{a\in\mathcal O}
   \bigl(\hat L_{r,a}+\hat\ell_{r,a}\bigr)
 \le \hat L_{r,i_r^{\mathcal O}}+\hat\ell_{r,i_r^{\mathcal O}}
 =M_r^{\mathcal O}+\hat\ell_{r,i_r^{\mathcal O}}.
\]
This proves the first inequality in
\eq{eq:moving-minimum-comparison}; summing it over the block proves the
second.  Thus, although the minimizing arm may change, the
increment is controlled round by round by an optimal arm chosen before the
current importance-weighted loss is observed.  Since every such arm has mean
\(\mu_*\), its centered increments remain martingale differences.  This is the
new moving-minimum comparison needed before the scalar calculations of
\citet[Appendix~F]{zhan2026does} can be applied.  It is independent of the sign
of \(D_t\) and will be used on both half-lines below.

Following the two branches of \(g_t\), we consider the cases \(D_t\ge0\) and
\(D_t<0\) separately.

\subsection{The positive half-line}

On the positive half-line,
\[
 g_t(y)=\frac{(y+\Delta t)^2}{\Delta^2t}.
\]
Throughout this subsection, suppose \(D_t\ge0\) and write
\(x:=D_t+\Delta t\).
Since
\[
 D_{t+L}-D_t
 =\bigl(M_{t+L}^{\mathcal O}-M_t^{\mathcal O}\bigr)
  -(N_{t+L}^{\mathcal S}-N_t^{\mathcal S}),
\]
the computation of the Lyapunov ratio
\[
 \frac{\E[g_{t+L}(D_{t+L})\mid\mathscr F_{t-1}]}{g_t(D_t)}
\]
requires a conditional second-moment bound for
\(M_{t+L}^{\mathcal O}-M_t^{\mathcal O}-\mu_*L\).  The increment of
\(N_t^{\mathcal S}\) is controlled separately.
Equation~\eqref{eq:moving-minimum-comparison}
reduces the stochastic part of this bound to the conditional second moment of
\[
 \sum_{r=t}^{t+L-1}
 \bigl(\hat\ell_{r,i_r^{\mathcal O}}-\mu_*\bigr).
\]
\noindent\begin{minipage}{\textwidth}
For importance-weighted loss estimates, this moment is controlled through the
inverse sampling probabilities.  Since \(i_r^{\mathcal O}\) attains
\(M_r^{\mathcal O}\), we have
\(\underline{\hat L}_{r,i_r^{\mathcal O}}=D_r^+\).
Following the lower sampling-probability bound in
Lemma~\ref{4-2}, as in \citet[Appendix~F]{zhan2026does}, define

\begin{equation}
 s_r:=\left(
  \frac{\alpha D_r^+}{2\sqrt r}+\sqrt d
 \right)^2.
 \label{eq:sr}
\end{equation}
\end{minipage}
The lower sampling-probability bound in Lemma~\ref{4-2} then gives
\(p_{r,i_r^{\mathcal O}}^{-1}\le s_r\).
Moreover, because \(i_r^{\mathcal O}\) is chosen before round \(r\) and every
arm in \(\mathcal O\) has mean \(\mu_*\),
\[
 \E[\hat\ell_{r,i_r^{\mathcal O}}\mid\mathscr F_{r-1}]=\mu_*,
 \qquad
 \E[\hat\ell_{r,i_r^{\mathcal O}}^2\mid\mathscr F_{r-1}]
 \le\frac{\mu_*}{p_{r,i_r^{\mathcal O}}}
 \le\mu_*s_r.
\]
Thus the centered summands below are martingale differences, and martingale
orthogonality and the tower property give
\begin{align}
 &\E\left[
  \left(
   \sum_{r=t}^{t+L-1}
   \bigl(\hat\ell_{r,i_r^{\mathcal O}}-\mu_*\bigr)
  \right)^2
  \,\middle|\,\mathscr F_{t-1}
 \right] \notag\\
 &\qquad\le
 \sum_{r=t}^{t+L-1}
 \E[\hat\ell_{r,i_r^{\mathcal O}}^2\mid\mathscr F_{t-1}]
 \le
 \mu_*\sum_{r=t}^{t+L-1}\E[s_r\mid\mathscr F_{t-1}].
 \label{eq:moving-minimum-to-sr}
\end{align}

It remains to control \(s_r\).  Since \(D_t=x-\Delta t\ge0\), the initial
estimate of \citet[Lemma~F.1]{zhan2026does} applies without change.  For the
one-step estimate, \(N_r^{\mathcal S}\) is nondecreasing and
\eq{eq:moving-minimum-comparison} gives
\[
 D_{r+1}^+-D_r^+
 \le (D_{r+1}-D_r)^+
 \le \hat\ell_{r,i_r^{\mathcal O}}.
\]
Together with the two conditional moment bounds above, these are exactly the
inputs to the scalar argument of \citet[Lemma~F.2]{zhan2026does}.  Iterating
that recursion as in \citet[Lemma~F.3]{zhan2026does} yields
\begin{align}
 \sum_{r=t}^{t+L-1}\E[s_r\mid\mathscr F_{t-1}]
 \le{}&\exp\left(\frac{C_\alpha L}{t}\right)
 \left[
  L\left(
   \frac{\alpha^2}{4}
   +\frac{\alpha\sqrt d}{4\Delta\sqrt t}
   +\frac{d}{\Delta^2t}
  \right)\frac{x^2}{t}
  +\frac{\alpha^2L^2}{8}
 \right].
 \label{eq:block-sum}
\end{align}

\begin{Cor}[Second moment of the optimal-set increment]
\label{cor:positive-optimal-increment}
There exists \(C_\alpha>0\), depending only on \(\alpha\), such that, for
every \(\rho\in(0,2)\), if \(D_t\ge0\) and
\(t\ge C_\alpha\rho^{-2}dL/\Delta^2\), then
\begin{align}
 \E\left[
  M_{t+L}^{\mathcal O}-M_t^{\mathcal O}-\mu_*L
  \,\middle|\,\mathscr F_{t-1}
 \right]&\le0,
 \label{eq:M-mean-positive}\\
 \E\left[
  \left(M_{t+L}^{\mathcal O}-M_t^{\mathcal O}-\mu_*L\right)^2
  \,\middle|\,\mathscr F_{t-1}
 \right]
 &\le
 \mu_*\left(\frac{\alpha^2}{4}+\frac{\rho}{4}\right)
 \frac{L(D_t+\Delta t)^2}{t}.
 \label{eq:M-second-positive}
\end{align}
\end{Cor}

\begin{proof}
Since \(D_t\ge0\), we have \(x\ge\Delta t\).  Equation~\eqref{eq:block-sum}, the stated
lower bound on \(t\), and the standing conditions
\(d\ge2\), \(0<\Delta\le1\), \(L\ge1\), and \(0<\rho<2\) give, after
increasing \(C_\alpha\) if necessary,
\[
 \sum_{r=t}^{t+L-1}\E[s_r\mid\mathscr F_{t-1}]
 \le\left(\frac{\alpha^2}{4}+\frac{\rho}{8}\right)
       \frac{Lx^2}{t}.
\]
Equation~\eqref{eq:moving-minimum-comparison} gives
\[
 M_{t+L}^{\mathcal O}-M_t^{\mathcal O}-\mu_*L
 \le
 \sum_{r=t}^{t+L-1}
 \bigl(\hat\ell_{r,i_r^{\mathcal O}}-\mu_*\bigr),
 \qquad
 \E\left[
  \sum_{r=t}^{t+L-1}
  \bigl(\hat\ell_{r,i_r^{\mathcal O}}-\mu_*\bigr)
  \,\middle|\,\mathscr F_{t-1}
 \right]=0,
\]
which proves \eq{eq:M-mean-positive}.  Since
\(M_{t+L}^{\mathcal O}\ge M_t^{\mathcal O}\), considering separately the
two signs of the centered increment also gives
\[
 \left(M_{t+L}^{\mathcal O}-M_t^{\mathcal O}-\mu_*L\right)^2
 \le
 \left(
  \sum_{r=t}^{t+L-1}
  \bigl(\hat\ell_{r,i_r^{\mathcal O}}-\mu_*\bigr)
 \right)^2
 +\mu_*^2L^2.
\]
Together with \eq{eq:moving-minimum-to-sr}, the first display of this proof
bounds the first term by
\[
 \mu_*\left(\frac{\alpha^2}{4}+\frac{\rho}{8}\right)
 \frac{Lx^2}{t}.
\]
If \(\mu_*=0\), the remaining term vanishes.  Otherwise, \(x\ge\Delta t\)
and the same lower bound on \(t\) give
\[
 \frac{\mu_*^2L^2}{(\mu_*\rho/8)Lx^2/t}
 \le\frac{8L}{\rho\Delta^2t}
 \le\frac{8\rho}{C_\alpha d}
 \le1
\]
after increasing \(C_\alpha\) if necessary.  Thus
\(\mu_*^2L^2\le(\mu_*\rho/8)Lx^2/t\).
Combining the last three displays proves \eq{eq:M-second-positive}.
\end{proof}

\subsection{The negative half-line}

On the negative half-line,
\[
 g_t(y)=\frac{4t^2}{(2\sqrt t-\alpha y)^2}.
\]
The ratio \(g_{t+L}(D_{t+L})/g_t(D_t)\) is controlled by applying
$(1-v)^{-2}\le1+2v+8v^2$ to the normalized change of the inverse-square
denominator.  The resulting calculation requires conditional moment and
maximal bounds for the upper fluctuation of
\(M_u^{\mathcal O}-M_t^{\mathcal O}\).

This case is somewhat simpler than the positive half-line because, with high
probability, \(i_r^{\mathcal O}\) continues to be sampled with probability at
least of order \(1/d\) throughout the block.  We find that, for a suitable
horizon \(L\), the gap process \((D_u)_{u\ge t}\) stays below the positive
level \(\sqrt t/\alpha\) throughout the next \(L\) rounds, namely,
\(D_u<\sqrt t/\alpha\) for every \(t\le u\le t+L\), with high conditional
probability; see Lemma~\ref{lem:negative-persistence}.  The inequality
$D_r<\sqrt t/\alpha$ is useful because it implies
\(\underline{\hat L}_{r,i_r^{\mathcal O}}=D_r^+<\sqrt t/\alpha\le
\sqrt r/\alpha\).  The lower sampling-probability bound in
Lemma~\ref{4-2} then gives
\[
 p_{r,i_r^{\mathcal O}}^{-1}
 \le\left(\sqrt d+\frac{\alpha D_r^+}{2\sqrt r}\right)^2
 \le\left(\sqrt d+\frac12\right)^2
 \le2d.
\]
Consequently,
$\hat\ell_{r,i_r^{\mathcal O}}\le
p_{r,i_r^{\mathcal O}}^{-1}\le2d$, and standard martingale inequalities
apply directly.

To formalize the event $D_u<\sqrt t/\alpha$ throughout the next $L$ rounds,
suppose \(D_t<0\) and define
\[
 \kappa_t:=\inf\left\{u\ge t:D_u\ge\frac{\sqrt t}{\alpha}\right\}.
\]
Since \(D_u\) is \(\mathscr F_{u-1}\)-measurable, \(\kappa_t\) is a stopping time
for \((\mathscr F_{u-1})_{u\ge t}\).  We therefore study the growth of
\(M_u^{\mathcal O}\) for \(t\le u\le(t+L)\wedge\kappa_t\).

\begin{Lem}[Maximal fluctuation before $D_u$ reaches $\sqrt t/\alpha$]
\label{lem:negative-persistence}
Suppose \(D_t<0\).  For every \(x\ge0\),
\[
 \P\left(
  \max_{t\le u\le(t+L)\wedge\kappa_t}
  \left(M_u^{\mathcal O}-M_t^{\mathcal O}-\mu_*(u-t)\right)\ge x
  \,\middle|\,\mathscr F_{t-1}
 \right)
 \le
 \exp\left(-\frac{x^2}{4dL+4dx/3}\right).
\]
Moreover, if
\[
 \mu_*L\le\frac14\left(\frac{\sqrt t}{\alpha}-D_t\right),
\]
then
\[
 \P\left(\kappa_t\le t+L\mid\mathscr F_{t-1}\right)
 \le
 \exp\left\{
 -\frac{(\sqrt t/\alpha-D_t)^2}
 {8\bigl(dL+d(\sqrt t/\alpha-D_t)\bigr)}
 \right\}.
\]
\end{Lem}

\begin{proof}
For every \(r<\kappa_t\), the same application of
Lemma~\ref{4-2} gives
\(p_{r,i_r^{\mathcal O}}^{-1}\le2d\).  Therefore, the stopped
centered increments
\[
 \1_{\{r<\kappa_t\}}
 \bigl(\hat\ell_{r,i_r^{\mathcal O}}-\mu_*\bigr)
\]
are martingale differences, are bounded above by \(2d\), and have predictable
quadratic variation at most \(2dL\) over the block.  The moving-minimum
comparison and the maximal Freedman inequality give the stated maximal tail.

If \(\kappa_t\le t+L\), the monotonicity of \(N_u^{\mathcal S}\) gives
 \[
 M_{\kappa_t}^{\mathcal O}-M_t^{\mathcal O}
 \ge\frac{\sqrt t}{\alpha}-D_t.
 \]
Under the stated condition, it follows that
\[
 M_{\kappa_t}^{\mathcal O}-M_t^{\mathcal O}
 -\mu_*(\kappa_t-t)
 \ge\frac34\left(\frac{\sqrt t}{\alpha}-D_t\right).
\]
Applying the maximal tail with
$x=3(\sqrt t/\alpha-D_t)/4$ proves the stated bound on
$\P(\kappa_t\le t+L\mid\mathscr F_{t-1})$.
\end{proof}

The bound on $\P(\kappa_t\le t+L\mid\mathscr F_{t-1})$ shows that the planned
endpoint \(t+L\) lies before \(\kappa_t\) with high probability whenever the deterministic growth
\(\mu_*L\) is sufficiently smaller than the initial distance to
\(\sqrt t/\alpha\).  We next record the conditional moment bounds that hold up
to this boundary.

\begin{Lem}[Stopped maximal optimal-set increment]
\label{lem:stopped-optimal-set-increment}
Suppose \(D_t<0\).  Then
\begin{align}
 \E\left[
  \max_{t\le u\le(t+L)\wedge\kappa_t}
  \left(M_u^{\mathcal O}-M_t^{\mathcal O}-\mu_*(u-t)\right)_+
  \,\middle|\,\mathscr F_{t-1}
 \right]&\le\sqrt{8dL},\notag\\
 \E\left[
  \max_{t\le u\le(t+L)\wedge\kappa_t}
  \left(M_u^{\mathcal O}-M_t^{\mathcal O}-\mu_*(u-t)\right)_+^2
  \,\middle|\,\mathscr F_{t-1}
 \right]&\le8dL.
 \label{eq:stopped-M-maximal}
\end{align}
\end{Lem}

\begin{proof}
Because \(D_u\) is \(\mathscr F_{u-1}\)-measurable, \(\kappa_t\) is a stopping time
for \((\mathscr F_{u-1})_{u\ge t}\), so
\(\1_{\{r<\kappa_t\}}\in\mathscr F_{r-1}\).  Since
\(i_r^{\mathcal O}\) is chosen before round \(r\) and has mean \(\mu_*\),
the stopped centered increments
\[
 \1_{\{r<\kappa_t\}}
 \bigl(\hat\ell_{r,i_r^{\mathcal O}}-\mu_*\bigr)
\]
are martingale differences.  The inequality
$p_{r,i_r^{\mathcal O}}^{-1}\le2d$ for $r<\kappa_t$ further implies
\[
 \left|\1_{\{r<\kappa_t\}}
 \bigl(\hat\ell_{r,i_r^{\mathcal O}}-\mu_*\bigr)\right|\le2d,
 \qquad
 \E\left[
 \1_{\{r<\kappa_t\}}
 \bigl(\hat\ell_{r,i_r^{\mathcal O}}-\mu_*\bigr)^2
 \middle|\mathscr F_{r-1}\right]
 \le\1_{\{r<\kappa_t\}}
       \frac{\mu_*}{p_{r,i_r^{\mathcal O}}}
 \le2d.
\]
Thus the predictable quadratic variation over the block is at most \(2dL\).
Moreover,
\[
 \sum_{r=t}^{t+L-1}\1_{\{r<\kappa_t\}}
 \bigl(\hat\ell_{r,i_r^{\mathcal O}}-\mu_*\bigr)
 =\sum_{r=t}^{(t+L)\wedge\kappa_t-1}
   \bigl(\hat\ell_{r,i_r^{\mathcal O}}-\mu_*\bigr).
\]
Martingale orthogonality gives
\[
 \E\left[
  \left(\sum_{r=t}^{t+L-1}\1_{\{r<\kappa_t\}}
  \bigl(\hat\ell_{r,i_r^{\mathcal O}}-\mu_*\bigr)\right)^2
  \,\middle|\,\mathscr F_{t-1}
 \right]\le2dL.
\]

For every \(u\le(t+L)\wedge\kappa_t\), summing the first inequality in
\eq{eq:moving-minimum-comparison} and subtracting \(\mu_*(u-t)\) gives
\[
 M_u^{\mathcal O}-M_t^{\mathcal O}-\mu_*(u-t)
 \le\sum_{r=t}^{u-1}\1_{\{r<\kappa_t\}}
 \bigl(\hat\ell_{r,i_r^{\mathcal O}}-\mu_*\bigr).
\]
Consequently,
\[
 \max_{t\le u\le(t+L)\wedge\kappa_t}
 \left(M_u^{\mathcal O}-M_t^{\mathcal O}-\mu_*(u-t)\right)_+^2
 \le
 \max_{t\le u\le t+L}
 \left|\sum_{r=t}^{u-1}\1_{\{r<\kappa_t\}}
 \bigl(\hat\ell_{r,i_r^{\mathcal O}}-\mu_*\bigr)\right|^2.
\]
Doob's \(L^2\) inequality gives the second bound in
\eq{eq:stopped-M-maximal}; conditional Cauchy--Schwarz then gives the first.
\end{proof}

\section{Proof of the Lower Bound in Lemma \ref{simlow}}\label{pfsimlow}
In this appendix, we prove Lemma~\ref{simlow}.
Throughout this appendix, we assume that the optimal arm is
unique, so $\mathcal O=\{i_*\}$.
We first reduce the desired lower bound to a uniform
second-moment bound for $D_{\Psi}(e_{i_*},p_t)$. We then obtain the required
linear estimate for $\E[\underline{\hat L}_{t,i_*}^2]$ directly from the
fixed-arm maximal bound in Appendix~\ref{sec:fixed-arm-moments} and substitute
it into the reduction.

We first compare the instantaneous expected loss gap with the Bregman divergence and identify the moment estimate needed to remove the resulting truncation. Since $D_{\Psi}(e_{i_*},p_t)$ can be unbounded when $p_{t,i_*}$ is small, the pointwise comparison below must be truncated.

\begin{Lem}\label{breglow}
    For any $t\ge 1$, {if $i_t^*=i_*$}, $\agp{p_t-e_{i_*},\mu}\ge \frac{\Delta}{16d}D^2_{\Psi}(e_{i_*},p_t)$; {if $i_t^*\ne i_*$}, $\agp{p_t-e_{i_*},\mu}\ge \frac{\Delta}{2}$. Hence, we have
    \[
    \agp{p_t-e_{i_*},\mu}\ge \frac{\Delta}{16d}D^2_{\Psi}(e_{i_*},p_t)\wedge \frac{\Delta}{2}.
    \]
\end{Lem}
The identity of the empirical minimizer determines whether $p_{t,i_*}$ is bounded from below or from above, so we treat the two cases separately.
\begin{proof}[Proof of Lemma \ref{breglow}]
    {If $i_{t}^*=i_*$}, for any $i\ne i_*$, $\hat{L}_{t,i}\ge\hat{L}_{t,i_*}$, which implies that $p_{t,i}\leq p_{t,i_*}$. Hence $p_{t,i_*}\ge \frac{1}{d}$. Then by the definition, \[
        \begin{aligned}
         D_{\Psi}(e_{i_*},p_t)=&2\sum_{i\ne i_*}\sqrt{p_{t,i}}+\frac{2(1-\sqrt{p_{t,i_*}})^2}{\sqrt{p_{t,i_*}}}\\\leq &2\sum_{i\ne i_*}\sqrt{p_{t,i}}+2\sqrt{d}(1-\sqrt{p_{t,i_*}})^2  \\
         \leq &2\sum_{i\ne i_*}\sqrt{p_{t,i}}+2\sqrt{d}(1-p_{t,i_*})^2.
        \end{aligned}
        \]
        By Jensen's inequality,
        $
        \sum_{i\ne i_*}\sqrt{p_{t,i}}\leq \sqrt{d\sum_{i\ne i_*} p_{t,i}}.
        $
        Also, note that
        $
        2\sqrt{d}(1-p_{t,i_*})^2\leq
        2\sqrt{d\sum_{i\ne i_*} p_{t,i}},
        $
        then we have
        \[
        D_{\Psi}(e_{i_*},p_t)\leq 4\sqrt{d\sum_{i\ne i_*} p_{t,i}}\leq 4\sqrt{\frac{d}{\Delta}\agp{p_t-e_{i_*},\mu}},
        \]
        which implies that
        $
    \agp{p_t-e_{i_*},\mu}\ge \frac{\Delta}{16d}D^2_{\Psi}(e_{i_*},p_t).
    $

    {If $i_t^*\ne i_*$}, similarly, there exists $i'\ne i_*$ such that $p_{t,i'}\ge p_{t,i_*}$ and hence $p_{t,i_*}\leq \frac12.$ Then clearly, we have
    $
    \agp{p_t-e_{i_*},\mu}\ge\frac{\Delta}{2}.
    $
\end{proof}

Combining Lemma~\ref{breglow} and \eq{simpleregret}, we have
\begin{equation}\label{simmid}
    \operatorname{Reg}^{\mathrm{sim}}_t\ge \E\left[
\frac{\Delta}{16d}D^2_{\Psi}(e_{i_*},p_t)\wedge \frac{\Delta}{2}
\right].
\end{equation}
Equivalently,
\[
\operatorname{Reg}^{\mathrm{sim}}_t
\geq \frac{\Delta}{16d}\E\left[D_{\Psi}^2(e_{i_*},p_t)\wedge 8d\right]
\geq \frac{\Delta}{16d}\E\left[D_{\Psi}^2(e_{i_*},p_t)\wedge d\right].
\]
Lemma~\ref{m1m2}, applied with $X=D_{\Psi}(e_{i_*},p_t)$ and $M_2=d$, shows that it is therefore enough to prove
\[
\E[D^2_{\Psi}(e_{i_*},p_t)]\leq \frac{Cd\,e^{\alpha^2\mu_{i_*}}}{1-\alpha^2\mu_{i_*}}.
\]
Indeed, since $e^{\alpha^2\mu_{i_*}}/(1-\alpha^2\mu_{i_*})\geq1$,
\[
\left(1+\sqrt{\frac{C e^{\alpha^2\mu_{i_*}}}{1-\alpha^2\mu_{i_*}}}\right)^{-2}
\geq \frac{(1-\alpha^2\mu_{i_*})e^{-\alpha^2\mu_{i_*}}}{(1+\sqrt{C})^2}.
\]
Thus, choosing $M_1$ slightly larger than the preceding second-moment bound gives the factor required in Lemma~\ref{simlow}.

To reduce this remaining moment estimate to a quantity controlled below, note that Jensen's inequality gives
\[
    D_{\Psi}(e_{i_*},p_t)=2\sum_{i=1}^d\sqrt{p_{t,i}}+\frac{2}{\sqrt{p_{t,i_*}}}-4\leq 2\sqrt{d}+\frac{2}{\sqrt{p_{t,i_*}}},
\]
and Lemma~\ref{4-2} gives
\begin{equation}\label{upper1/p}
    \frac{1}{p_{t,i_*}}\leq 2d+\left(\eta_t\underline{\hat{L}}_{t,i_*}\right)^{2}/2.
\end{equation}
Thus, by $(a+b)^2\leq 2(a^2+b^2)$,
\[
    D_{\Psi}^2(e_{i_*},p_t)\leq 8d+\frac{8}{p_{t,i_*}}
    \leq 24d+4\left(\eta_t\underline{\hat{L}}_{t,i_*}\right)^{2}.
\]
Since $\eta_t^2=\alpha^2/t$, a linear bound for $\E[\underline{\hat{L}}_{t,i_*}^2]$ is sufficient to obtain the required uniform second-moment bound for the Bregman divergence.
\begingroup

Taking $s=1$, choosing the deterministic optimal arm
$i=i_*$, and taking a deterministic endpoint in
Lemma~\ref{lem:fixed-arm-maximal}, we obtain
\begin{equation}
\label{eq:fixed-arm-linear-specialization}
\begin{aligned}
\E[\underline{\hat L}_{t,i_*}^{2}]
&\le \E\left[\max_{1\le u\le t}
\underline{\hat L}_{u,i_*}^{2}\right]\\
&\le C\left[
\frac d{\alpha^2}
+(1+d\mu_{i_*})
\frac{e^{\alpha^2\mu_{i_*}}}{1-\alpha^2\mu_{i_*}}
\right]t.
\end{aligned}
\end{equation}
\endgroup

We now substitute the linear estimate in
\eq{eq:fixed-arm-linear-specialization} into the pointwise Bregman bound
above. Since $\eta_t^2=\alpha^2/t$,
\eq{eq:fixed-arm-linear-specialization} gives
\[
\begin{aligned}
\E[D^2_{\Psi}(e_{i_*},p_t)]
&\leq 24d+4\eta_t^2\E[\underline{\hat{L}}_{t,i_*}^2]\\
&\leq 24d+4C\left[d+\alpha^2(1+d\mu_{i_*})\frac{e^{\alpha^2\mu_{i_*}}}{1-\alpha^2\mu_{i_*}}\right]\\
&\leq \frac{Cd\,e^{\alpha^2\mu_{i_*}}}{1-\alpha^2\mu_{i_*}}.
\end{aligned}
\]
Thus the linear growth in
\eq{eq:fixed-arm-linear-specialization}
is exactly canceled by the factor $\eta_t^2$, making the Bregman second moment uniform in $t$. Applying Lemma~\ref{m1m2} with $X=D_{\Psi}(e_{i_*},p_t)$ and $M_2=d$, there exists a constant $C'>0$ such that
\begin{equation}\label{applym1m2}
    \E\left[D^2_{\Psi}(e_{i_*},p_t)\wedge d\right]
\ge \frac{C'(1-\alpha^2\mu_{i_*})}{e^{\alpha^2\mu_{i_*}}} \left(\E[D_{\Psi}(e_{i_*},p_t)]\right)^2.
\end{equation}
Finally, combining \eq{simmid} and \eq{applym1m2}, together with the displayed truncation comparison, yields
\[
\begin{aligned}
\operatorname{Reg}^{\mathrm{sim}}_t
&\geq \frac{\Delta}{16d}\E\left[D_{\Psi}^2(e_{i_*},p_t)\wedge 8d\right]\\
&\geq \frac{\Delta}{16d}\E\left[D_{\Psi}^2(e_{i_*},p_t)\wedge d\right]\\
&\geq \frac{C(1-\alpha^2\mu_{i_*})\Delta}{d\,e^{\alpha^2\mu_{i_*}}}
\left(\E[D_{\Psi}(e_{i_*},p_t)]\right)^2,
\end{aligned}
\]
where the numerical factor is absorbed into the universal constant $C$. This proves Lemma~\ref{simlow}.

\section{Other Omitted Proofs}
This appendix gives the omitted proof of the Bregman
decomposition and then proves the Bregman-divergence and simple-regret lower
bounds in Theorem~\ref{thm:lower}.
\subsection{Proof for Lemma \ref{decompo}}\label{proofdecompo}
\begin{proof}
Because $\hat{L}_{t+1}=\hat{L}_t+\hat{\ell}_t$, Lemma~\ref{exist}, applied to $p_t$ and $\widetilde p_{t+1}$, gives
\[
\eta_t\agp{p_t-e_{i_*},\hat{\ell}_t}
=\agp{p_t-e_{i_*},\nabla\Psi(p_t)-\nabla\Psi(\widetilde p_{t+1})}.
\]
The simplex multipliers vanish in the inner product because $p_t$ and $e_{i_*}$ both have unit mass. Applying Lemma~\ref{xyz} with $x=p_t$, $y=\widetilde p_{t+1}$, and $z=e_{i_*}$ gives
\[
\begin{aligned}
\agp{p_t-e_{i_*},\hat{\ell}_t}
={}&\frac{1}{\eta_t}D_\Psi(p_t,\widetilde p_{t+1})
+\frac{1}{\eta_t}D_\Psi(e_{i_*},p_t)\\
&-\frac{1}{\eta_t}D_\Psi(e_{i_*},\widetilde p_{t+1}).
\end{aligned}
\]
It remains to compare the last term with the Bregman divergence at $p_{t+1}$. By the definition of the Bregman divergence and Lemma~\ref{exist}, applied to $p_{t+1}$ and $\widetilde p_{t+1}$,
\[
\begin{aligned}
&\frac{1}{\eta_{t+1}}D_\Psi(e_{i_*},p_{t+1})
-\frac{1}{\eta_t}D_\Psi(e_{i_*},\widetilde p_{t+1})\\
={}&\left(\frac{1}{\eta_{t+1}}-\frac{1}{\eta_t}\right)
\left(\Psi(e_{i_*})-\Psi(p_{t+1})\right)\\
&+\frac{1}{\eta_t}\left[\Psi(\widetilde p_{t+1})-\Psi(p_{t+1})
+\eta_t\agp{\widetilde p_{t+1}-p_{t+1},\hat L_{t+1}}\right].
\end{aligned}
\]
Since $\widetilde p_{t+1}$ minimizes the FTRL objective with learning rate $\eta_t$, the term in square brackets is nonpositive. Moreover, $\eta_t\geq\eta_{t+1}$ and
\[
\begin{aligned}
\Psi(e_{i_*})-\Psi(p_{t+1})
&=4\left(\sum_i\sqrt{p_{t+1,i}}-1\right)\\
&\leq4\sum_{i\ne i_*}\sqrt{p_{t+1,i}}
\leq2D_\Psi(e_{i_*},p_{t+1}).
\end{aligned}
\]
Therefore,
\[
\begin{aligned}
&\frac{1}{\eta_{t+1}}D_\Psi(e_{i_*},p_{t+1})
-\frac{1}{\eta_t}D_\Psi(e_{i_*},\widetilde p_{t+1})\\
&\quad\leq2\left(\frac{1}{\eta_{t+1}}-\frac{1}{\eta_t}\right)D_\Psi(e_{i_*},p_{t+1}).
\end{aligned}
\]
Adding and subtracting $\eta_{t+1}^{-1}D_\Psi(e_{i_*},p_{t+1})$ in the exact identity and substituting the preceding bound gives
\[
\begin{aligned}
\agp{p_t-e_{i_*},\hat{\ell}_t}
\leq{}&\frac{1}{\eta_t}D_\Psi(p_t,\widetilde p_{t+1})
+\frac{1}{\eta_t}D_\Psi(e_{i_*},p_t)\\
&-\left(\frac{2}{\eta_t}-\frac{1}{\eta_{t+1}}\right)D_\Psi(e_{i_*},p_{t+1}),
\end{aligned}
\]
which proves the claim.
\end{proof}
\subsection{Proof of Theorem \ref{thm:lower}}\label{sec:pflower}
    Combining \eq{2sqrt} and Lemma \ref{4-2}, we have
    \[
    \E[D_{\Psi}(e_{i_*},p_t)]\ge2\sum_{i \ne i_*} \E[\sqrt{p_{t,i}}]\ge 4\sum_{i \ne i_*}\E\left[\left(\eta_t\underline{\hat{L}}_{t,i}+2\sqrt{d}\right)^{-1}\right].
    \]
    Then, by Jensen's inequality, it follows that
    \[
    \E[D_{\Psi}(e_{i_*},p_t)]\ge4\sum_{i \ne i_*}\left(\eta_t\E\left[\underline{\hat{L}}_{t,i}\right]+2\sqrt{d}\right)^{-1}.
    \]
    Hence, it suffices to bound $\E\left[\underline{\hat{L}}_{t,i}\right]$ for any $i\ne i_*$. Note that by the definition, we have
    \[
    \underline{\hat{L}}_{t,i}=\hat{L}_{t,i}-\min_{i'\in\mathcal{I}}\hat{L}_{t,i'}=\left(\hat{L}_{t,i}-\hat{L}_{t,i_*}\right)+\left(\hat{L}_{t,i_*}-\min_{i'\in\mathcal{I}}\hat{L}_{t,i'}\right)=\left(\hat{L}_{t,i}-\hat{L}_{t,i_*}\right)+\underline{\hat{L}}_{t,i_*}.
    \]
    On one hand, since $\hat{L}_{t}$ is an unbiased estimator for $(t-1)\mu$, we have
    \[
    \E\left[\left(\hat{L}_{t,i}-\hat{L}_{t,i_*}\right)\right]=(t-1)(\mu_i-\mu_{i_*})\leq t\Delta_i.
    \]
    On the other hand, by \eq{eq:fixed-arm-linear-specialization} and Cauchy-Schwarz inequality, we have
    \[
    \E\left[\underline{\hat{L}}_{t,i_*}\right]\leq \left(\E\left[\underline{\hat{L}}_{t,i_*}^2\right]\right)^{\frac12}\leq C_{\alpha}\sqrt{dt},
    \]
    where $C_{\alpha}=\left(\frac{C\,e^{\alpha^2}}{\alpha^2(1-\alpha^2)}\right)^{\frac12}$ and $C$ is the universal constant in \eq{eq:fixed-arm-linear-specialization}. Hence,
    \[
    \E\left[\underline{\hat{L}}_{t,i}\right]\leq t\Delta_i+C_{\alpha}\sqrt{dt}.
    \]
    Recall that $\eta_t=\alpha/\sqrt{t}$, then we have
    \[
    \E[D_{\Psi}(e_{i_*},p_t)]\ge4\sum_{i \ne i_*}\left(\eta_t\E\left[\underline{\hat{L}}_{t,i}\right]+2\sqrt{d}\right)^{-1}\ge4\sum_{i \ne i_*}\left(\alpha\Delta_i\sqrt{t}+(2+\alpha C_{\alpha})\sqrt{d}\right)^{-1}.
    \]
    When $t\ge C'_{\alpha}\frac{d}{\Delta^2}$, where $C'_\alpha=\left(\frac{2+\alpha C_{\alpha}}{\alpha}\right)^2$, for any $i\ne i_*$, we have $\alpha\Delta_i\sqrt{t}\ge(2+\alpha C_{\alpha})\sqrt{d}$. Therefore,
    {
    \[
    \E[D_{\Psi}(e_{i_*},p_t)]
    \ge2\alpha^{-1}\sum_{i \ne i_*}\frac{t^{-1/2}}{\Delta_i}
    \ge \frac{2}{\alpha\Delta}t^{-1/2}.
    \]}

\begingroup

For the simple-regret lower bound, we use the same estimate of
$\E[\underline{\hat L}_{t,i}]$ and apply the lower sampling-probability bound in
Lemma~\ref{4-2} directly.
For every $i\ne i_*$, Lemma~\ref{4-2} and Jensen's inequality give
\[
\begin{aligned}
\E[p_{t,i}]
&\ge 4\E\left[\left(
\eta_t\underline{\hat L}_{t,i}+2\sqrt d
\right)^{-2}\right]\\
&\ge 4\left(
\eta_t\E[\underline{\hat L}_{t,i}]+2\sqrt d
\right)^{-2}.
\end{aligned}
\]
Moreover, the preceding first-moment estimate and
$t\ge C'_\alpha d/\Delta^2$ imply
\[
\begin{aligned}
\eta_t\E[\underline{\hat L}_{t,i}]+2\sqrt d
&\le \alpha\Delta_i\sqrt t+(2+\alpha C_\alpha)\sqrt d\\
&\le 2\alpha\Delta_i\sqrt t.
\end{aligned}
\]
Consequently,
\[
\E[p_{t,i}]\ge \frac{1}{\alpha^2\Delta_i^2t}.
\]
Multiplying by $\Delta_i$, summing over $i\ne i_*$, and using
\eq{simpleregret}, we obtain
\[
\operatorname{Reg}^{\mathrm{sim}}_t
\ge \frac{1}{\alpha^2t}\sum_{i\ne i_*}\frac{1}{\Delta_i}
\ge \frac{1}{\alpha^2\Delta t}.
\]
\endgroup

\section{Derived Results Based on \citet{zimmert2021tsallis}}
For completeness, this section derives Lemma~\ref{stab} and
records the stability identities used in
Appendix~\ref{sec:fixed-arm-moments} from the following results of
\citet{pmlr-v89-zimmert19a}. We follow the definition there: for any $\lambda\in\R^d$,
\[
\Phi_t(\lambda):=\max _{w \in \Delta^{d-1}}\left\{\langle w,\lambda\rangle-\eta_t^{-1}\Psi(w)\right\}.
\]

\begin{Lem}[Lemma 11 (1) in \citep{pmlr-v89-zimmert19a}]\label{111}
If $\eta_t>0$, then
    \[
    \agp{p_t,\hat{\ell}_t}+\Phi_t\left(-\hat{L}_{t+1}\right)-\Phi_t\left(-\hat{L}_{t}\right)\leq \frac{\eta_t}{2}\sum_{i=1}^d \frac{A_{t,i}}{\sqrt{p_{t,i}}}.
    \]
\end{Lem}

\begin{Lem}[Lemma 11 (2) in \citep{pmlr-v89-zimmert19a}]\label{112}
If $0<\eta_t< 1$, then
    \[
    \mathbb{E}\left[
    \agp{p_t,\hat{\ell}_t}
    +\Phi_t\left(-\hat{L}_{t+1}\right)-\Phi_t\left(-\hat{L}_{t}\right)\bigg{|}\mathscr{F}_{t-1}\right]\leq \frac{\eta_t^2}{2}+\sum_{i=1}^d \frac{\eta_t}{2} p_{t, i}^{\frac{1}{2}}\left(1-p_{t, i}\right).
    \]
\end{Lem}
\begin{Rema}
    First, we have adapted the remaining notations to match those used in this paper. It is worth noting that in \cite{pmlr-v89-zimmert19a}, the subscript of $\hat{L}_t$ is smaller by one compared to ours; that is, their $\hat{L}_t$ actually corresponds to $\hat{L}_{t+1}$ in our notation. Second, we present the deterministic (conditional expectation) version of their result, which follows directly from the proof of Lemma 11 in their paper.
\end{Rema}
To derive Lemma~\ref{stab} and the interval stability estimate
used in Appendix~\ref{sec:fixed-arm-moments}, we need the following
observations:
\begin{Lem}\label{stabequi}
    Let $\widetilde{p}_{t+1}=\phi(-\eta_t\hat{L}_{t+1})$, then
    \[
    \agp{p_t,\hat{\ell}_t}
    +\Phi_t\left(-\hat{L}_{t+1}\right)-\Phi_t\left(-\hat{L}_{t}\right)=\frac{1}{\eta_t}D_\Psi(p_t,\widetilde{p}_{t+1}).
    \]
    \begin{proof}
        By the definition of $\Phi$ and $\phi$, one can see that
        \[
        \Phi_t(-\hat{L}_{t+1})=\agp{\widetilde{p}_{t+1},-\hat{L}_{t+1}}-\eta_t^{-1}\Psi(\widetilde{p}_{t+1}), \Phi_t(-\hat{L}_{t})=\agp{p_t,-\hat{L}_{t}}-\eta_t^{-1}\Psi(p_t).
        \]
        Hence, the left hand equals
        \begin{equation}\label{left}
            \eta_t^{-1}\Psi(p_t)-\eta_t^{-1}\Psi(\widetilde{p}_{t+1})-\agp{p_t-\widetilde{p}_{t+1},-\hat{L}_{t+1}}.
        \end{equation}
        By Lemma~\ref{exist}, there exists $\nu\in\R$ such that
        \[
        \nabla\Psi(\widetilde{p}_{t+1})=-\eta_t\hat{L}_{t+1}+\nu\mathbf{1} ,
        \]
        then \eq{left} is equal to
        \[
        \eta_t^{-1}\Psi(p_t)-\eta_t^{-1}\Psi(\widetilde{p}_{t+1})-\eta_t^{-1}\agp{p_t-\widetilde{p}_{t+1},\nabla\Psi(\widetilde{p}_{t+1})}=\frac{1}{\eta_t}D_\Psi(p_t,\widetilde{p}_{t+1}).
        \]
    \end{proof}
\end{Lem}

\begin{Lem}\label{x1}
Let $\widetilde{p}_{t+1}=\phi(-\eta_t\hat{L}_{t+1})$, then
    \[
    \agp{p_t-p_{t+1},\hat{\ell}_{t}}-\frac{1}{\eta_{t}}D_{\Psi}(p_{t+1},p_t)\leq \frac{1}{\eta_t}D_\Psi(p_t,\widetilde{p}_{t+1}).
    \]
    \begin{proof}
By Lemma~\ref{exist}, there exist $\nu_1,\nu_2$ and $\nu_3\in\R$ such that
\begin{equation}
    \nabla\Psi(p_t)=-\eta_t\hat{L}_t+\nu_1\mathbf{1}, \nabla\Psi(\widetilde{p}_{t+1})=-\eta_t\hat{L}_{t+1}+\nu_2\mathbf{1} \,\text{and }\nabla\Psi(p_{t+1})=-\eta_{t+1}\hat{L}_{t+1}+\nu_3\mathbf{1}.
\end{equation}
Then 
\[
\agp{p_t-p_{t+1},\eta_t\hat{\ell}_{t}}=\agp{p_t-p_{t+1},\nabla\Psi(p_t)-\nabla\Psi(\widetilde{p}_{t+1})},
\]
which, by Lemma \ref{xyz} ($x=p_t, y=\widetilde{p}_{t+1}$ and $z=p_{t+1}$), equals
\[
D_\Psi(p_t,\widetilde{p}_{t+1})+D_\Psi(p_{t+1},p_t)-D_\Psi(p_{t+1},\widetilde{p}_{t+1}).
\]
Hence,
\[
\agp{p_t-p_{t+1},\hat{\ell}_{t}}-\frac{1}{\eta_{t}}D_{\Psi}(p_{t+1},p_t)=\frac{1}{\eta_t}\left(D_\Psi(p_t,\widetilde{p}_{t+1})-D_\Psi(p_{t+1},\widetilde{p}_{t+1})\right)\leq \frac{1}{\eta_t}D_\Psi(p_t,\widetilde{p}_{t+1}).
\]
    \end{proof}
\end{Lem}
Combining Lemma \ref{stabequi} and Lemma \ref{x1}, for any $t\ge 1$, we have
\begin{equation}\label{bri}
    \agp{p_t-p_{t+1},\hat{\ell}_{t}}-\frac{1}{\eta_{t}}D_{\Psi}(p_{t+1},p_t)\leq \agp{p_t,\hat{\ell}_t}
    +\Phi_t\left(-\hat{L}_{t+1}\right)-\Phi_t\left(-\hat{L}_{t}\right).
\end{equation}

\subsection{Proof for Lemma \ref{stab}}\label{proof1}
Note that $1>\eta_{t}\ge \eta_{t+1}$, then combining Lemma \ref{stabequi} and Lemma \ref{112}, we have
\[
\E_{t-1}[\mathbf{\uppercase\expandafter{\romannumeral1}}]\leq \frac{\eta_t^2}{2}+\sum_{i=1}^d \frac{\eta_t}{2} p_{t, i}^{\frac{1}{2}}\left(1-p_{t, i}\right).
\]
Clearly, one can see that
\[
\sum_{i=1}^d  p_{t, i}^{\frac{1}{2}}\left(1-p_{t, i}\right)\leq 1-p_{t, i_*}+ \sum_{i\ne i_*}^d  p_{t, i}^{\frac{1}{2}}\left(1-p_{t, i}\right)\leq \sum_{i\ne i_*}^d  p_{t, i}+\sum_{i\ne i_*}^d  p_{t, i}^{\frac{1}{2}}\leq 2\sum_{i\ne i_*}^d  p_{t, i}^{\frac{1}{2}}.
\]
Then since $\eta_t\leq \frac{1}{\sqrt{t}}$, we have
\[
\E_{t-1}[\mathbf{\uppercase\expandafter{\romannumeral1}}]\leq \frac{1}{2t}+\sum_{i\ne i_*}^d\sqrt{\frac{p_{t,i}}{t}}.
\]

\section{Auxiliary Lemmas}
This appendix collects the truncated-moment inequality, the
sequence estimate, and the generalized Pythagoras identity used in the
Bregman-divergence arguments, followed by the finite-horizon stopping-time
decomposition used in the simple-regret proof.
    \begin{Lem}\label{m1m2}
        If a nonnegative random variable $X$ satisfies $\mathbb{E}[X^2]<M_1$, then for any $M_2>0$ we have
\[
\mathbb{E}[X^2 \wedge M_2] \;\ge\; \left(1+\sqrt{\frac{M_1}{M_2}}\right)^{-2} \, (\mathbb{E}[X])^2.
\]
\begin{proof}
    On one hand, by Jensen's inequality, we have
    \[
    \E[X\cdot\1_{\{X\leq \sqrt{M_2}\}}]\leq \sqrt{\E[X^2\cdot\1_{\{X\leq \sqrt{M_2}\}}]}\leq \sqrt{\mathbb{E}[X^2 \wedge M_2]}.
    \]
    On the other hand, by Cauchy-Schwarz inequality, we have
    \[
    \E[X\cdot\1_{\{X> \sqrt{M_2}\}}]\leq \sqrt{\E[X^2]\P(X>\sqrt{M_2})}\leq \sqrt{M_1\P(X>\sqrt{M_2})}\leq \sqrt{\frac{M_1}{M_2}\mathbb{E}[X^2 \wedge M_2]}.
    \]
    Hence,
    \[
    \E[X]=\E[X\cdot\1_{\{X\leq \sqrt{M_2}\}}]+\E[X\cdot\1_{\{X> \sqrt{M_2}\}}]\leq \left(1+\sqrt{\frac{M_1}{M_2}}\right)\sqrt{\mathbb{E}[X^2 \wedge M_2]},
    \]
    which completes the proof.
\end{proof}
    \end{Lem}

\begingroup
\begin{Lem}\label{ratelemma}
    Given a nonnegative sequence $(x_t)_{t=1}^{+\infty}$, if there exist constants $C_1,C_2>0$ such that for any $t\geq 1$,
\[
x_t^2\leq\frac{C_1}{t}+C_2\sqrt{t}\left[x_t-\left(2-\sqrt{\frac{t+1}{t}}\right)x_{t+1}\right],
\]
then for any $t\geq 1$, we have
\[
x_t\leq\frac{C}{\sqrt{t}},
\]
where $C=x_1\vee2\sqrt{C_1}\vee2C_2\vee\left(C_2+\frac{6C_1}{C_2}\right)$.
\begin{proof}
    We prove $\sqrt{t}\,x_t\leq C$ by induction. The assumed recursion gives
    \[
    \left(2\sqrt{\frac{t}{t+1}}-1\right)\sqrt{t+1}\,x_{t+1}
    \leq \sqrt{t}\,x_t-\frac{t x_t^2-C_1}{C_2t}.
    \]
    Assume $\sqrt{t}\,x_t\leq C$. If $C\leq C_2t/2$, the right-hand side of the preceding inequality, as a function of $\sqrt{t}\,x_t$, is increasing on $[0,C]$. Hence it is at most
    \[
    C-\frac{C^2-C_1}{C_2t}
    \leq C-\frac{C}{t}
    \leq\left(2\sqrt{\frac{t}{t+1}}-1\right)C,
    \]
    where the last inequality follows from
    \[
    2t\left(1-\sqrt{\frac{t}{t+1}}\right)\leq1.
    \]
    If $C>C_2t/2$, the same right-hand side is at most
    \[
    \frac{C_2t}{4}+\frac{C_1}{C_2t}.
    \]
    Since
    \[
    2\sqrt{\frac{t}{t+1}}-1\geq\frac{t}{t+2}
    \quad\text{and}\quad t<\frac{2C}{C_2},
    \]
    we obtain
    \[
    \begin{aligned}
    \frac{\frac{C_2t}{4}+\frac{C_1}{C_2t}}{2\sqrt{t/(t+1)}-1}
    &\leq\frac{C_2(t+2)}{4}+\frac{C_1(t+2)}{C_2t^2}\\
    &<\frac{C}{2}+\frac{C_2}{2}+\frac{3C_1}{C_2}
    \leq C.
    \end{aligned}
    \]
    Thus $\sqrt{t+1}\,x_{t+1}\leq C$ in both cases. Since $x_1\leq C$, the induction is complete.
\end{proof}
\end{Lem}
\endgroup

\begin{Lem}[Generalized Pythagoras Identity]\label{xyz}
    \[
    D_\Psi(x,y)+D_\Psi(z,x)-D_\Psi(z,y)=\agp{\nabla\Psi(x)-\nabla\Psi(y),x-z}.
    \]
    \begin{proof}
        It suffices to expand the left hand by the definition that
        \[
        D_\Psi(x,y)=\Psi(x)-\Psi(y)-\agp{x-y,\nabla\Psi(y)}.
        \]
    \end{proof}
\end{Lem}

\begin{Lem}[Finite-horizon stopping-time decomposition]
\label{lem:finite-horizon-stopping}
Let $(X_j)_{j\ge1}$ be a stochastic process adapted to a filtration
$(\mathscr F_j)_{j\ge1}$, let $(H_j)_{j\ge1}$ be a sequence of nonnegative
adapted random variables, and let $(\mathcal M_j)_{j\ge1}$ be a sequence of
Borel subsets of $\mathbb R$.  Suppose that
for every $j\ge1$,
\[
\E[H_{j+1}\mid\mathscr F_j]\le H_j
\qquad\text{on }\{X_j\in\mathcal M_j\}.
\]
Define
\[
\sigma_0:=1,
\qquad
\tau_k:=\inf\{j\ge\sigma_k:X_j\notin\mathcal M_j\},
\]
and, for $k\ge1$,
\[
\sigma_k:=\inf\{j>\tau_{k-1}:X_j\in\mathcal M_j\},
\]
with the convention that the infimum of the empty set is infinite.  Then, for
every bounded $(\mathscr F_j)$-stopping time $T$ taking values in the positive
integers,
\[
\E[H_T\1_{\{X_T\in\mathcal M_T\}}]
\le
\E[H_1\1_{\{X_1\in\mathcal M_1\}}]
+\sum_{k\ge1}\E[H_{\sigma_k}\1_{\{\sigma_k\le T\}}].
\]
\end{Lem}

\begin{proof}
The stopped-supermartingale conclusion of
\citet[Appendix~G, Lemma~G.2]{zhan2026does} applies on every visit interval
$[\sigma_k,\tau_k)$.  We only adapt the second part of their proof to the
bounded stopping time $T$ and retain the finite-horizon entrance events.

The visit intervals are disjoint and
\[
\{X_T\in\mathcal M_T\}
=\bigcup_{k\ge0}\{\sigma_k\le T<\tau_k\}.
\]
For $k\ge1$, the event $\{\sigma_k\le T\}$ belongs to
$\mathscr F_{\sigma_k}$.  On that event,
$\sigma_k\le T\wedge\tau_k$, and bounded optional sampling on the $k$th
stopped process, together with nonnegativity, gives
\[
\E[H_T\1_{\{\sigma_k\le T<\tau_k\}}]
\le
\E[H_{T\wedge\tau_k}\1_{\{\sigma_k\le T\}}]
\le
\E[H_{\sigma_k}\1_{\{\sigma_k\le T\}}].
\]
For $k=0$, the event $\{1<\tau_0\}$ is
$\{X_1\in\mathcal M_1\}$, and the same stopped-supermartingale estimate gives
\[
\E[H_T\1_{\{1\le T<\tau_0\}}]
\le
\E[H_1\1_{\{X_1\in\mathcal M_1\}}].
\]
Summing over the disjoint visit intervals proves the claim.  Only finitely many
terms can be nonzero because $T$ is bounded.
\end{proof}

\section{Verification of the parameter constraints}
\label{sec:parameter-verification}

This appendix verifies the parameter conditions used in the local and global
estimates.  Recall from Appendix~\ref{sec:parameters} that
\[
A=\left\lceil C\Delta^{-2}\log(2|\mathcal S|)\right\rceil,
\qquad
\beta=(C_\alpha A)^{-1},
\]
and that $N$ is larger than
$C'_\alpha\max\{A^2d^4,\beta^{-2}\}$.  We first fix the universal constant in
$A$, then the denominator constant in the definition of $\beta$, and finally
the leading constant in the lower bound on $N$.  This is the order used in
all verifications below.

\subsection{Conditions for the local estimates}

Suppose that $t\ge N$ and $D_t\ge-\beta t$.  The definition of $h_t$ gives
\[
Ad\le h_t\le C_\alpha A(d+\beta^2t),
\qquad
\frac{h_t}{t}
\le\frac1t+C_\alpha A\left(\frac d t+\beta^2\right)
\le\frac18,
\]
after the denominator constant in $\beta$ and the leading constant in $N$
are fixed; here $A\ge1$ and $d\ge2$ absorb the ceiling in $h_t$.
This verifies \eq{eq:adaptive-block-scales}.  The same choices give the
common conditions
\begin{equation}
\label{eq:verification-basic-scales}
\begin{gathered}
A\ge1,
\qquad
A\beta\le1,
\qquad
\sqrt t\ge C_\alpha Ad^2,
\\
t\ge\max\{A^2d,d\Delta^{-2},d\},
\qquad
\alpha\Delta\sqrt t\ge2,
\\
A\Delta^2\ge C\log(2|\mathcal S|),
\qquad
\beta=(C_\alpha A)^{-1}.
\end{gathered}
\end{equation}
Indeed, the $A^2d^4$ term in the lower bound on $N$ gives the displayed
lower bounds on $t$ and $\sqrt t$ after its leading constant is fixed.
Together with $A\Delta^2\ge C\log 2$ and $d\ge2$, the square-root bound gives
$\alpha\Delta\sqrt t\ge2$.  The inequality $A\beta\le1$ follows directly
from the denominator chosen in the definition of $\beta$.
In particular, the universal constant in $A$ can be chosen so that
\[
A\ge C\varepsilon^{-2}\Delta^{-2}\log(2|\mathcal S|),
\qquad \varepsilon\in\{1/16,1/4\}.
\]
Thus all assumptions of Corollary~\ref{cor:state-dependent-horizon} used in
the two local proofs hold.
The same definition of $\beta$, together with
$A\ge C\Delta^{-2}\log 2$ and $\Delta\le1$, also gives
\[
\beta\le\Delta.
\]

On the positive half-line, $h_t=\lceil Ad\rceil\le2Ad$.  The lower bound on
$N$, together with $d\ge2$ and $A\Delta^2\ge C\log 2$, gives
\[
(12dh_t)^2\le C A^2d^4\le t,
\qquad
t\ge A^2d,
\qquad
t\ge C_\alpha\frac{dh_t}{\Delta^2},
\qquad
h_t\le\min\left\{\frac t8,\Delta t\right\}.
\]
Moreover,
\[
\frac{4dh_t+2\sqrt t/\alpha}{\Delta t}
\le
\frac{CAd^2}{\Delta t}+\frac{2}{\alpha\Delta\sqrt t}
\le1.
\]
The bounds $(12dh_t)^2\le t$ and
$4dh_t+2\sqrt t/\alpha\le\Delta t$, together with the common condition
$\alpha\Delta\sqrt t\ge2$ in \eq{eq:verification-basic-scales}, verify
\eq{eq:positive-parameter-conditions}.  The remaining bounds in the first
display are used in the positive-starting-point parts of the contraction and
maximal proofs.  Moreover,
\[
\begin{aligned}
\frac{C\log(2|\mathcal S|)}{A}
\frac{h_t^2}{\Delta^2t^2}
&=\frac{h_t}{t}\,
\frac{C\log(2|\mathcal S|)}{A}\,
\frac{h_t}{\Delta^2t}\\
&\le\frac{h_t}{24t}.
\end{aligned}
\]
The last step uses $A\Delta^2\ge C\log(2|\mathcal S|)$ and
$t\ge C_\alpha dh_t/\Delta^2$.  This is the second-moment remainder in the
positive-axis ratio calculation.

For the negative strip, the inverse-square definition gives
\[
\sqrt{g_t(D_t)}=
\frac{2t}{2\sqrt t+\alpha(-D_t)}.
\]
The denominator constant in $\beta$ first makes the term containing
$A\beta$ below small; the leading constant in $N$ then makes the term
containing $Ad/\sqrt t$ small.  The definition of $h_t$ therefore gives
\[
\frac{h_t\sqrt{g_t(D_t)}}{t}
\le C_\alpha A\left(\frac d{\sqrt t}+\beta\right)
\le\frac18.
\]
The lower bound $\sqrt d\ge1$ in the definition of $h_t$ also gives
\[
\frac{h_t}{t}g_t(D_t)\ge A.
\]
Moreover, $D_t<0$ gives $\sqrt{g_t(D_t)}\le\sqrt t$, whereas
$t\ge\beta^{-2}$, $-D_t\le\beta t$, and $\beta\le\Delta$ give
\[
\Delta\sqrt{g_t(D_t)}
\ge\frac{2\Delta t}{2\sqrt t+\alpha\beta t}
\ge\frac23.
\]
The proof in Appendix~\ref{sec:local-estimates} uses
$h_t\sqrt{g_t(D_t)}/t\le1/8$ both to deduce a large value of $R_t^+$ from
$v_t>1/4$ and to obtain the denominator bound in the maximal estimate.  The
bound $(h_t/t)g_t(D_t)\ge A$ absorbs the two exponentially unlikely classes
of paths.  The bound $\Delta\sqrt{g_t(D_t)}\ge2/3$ converts the drift on
$\{\kappa_t>t+h_t,\ v_t\le1/4\}$ into a multiple of $h_t/t$.
Also,
\[
\frac{h_t}{\sqrt t/\alpha-D_t}
\le\frac{\alpha h_t\sqrt{g_t(D_t)}}{t}
\le\frac14.
\]
This verifies the condition used in Lemma~\ref{lem:negative-persistence} to
bound the probability that $D_u$ reaches $\sqrt t/\alpha$ before $t+h_t$.
The lower bound $h_t\ge Ad$ and the choice of $A$ imply
\[
\sqrt{8dh_t}\le\frac{\Delta h_t}{4}.
\]
Furthermore, $t/\sqrt{g_t(D_t)}\ge\sqrt t$, so the chosen leading constant
in $N$ makes
\[
\exp\left\{-\frac{t}{C_\alpha d\sqrt{g_t(D_t)}}\right\}\le\frac14.
\]

To verify \eq{eq:negative-normalized-remainders}, divide its four positive
terms by $\Delta h_t\sqrt{g_t(D_t)}/t$.  The resulting ratios satisfy, by
\eq{eq:verification-basic-scales},
\begin{align*}
\frac1{\Delta\sqrt{g_t(D_t)}}
&\le\frac1{\Delta\sqrt t}+\frac{\alpha\beta}{2\Delta},\\
\frac{d\sqrt{g_t(D_t)}}{\Delta t}
&\le\frac d{\Delta\sqrt t},\\
\frac{\log(2|\mathcal S|)h_t\sqrt{g_t(D_t)}}{A\Delta t}
&\le\frac{C_\alpha\log(2|\mathcal S|)}\Delta
\left(\frac d{\sqrt t}+\beta\right),\\
\frac{\Delta h_t\sqrt{g_t(D_t)}}t
&\le C_\alpha\Delta A
\left(\frac d{\sqrt t}+\beta\right).
\end{align*}
The four right-hand sides satisfy
\begin{align*}
\frac1{\Delta\sqrt t}+\frac\beta\Delta
&\le \frac{1}{C_\alpha\Delta Ad^2}
+\frac{1}{C_\alpha\Delta A},\\
\frac d{\Delta\sqrt t}
&\le\frac{1}{C_\alpha\Delta Ad},\\
\frac{\log(2|\mathcal S|)}\Delta
\left(\frac d{\sqrt t}+\beta\right)
&\le\frac{C}{C_\alpha},\\
\Delta A\left(\frac d{\sqrt t}+\beta\right)
&\le\frac{C}{C_\alpha}.
\end{align*}
Since $A\Delta^2\ge C\log(2|\mathcal S|)$ and $d\ge2$, all four quantities
can be made jointly smaller than the fixed fraction required by the
elementary bound on $(1-v_t)^{-2}$.  More precisely, after fixing the constant
multiplying the second-order terms in Appendix~\ref{sec:local-estimates}, the
denominator constant
in $\beta$ and the leading constant in $N$ can be chosen so that
\[
C_\alpha\left[
\frac1{\Delta\sqrt{g_t(D_t)}}
+\frac{d\sqrt{g_t(D_t)}}{\Delta t}
+\frac{\log(2|\mathcal S|)h_t\sqrt{g_t(D_t)}}{A\Delta t}
+\frac{\Delta h_t\sqrt{g_t(D_t)}}t
\right]
\le\frac{\alpha}{8}.
\]
Multiplying this joint bound by
$\Delta h_t\sqrt{g_t(D_t)}/t$ proves
\eq{eq:negative-normalized-remainders}.  Appendix~\ref{sec:local-estimates}
then uses $\Delta\sqrt{g_t(D_t)}\ge2/3$ to obtain
\eq{eq:negative-no-hit-good}.

Finally, each of the classes
$\{\kappa_t>t+h_t,\ v_t>1/4\}$ and
$\{\kappa_t\le t+h_t\}$ contributes at most
$C_\alpha t\exp\{-t/(C_\alpha d\sqrt{g_t(D_t)})\}$.  Since
$t/\sqrt{g_t(D_t)}\ge\sqrt t$ and $e^{-x}\le Cx^{-4}$,
\[
C_\alpha t\exp\left\{-\frac{t}
{C_\alpha d\sqrt{g_t(D_t)}}\right\}
\le\frac{C_\alpha d^4}{t}
\le\frac{A}{8C_\alpha}.
\]
The inequality $C_\alpha d^4/t\le A/(8C_\alpha)$ is obtained by choosing the
final leading constant in $N$; this is possible because
$t\ge C'_\alpha A^2d^4$ and
$A\ge1$.  Comparing this display with
$(h_t/t)g_t(D_t)\ge A$ gives
\eq{eq:exceptional-payment-absorption}.

\subsection{Conditions for the global estimates}

In addition to the already verified inequality $\beta\le\Delta$, the choices
above imply
\[
N\ge\max\{\beta^{-2},d\Delta^{-2},d\},
\qquad
\beta N\ge2(d+\beta).
\]
Indeed,
\[
A^2d^4\ge d\Delta^{-2},
\qquad
\beta N\ge C'_\alpha\beta A^2d^4
\ge C A d^4\ge2(d+\beta),
\]
after the leading constant in $N$ is fixed.  The lower bound on $N$ also
contains $\beta^{-2}$ and dominates $d$, and its leading constant ensures
$N\ge8$.  At a skeleton node in the strip,
\eq{eq:adaptive-block-scales} gives $h_t\le t/8$; outside the strip, the
definition gives $h_t=1\le t/8$.  These two cases verify
\eq{eq:skeleton-step-size}.  The bound
$\beta N\ge2(d+\beta)$ is used in the entrance-size step of the proof of
Lemma~\ref{lem:entrance-budget}, leading to
\eq{eq:weighted-entrance-potential}, while $N\ge d\Delta^{-2}$ is used in
\eq{eq:initial-potential}.

\subsection{Simplification of the final dependence}

Since $|\mathcal S|\le d$,
\begin{align*}
A&=O\bigl(\Delta^{-2}\log(2|\mathcal S|)\bigr),\\
\beta^{-2}&=O_\alpha\bigl(
\Delta^{-4}\log^2(2|\mathcal S|)\bigr).
\end{align*}
It follows directly from the definition of $N$ that
\[
N=O_\alpha\left(
d^4\Delta^{-4}\log^2(2|\mathcal S|)
\right),
\]
whereas the entrance contribution satisfies
\[
\beta^{-2}(1+|\mathcal S|\beta^{-2})
=O_\alpha\left(
\Delta^{-4}\log^2(2|\mathcal S|)
+|\mathcal S|\Delta^{-8}\log^4(2|\mathcal S|)
\right).
\]
Using $|\mathcal S|\le d$ and $\Delta\le1$, the complete numerator is at
most $C_\alpha d^4\Delta^{-8}\log^4(2d)$, as used in the proof of
Theorem~\ref{thm:coarse-uniform-moment}.  The preceding expression is more
informative when $|\mathcal S|\ll d$.

\end{document}